\documentclass{article}

\usepackage{arxiv}

\usepackage[utf8]{inputenc} % allow utf-8 input
\usepackage[T1]{fontenc}    % use 8-bit T1 fonts
\usepackage{natbib}         % citation commands
\usepackage{hyperref}       % hyperlinks
\usepackage{url}            % simple URL typesetting
\usepackage{booktabs}       % professional-quality tables
\usepackage{amsfonts}       % blackboard math symbols
\usepackage{nicefrac}       % compact symbols for 1/2, etc.
\usepackage{microtype}      % microtypography
\usepackage{xcolor}         % colors

\newif\ifshowchanges
\showchangestrue
\definecolor{revisionred}{RGB}{200,0,0}
\definecolor{revisiongreen}{RGB}{0,115,55}

\definecolor{revisionblue}{RGB}{0,76,190}

\usepackage{amsmath}        % align, gather, etc. (already implicit, but be explicit)
\usepackage{amssymb}        % \mathbb, \mathcal, etc.
\usepackage{amsthm}         % theorem environments + proof environment
\usepackage{mathtools}      % small fixes/extensions over amsmath

\newtheorem{theorem}{Theorem}
\newtheorem{lemma}{Lemma}
\newtheorem{proposition}{Proposition}
\newtheorem{corollary}{Corollary}
\theoremstyle{definition}

\newtheorem{assumption}{Assumption}
\theoremstyle{remark}
\newtheorem{remark}{Remark}

\newcommand{\E}{\mathbb{E}}
\newcommand{\one}{\mathbf{1}}
\newcommand{\norm}[1]{\left\lVert #1 \right\rVert}

\usepackage{algorithm}
\usepackage{algpseudocode}
\newcommand{\acc}[4]{\ensuremath{#1\pm#2\,/\,#3\pm#4}}
\usepackage{graphicx}
\usepackage{subcaption}
\usepackage{multirow}

\newenvironment{ack}{\section*{Acknowledgments and Disclosure of Funding}}{}

\title{DPA: Decentralized Primal Averaging with Quasi-Global Momentum for Highly Heterogeneous Data}

\author{%
  Diyako Ghaderyan\thanks{%
    Department of Information and Communications Engineering,
    Aalto University, 02150 Espoo, Finland.
    E-mail: \texttt{diyako.ghaderyan@aalto.fi}.%
  }
  \And
  Stefan Werner\thanks{%
    Department of Electronic Systems,
    Faculty of Information Technology and Electrical Engineering,
    Norwegian University of Science and Technology,
    7034 Trondheim, Norway; also with the Department of Information
    and Communications Engineering, Aalto University,
    02150 Espoo, Finland.
    E-mail: \texttt{stefan.werner@ntnu.no}.%
  }
}

\date{}
\renewcommand{\shorttitle}{DPA: Decentralized Primal Averaging}

\begin{document}

\maketitle

\begin{abstract}
Decentralized training of deep learning models {keeps training data
local and supports} on-device learning over networks. In realistic scenarios,
heterogeneity across agents' local data poses an optimization challenge and
can severely degrade test accuracy, especially on sparse communication
graphs. Building on recent advances in centralized primal averaging, we
propose Decentralized Primal Averaging (DPA), a decentralized learning
algorithm that combines primal averaging with {a local quasi-global momentum
buffer}. DPA maintains three coupled sequences for optimization, gradient
evaluation, and output averaging, while using consecutive mixed iterates to
construct a network-aware momentum direction without transmitting an
auxiliary tracking variable. For smooth nonconvex objectives, {{we establish a convergence bound for DPA and identify conditions for linear speedup in the number of agents for sufficiently many iterations.}} We also introduce DPA-1G,
{a variant that uses one communication round per iteration and matches} {the} {{per-iteration}}
{communication cost} of DSGD-style baselines. Experiments following the GUT
decentralized image-classification benchmark show that DPA {compares
favorably with} DSGD, gradient tracking, QG-DSGDm, momentum tracking, and
QG-GUTm across datasets, architectures, topologies, and heterogeneity levels,
with the largest gains under high heterogeneity {on sparse rings. At the same total communication cost, both DPA variants achieve
higher mean test accuracy than the best baseline on CIFAR-10
and CIFAR-100 with ResNet-20. {Our code is publicly available.}\footnote{%
{\url{https://github.com/Diyako754/DPA}}}}
\end{abstract}

% ============================================================
%  MAIN PAPER
% ============================================================

\section{Introduction}
\label{sec:intro}

Training modern machine learning models, especially deep neural networks,
often requires large datasets and {substantial} computational resources. This has
motivated distributed training methods, where the workload is split across
{many agents}. Early work on decentralized optimization~\citep{tsitsiklis1984problems}
and gossip-style protocols~\citep{xiao2004fast} {{showed how agents can reach agreement} without a central coordinator}. More recent work has
shown that decentralized stochastic gradient methods can be effective for
large-scale machine learning~\citep{lian2017can}. In contrast, the classical
master--worker setup relies on a centralized parameter server, where workers
read model parameters and send gradient updates to shared server
shards~\citep{dean2012large}. Although this approach is effective, the server
can become a communication bottleneck as the number of workers increases.

Decentralized learning provides an alternative that avoids a central server.
It is also useful when data should remain local~\citep{nedic2020distributed}. {In this setting, agents}
are connected through a communication graph and exchange messages only with
their neighbors. This reduces the per-iteration communication to local
peer-to-peer exchanges~\citep{assran2019stochastic}. Under standard
assumptions, decentralized stochastic gradient methods can achieve the same
$\mathcal{O}(1/\sqrt{nT})$ rate as centralized mini-batch SGD, giving a linear
speedup with respect to the number of agents $n$ {for sufficiently many iterations}~\citep{koloskova2020unified,lian2017can}.

A key step in these methods is gossip averaging. At each iteration, every
agent mixes its local model with the models received from neighboring agents,
using weights defined by the communication graph~\citep{xiao2004fast}. Since
this mixing is local, information does not spread through the whole network
immediately. As a result, the local models can differ from each other during
training, especially in the early iterations. The communication graph therefore
affects both the {communication cost} and the convergence behavior. Dense graphs
spread information faster but require more communication per agent, while
sparse graphs {require less communication per agent} but usually mix more slowly~\citep{ghaderyan2023fast,pu2021distributed}.

The problem becomes harder when the data are heterogeneous across agents. In
many practical settings, different agents observe data from different
distributions. Then the local objectives may be quite different, and the local
gradients may point in different directions. This can increase model drift,
slow down training, and reduce test accuracy~\citep{DBLP:conf/icml/0004KSJ21,hsieh2020non}.
Several methods have been proposed to reduce this effect. {Gradient
tracking (GT)}~\citep{di2016next,nedic2017achieving,koloskova2021improved} and
related methods such as $D^2$~\citep{tang2018d} correct local updates using
information that better approximates the global descent direction. These
methods can obtain convergence guarantees that are less affected by data
heterogeneity. However, {gradient-tracking methods} usually require each agent to communicate an
extra model-sized tracking variable, which increases the per-iteration
{communication cost}. {Other methods, such as QG-DSGDm}~\citep{DBLP:conf/icml/0004KSJ21} and RelaySum~\citep{vogels2021relaysum},
avoid this extra communication by changing the local update or {the communication scheme}. {QG-DSGDm preserves DSGD's per-iteration communication cost. RelaySum has the same communication cost as gossip averaging on the same spanning tree.}

Motivated by these limitations, we propose {a decentralized learning algorithm} that
combines two ideas. First, inspired by primal averaging and schedule-free
optimization~\citep{defazio2025smoothing,defazio2024road}, we maintain
separate sequences for optimization, gradient evaluation, and {output averaging}.
This {allows} {the output model} to move more smoothly than the noisy local
{optimization sequence}. Second, inspired by quasi-global
momentum~\citep{DBLP:conf/icml/0004KSJ21}, we build {a local quasi-global momentum buffer} from
consecutive mixed iterates. {This buffer reflects} the effect of network
mixing, but does not require agents to send an extra momentum or tracking
buffer. The resulting method, \emph{Decentralized Primal Averaging} (DPA), is
designed to improve training under heterogeneous data {using communication between neighbors}. Our main contributions are as follows.

\subsection{Contributions}
\begin{itemize}
    \item {We propose \emph{Decentralized Primal Averaging} (DPA),
    which combines primal averaging~\citep{defazio2024road,defazio2025smoothing}
    with local quasi-global momentum~\citep{DBLP:conf/icml/0004KSJ21} for decentralized
    learning with heterogeneous data. DPA maintains {separate optimization,
    gradient-evaluation, and output sequences}, and {uses a
    second communication round to mix the locally averaged models} while keeping the momentum
    buffer local.}

    \item {For smooth nonconvex objectives, we establish a convergence
    bound for the expected squared gradient norm at {DPA's network-average
    output model,} averaged over iterations. The bound explicitly accounts for
    network connectivity and data heterogeneity and identifies conditions
    for linear speedup in the number of agents.}

    \item {We introduce DPA-1G, a variant that uses one {communication round}
    per iteration, matching {DSGD's per-iteration communication cost}.
    It periodically replaces a stochastic-gradient update with gossip
    averaging of the output models.}

    \item {On CIFAR-10 and CIFAR-100 with ResNet-20 on a 16-agent
    ring ($\alpha=0.01$), DPA reaches high test accuracy in fewer epochs
    than QG-DSGDm and QG-GUTm, using a constant learning rate after linear
    warmup. At the full matched communication budget, both DPA variants
    outperform the strongest baseline, with DPA gains of $7.48$ and
    $16.03$ percentage points in mean test accuracy, respectively
    (Appendix~\ref{app:comm-matched}).}
\end{itemize}

\subsection{{Related work}}

Decentralized Stochastic Gradient Descent (DSGD)~\citep{lian2017can} is one
of the standard methods for peer-to-peer distributed learning. It alternates
between local stochastic-gradient steps and {gossip averaging steps}~\citep{xiao2004fast}, and has been widely used for training deep models
over decentralized networks~\citep{ying2021exponential,lian2017can}.
Stochastic Gradient Push (SGP)~\citep{assran2019stochastic} extends this idea
to directed and time-varying communication graphs. A unified analysis of
decentralized SGD {with time-varying topologies and local updates} was given
in~\citep{koloskova2020unified}. For nonconvex problems, NEXT~\citep{di2016next}
introduced a tracking mechanism for in-network optimization, which later
became a key idea behind gradient-tracking methods for decentralized
optimization~\citep{koloskova2021improved,pu2021distributed}. {AdGT~\citep{ghaderyan2026adgt} combines gradient tracking with adaptive per-agent
{learning rates} based on local smoothness and a common safety factor. It reduces tuning and often
converges faster than {gradient tracking with a} tuned {constant {learning rate}},
without extra communication rounds. The paper also provides a linear convergence guarantee.}

A common assumption in the analysis and use of DSGD is that the data are
independently and identically distributed (IID) across agents. This assumption
is often unrealistic in practical decentralized systems~\citep{hsieh2020non}.
In many applications, each agent observes data from a different distribution,
which is usually called the heterogeneous or non-IID setting. Data
heterogeneity makes the local objectives differ across agents, increases the
variation among local gradients, and can lead to unstable training and lower
test accuracy~\citep{DBLP:conf/icml/0004KSJ21}. Several methods have been proposed to
reduce this effect, including $D^2$~\citep{tang2018d} and gradient
tracking~\citep{nedic2017achieving,di2016next,koloskova2021improved}. These
methods correct local updates using extra information and can obtain
convergence guarantees that do not depend directly on the heterogeneity
parameter $\zeta^2${~\citep{koloskova2021improved}}. However,
gradient-tracking methods usually require each agent to communicate both the
model parameters and an additional tracking variable, which roughly doubles
the {{communication cost}} per iteration compared with vanilla DSGD.

Momentum-based methods provide another approach to decentralized training
with heterogeneous data. A direct momentum extension, DSGD with momentum
(DSGDm)~\citep{gao2020periodic}, can perform worse than DSGD when the data
are heterogeneous, because local momentum buffers can increase {model drift}.
Synchronizing the momentum buffers can improve performance
\citep{assran2019stochastic,yu2019linear}, and SlowMo~\citep{Wang2020SlowMo:}
uses a slow momentum update on globally averaged parameters. However, these
approaches require extra communication. QG-DSGDm~\citep{DBLP:conf/icml/0004KSJ21} avoids
this by using {quasi-global momentum}, which estimates a global update direction
from consecutive {mixed} iterates. RelaySum~\citep{vogels2021relaysum}
takes a different approach by changing {the communication scheme} itself, and
can be combined with other optimization methods. {Momentum
tracking (MT)}~\citep{takezawa2023momentum} combines momentum with gradient tracking
and {has} a convergence rate independent of data heterogeneity in the
nonconvex stochastic setting. QG-GUTm~\citep{aketi2023global} combines
{quasi-global momentum} with {global update tracking}, which tracks model updates
instead of gradients. This gives a tracking-like correction without extra
communication and improves over QG-DSGDm under {the same per-iteration
communication cost} on heterogeneous data.

\section{{Algorithm and setup}}
\label{sec:algorithm}

We consider a network of \(n\) agents connected by an undirected graph
\(\mathcal{G}=(\mathcal{N},\mathcal{E})\). An edge
\(\{i,j\}\in\mathcal{E}\) means that agents \(i\) and \(j\) can directly
exchange messages. We denote by \(\mathcal{N}(i)\) the neighbors of agent
\(i\), including \(i\) itself. Let \(W\in\mathbb{R}^{n\times n}\) be the
mixing matrix associated with the graph. We assume that \(W\) is symmetric and doubly stochastic
\citep{lian2017can,koloskova2020unified}. Communication is synchronous.

Each agent \(i\) has a local objective
\(f_i:\mathbb{R}^d\to\mathbb{R}\). The goal is to minimize the average
objective
\begin{equation}
\label{problem_1}
\min_{x\in\mathbb{R}^d}\; f(x)=\frac{1}{n}\sum_{i=1}^{n} f_i(x),
\qquad
f_i(x)=\mathbb{E}_{\xi_i\sim\mathcal{D}_i}\bigl[F_i(x;\xi_i)\bigr],
\end{equation}
where \(F_i\) is the loss on one data sample, \(\xi_i\) is sampled from the
local distribution \(\mathcal{D}_i\), and \(f_i\) is the expected local loss.
At each iteration, agent \(i\) can compute a stochastic gradient
\(\nabla F_i(x;\xi_i)\). We assume that these gradients are unbiased and have
bounded variance, as stated in Assumption~\ref{ass:bounded-variance}.

\subsection{Decentralized Primal Averaging}
\label{subsec:dpa}

We propose \emph{Decentralized Primal Averaging} (DPA), a decentralized
learning algorithm based on primal averaging. The use of separate sequences
for optimization, gradient evaluation, and {output averaging} follows ideas from
primal-averaging and schedule-free methods
\citep{lan2012optimal,defazio2024road}. The closest centralized method is
Generalized Primal Averaging (GPA)~\citep{defazio2025smoothing}, which
separates the point where gradients are evaluated from {the output
sequence} in {a centralized setting}. {ScheduleFree+~\citep{defazio2026schedulefree+}, released after our submission, extends centralized schedule-free learning to large language models using inner momentum, a modified Polyak {learning rate}, and warm-started averaging.}

DPA adapts {primal averaging} to decentralized non-IID learning. This
{introduces} two {challenges}. First, {the output models}
must stay close across agents. A purely local averaging step
{need not keep the output models close} when the data are heterogeneous.
We therefore {use a second communication round to mix the locally averaged model}
\[
s_i^k=(1-a_{k+1})x_i^k+a_{k+1}y_i^{k+1}.
\]

Second, {different local objectives can still lead agents to update
their models in different directions}. To reduce this
drift, DPA uses a local quasi-global momentum buffer
\(u_i^k\)~\citep{DBLP:conf/icml/0004KSJ21}. {Each agent updates this buffer
from the difference between its consecutive mixed iterates \(y_i^k\) and
\(y_i^{k+1}\), without sending the buffer to its neighbors.}

The {nonconvex} analysis uses a Lyapunov argument {to control}
the consensus errors of \(X_k\), \(Y_k\), and \(U_k\), {and to
{bound the expected squared gradient norm, averaged over iterations, at} \(\bar z_k\), the network average of the
local gradient-evaluation points}. {These consensus errors do not
arise in centralized primal averaging.}
\begin{algorithm}[t]
\caption{Decentralized Primal Averaging (DPA)}
\label{alg:dpa}
\textbf{Input:} Initial model $x_i^0$ {at agent} $i$,
$y_i^0=x_i^0$, $u_i^0=0$, number of iterations $T$, {learning rate} $\eta$, {momentum parameters}
$\beta_1,\beta_2\in(0,1)$, schedules $\{a_k\}_{k\ge1}$ and
$\{b_k\}_{k\ge1}$, mixing weights $\{w_{ij}\}_{j\in\mathcal N(i)}$,
and neighbor set $\mathcal N(i)$ including $i$.

\begin{algorithmic}[1]
\Procedure{{AGENT-}$i$}{}
  \For{$k=0,1,\dots,T-1$}
    \State $z_i^k=b_{k+1}y_i^k+(1-b_{k+1})x_i^k$
    \State Sample $\xi_i^k\sim\mathcal D_i$ and compute
    $g_i^k=\nabla F_i(z_i^k;\xi_i^k)$
    \State $v_i^k=y_i^k-\eta(\beta_1u_i^k+g_i^k)$
    \State \textsc{SendReceive}$(v_i^k)$
    \State $y_i^{k+1}=\sum_{j\in\mathcal N(i)}w_{ij}v_j^k$
    \State $d_i^{k+1}=(y_i^k-y_i^{k+1})/\eta$
    \State $u_i^{k+1}=(1-\beta_2)u_i^k+\beta_2 d_i^{k+1}$
    \State $s_i^k=(1-a_{k+1})x_i^k+a_{k+1}y_i^{k+1}$
    \State \textsc{SendReceive}$(s_i^k)$
    \State $x_i^{k+1}=\sum_{j\in\mathcal N(i)}w_{ij}s_j^k$
  \EndFor
  \State \Return $x_i^T$
\EndProcedure
\end{algorithmic}
\end{algorithm}
\paragraph{DPA updates}
We now describe the DPA update. {The} full procedure is shown in
Algorithm~\ref{alg:dpa}. At iteration \(k\), agent \(i\) first forms the
gradient-evaluation point
\[
z_i^k=b_{k+1}y_i^k+(1-b_{k+1})x_i^k .
\]
{The coefficient} \(b_{k+1}\) controls where the stochastic gradient is evaluated.
Larger values place \(z_i^k\) closer to the current optimization {iterate}
\(y_i^k\), while smaller values place it closer to {the output model}
\(x_i^k\). Agent \(i\) then computes
\[
g_i^k=\nabla F_i(z_i^k;\xi_i^k)
\]
and {takes a local update step to obtain}
\[
v_i^k = y_i^k-\eta(\beta_1u_i^k+g_i^k).
\]
{After one communication round}, the optimization sequence is updated as
\[
y_i^{k+1}=\sum_{j\in\mathcal N(i)}w_{ij}v_j^k .
\]
The local quasi-global momentum buffer is then updated by
\[
u_i^{k+1} = (1-\beta_2)u_i^k
+ \beta_2 \frac{y_i^k-y_i^{k+1}}{\eta}.
\]
This buffer is not communicated. It depends on the mixed iterate
\(y_i^{k+1}\), so it reflects the effect of the network mixing without sending
an extra tracking variable. Finally, agent \(i\) updates {the output sequence}. It first forms
\[
s_i^k=(1-a_{k+1})x_i^k+a_{k+1}y_i^{k+1},
\]
and then {uses a second communication round} to obtain \(x_i^{k+1}\). {The averaging coefficient}
\(a_{k+1}\) {controls the weights in the local averaging update.
Smaller values give greater weight to the previous output model}
\(x_i^k\){, while larger values give greater weight to the updated
optimization iterate} {\(y_i^{k+1}\)}.

\paragraph{{Communication cost}}
DPA uses two {communication rounds} per iteration. The first round mixes the
{locally updated model} \(v_i^k\), and the second round mixes the {locally averaged model} \(s_i^k\). Thus, each iteration communicates two model-sized vectors{ to each neighbor}.
The momentum buffer \(u_i^k\) is kept local and is never sent to neighboring
agents. In Section~\ref{subsec:DPA-1G-main}, we introduce DPA-1G, a variant
that uses only one {communication round} per iteration.

\paragraph{Compact form}
Stack the local variables as
\(X_k=[x_1^k,\ldots,x_n^k]\), and define \(Y_k,Z_k,U_k\) in the same way.
Let
\[
G_k=[\nabla F_1(z_1^k;\xi_1^k),\ldots,\nabla F_n(z_n^k;\xi_n^k)].
\]
Then DPA can be written as
\begin{align}
  Z_k     &= b_{k+1}Y_k+(1-b_{k+1})X_k, \label{eq:dpa-z} \\
  Y_{k+1} &= \bigl(Y_k-\eta(\beta_1U_k+G_k)\bigr)W, \label{eq:dpa-y} \\
  U_{k+1} &= (1-\beta_2)U_k+\beta_2\frac{Y_k-Y_{k+1}}{\eta}, \label{eq:dpa-u} \\
  X_{k+1} &= \bigl((1-a_{k+1})X_k+a_{k+1}Y_{k+1}\bigr)W, \label{eq:dpa-x}
\end{align}
with \(Y_0=X_0\) and \(U_0=0\). {Write $\bar x_k=n^{-1}X_k\one$, and analogously for the other sequences.}

\begin{remark}[Difference from centralized primal averaging]
DPA differs from centralized primal-averaging methods such as {schedule-free
optimization} and GPA in both the optimization setting and the proof technique.
Those methods {maintain a single set of iterates}, so they do not need
to control consensus error across agents. Their theoretical guarantees are mainly based on convex stochastic optimization and online-to-batch {or regret} arguments, whereas our analysis treats the decentralized nonconvex setting.
DPA runs on a sparse peer-to-peer graph with heterogeneous local data. The
algorithm must therefore control consensus error in the optimization sequence,
the gradient-evaluation sequence, {the output sequence}, and the local
quasi-global momentum buffer. This leads to a decentralized nonconvex
Lyapunov analysis that {controls} the consensus errors of \(X_k\), \(Y_k\),
\(Z_k\), and \(U_k\), rather than the centralized online-to-batch arguments
used in earlier primal-averaging work.
\end{remark}

\begin{remark}[Role of quasi-global momentum in DPA]
DPA also differs from quasi-global momentum methods such as QG-DSGDm and
QG-GUTm. In those methods, the quasi-global momentum buffer
{helps} reduce drift under heterogeneous data. In DPA,
{primal averaging separates the local optimization iterates and
the gradient-evaluation points from {the output model returned by each}
agent.} The quasi-global momentum buffer adds another correction on top
of this structure. Therefore, setting \(\beta_1=0\) removes the momentum
term but does not reduce DPA to vanilla DSGD; the primal-averaging
updates remain active.
\end{remark}

\paragraph{Effect of momentum and the \(b_k\) schedule}
Appendix~\ref{app:ablation-bk} studies the effect of quasi-global momentum
and the \(b_k\) schedule. In Figure~\ref{fig:dpa-ablation-bk}, we compare DPA
with several variants on CIFAR-10/ResNet-20 using a ring topology with
\(n=16\) agents and \(\alpha=0.01\). Full DPA, which uses both quasi-global
momentum and a decreasing \(b_k\) schedule, {has higher test accuracy
early in training and at {the last epoch}} than the {variants without momentum}.

Among the {variants without momentum}, the decreasing \(b_k\) schedule performs
slightly better than {the fixed choices of \(b_k\) tested here}.
{Training with} \(b_k=0.2\) is unstable in this experiment.
These results support using a {decreasing coefficient}
{in this setting}.

\subsection{{{Variant with one communication round per iteration}}}
\label{subsec:DPA-1G-main}

DPA uses two {communication rounds} per iteration{, with} one for the optimization sequence
and one for {the output sequence}. To match {the per-iteration
{communication cost of}} DSGD, QG-DSGDm, and QG-GUTm, we introduce DPA-1G,
which performs exactly one \textsc{SendReceive} operation per iteration.

On most iterations, agents {take a local stochastic-gradient step,
{apply gossip averaging to the resulting optimization iterates}}, update the quasi-global momentum
buffer, and update {the output sequence} locally. Every \(T_x\)
iterations, the method skips the stochastic-gradient computation, keeps the
optimization sequence and momentum buffer fixed, and uses {the communication round to mix} {the output sequence}. In our experiments, we set
\(T_x=5\). Thus, four out of every five iterations perform
stochastic-gradient updates, and {the remaining iteration is gradient-free and mixes the locally averaged
models across neighbors}. The full pseudocode is given in
Appendix~\ref{app:DPA-1G}. {The convergence analysis covers DPA,
not DPA-1G.}

\subsection{Assumptions}
\label{subsec:assumptions}

The convergence analysis uses a constant {learning rate} \(\eta>0\) { within each run}. The
coefficients \(a_{k+1}\) and \(b_{k+1}\) {are common to all agents and chosen before sampling. They} may change with \(k\), as long as
they stay within the {$T$-independent} bounds given in Appendix~\ref{app:notation}.

We use the following standard assumptions.

\begin{assumption}[Smoothness]
\label{ass:smoothness}
Each $f_i$ is $L$-smooth: for all $x,y \in \mathbb{R}^d$ and $i \in [n]$,
\[
\|\nabla f_i(x) - \nabla f_i(y)\| \le L\|x-y\|.
\]
\end{assumption}

\begin{assumption}[Unbiased gradients and bounded variance]
\label{ass:bounded-variance}
For every $i \in [n]$ and $x \in \mathbb{R}^d$,
\[
\E[\nabla F_i(x;\xi_i)] = \nabla f_i(x),
\qquad
\E\|\nabla F_i(x;\xi_i) - \nabla f_i(x)\|^2 \le \sigma^2.
\]
{Given the pre-sampling history $\mathcal F_k$, both {statements} hold and agents sample independently.}
\end{assumption}

\begin{assumption}[Bounded heterogeneity]
\label{ass:bounded-heterogeneity}
For all $x \in \mathbb{R}^d$,
\[
\frac{1}{n} \sum_{i=1}^{n} \|\nabla f_i(x) - \nabla f(x)\|^2 \le \zeta^2.
\]
\end{assumption}

\begin{assumption}[Mixing matrix]
\label{ass:mixing}
$W$ is {fixed,} symmetric and doubly stochastic, {with $w_{ij}=0$ for $j\notin\mathcal N(i)$}. For some $\rho \in (0,1)$,
\[
\max\{|\lambda_2(W)|, |\lambda_n(W)|\} \le 1-\rho.
\]
\end{assumption}

\begin{assumption}[Lower bound]
\label{ass:lower-bound}
The global objective satisfies $f(x) \ge f^\star > -\infty$ for all
$x \in \mathbb{R}^d$.
\end{assumption}

\begin{assumption}[Consensus initialization]
\label{ass:consensus-init}
All agents start from the same model: $X_0=Y_0=x_0\one^\top$ and $U_0=0$.
\end{assumption}

\section{{Convergence analysis}}
\label{sec:analysis}

We now give the main convergence results for DPA. Since stochastic gradients
are evaluated at \(Z_k\), we first bound {the expected squared gradient norm at} \(\bar
z_k\){, averaged over iterations}. We then transfer this bound to {the network-average output
model}
\(\bar x_k\). The transfer uses the relation
\[
\bar z_k-\bar x_k=b_{k+1}(\bar y_k-\bar x_k),
\]
which follows directly from the definition of \(Z_k\).

The proof uses two main ingredients. First, we introduce {an auxiliary} iterate
that removes {the network-average quasi-global momentum term, so that the network-average}
dynamics resemble a centralized stochastic-gradient update. Second, we use a
weighted Lyapunov function to control the consensus errors of \(X_k\),
\(Y_k\), and \(U_k\). Full proofs and explicit constants are given in
Appendix~\ref{app:proofs}.

\subsection{{Bound at the network average of the gradient-evaluation points}}
\label{subsec:z-guarantee}

The constants \(\tilde\eta\), \(\chi\), \(\lambda_X\), and \(\lambda_Y\)
used below are defined in Appendix~\ref{app:notation} and
Proposition~\ref{prop:explicit-weights}; {$C_\Psi$ is given in
\eqref{eq:CPsi-def}} { and is proportional to
$\eta^2(4\zeta^2+2\sigma^2)$}.

\begin{theorem}[{Bound on the squared gradient norm at \(\bar z_k\)}]
\label{thm:full-rate-z}
Suppose Assumptions~\ref{ass:smoothness}--\ref{ass:consensus-init}
hold, {with the coefficient conditions of
Section~\ref{subsec:assumptions} and $\tilde\eta\le1/(8L)$}, together
with {the condition on the momentum parameters} \eqref{eq:momentum-smallness-explicit},
the consensus {learning-rate} condition \eqref{eq:eta-smallness-explicit},
and the {learning-rate condition ensuring \(K_g>0\)} \eqref{eq:Kg-stepsize-sufficient}.
Then, for every $T \ge 1$,
{\small
\begin{equation}
\frac1T \sum_{k=0}^{T-1} \E\|\nabla f(\bar z_k)\|^2
\le
\frac{16(f(\bar x_0) - f^\star)}{\tilde\eta\, T}
+
\frac{64 L^2 C_\Psi}{\chi \min\{\lambda_X, \lambda_Y\}}
+
16\!\left(
\frac{9}{4 a_{\min}^2}
+ \frac{3 \beta_1^2}{4 \beta_2 (1 - \beta_1)}
+ \frac12
\right)\!\tilde\eta L \cdot \frac{\sigma^2}{n}.
\label{eq:full-rate-z}
\end{equation}}
In particular, with $\eta = c/\sqrt T$ and {$0<c \le c_0$} as in
Remark~\ref{rem:uniform-smallness},
\begin{equation}
\frac1T \sum_{k=0}^{T-1} \E\|\nabla f(\bar z_k)\|^2
\le
\frac{C_{z,1}}{\sqrt T} + \frac{C_{z,2}}{T},
\label{eq:full-rate-z-explicit}
\end{equation}
where $C_{z,1}, C_{z,2}$ depend only on the problem, algorithm, and
network parameters; explicit forms are given in
Appendix~\ref{app:combine-z}.
\end{theorem}

Choosing \(\eta = c\sqrt{n/T}\) gives {the following bound} for
{the network average of the gradient-evaluation points.}

\begin{corollary}[Linear-speedup regime]
\label{cor:linear-speedup-z}
Suppose {all assumptions} of Theorem~\ref{thm:full-rate-z}
{other than the learning-rate conditions hold}, with the {learning rate} chosen as
\[
\eta = c\sqrt{\frac{n}{T}},
\]
where $c > 0$ is independent of {$n,T$}. Suppose $T$ is large
enough that
\begin{equation}
\frac{c^2 n}{T}
\le
\min\!\left\{
\frac{(1-\beta_1)^2}{64L^2},\;
\frac{a_{\min}^2 \beta_2^2 (1-\beta_1)^4}{192 L^2},\;
\eta_{\mathrm{cons}}^2
\right\}.
\label{eq:c-condition-speedup}
\end{equation}
Then
\begin{equation}
\frac1T \sum_{k=0}^{T-1} \E\|\nabla f(\bar z_k)\|^2
\le
\frac{D_1}{\sqrt{nT}} + \frac{D_2 n}{T},
\label{eq:linear-speedup-z}
\end{equation}
where $D_1, D_2$ are given in Appendix~\ref{app:combine-z}{, and
$\eta_{\mathrm{cons}}^2$ is the right-hand side of
\eqref{eq:eta-smallness-explicit}}.
\end{corollary}

{The bound gives linear speedup when $D_1$ stays bounded as $n$
grows and $T$ is large enough that $D_2n/T \le D_1/\sqrt{nT}$.}
The corresponding statement for the {network-average} output
sequence $\bar x_k$ {is given in Corollary~\ref{cor:linear-speedup-x}
in Appendix~\ref{app:linear-speedup-x}.}

\subsection{{Guarantee for the network-average output model}}
\label{subsec:x-guarantee}

We now move from the intermediate bound on \(\bar z_k\) to the
{{network-average output model}} \(\bar x_k\). Since
\[
\bar z_k-\bar x_k=b_{k+1}(\bar y_k-\bar x_k),
\]
it is enough to control {the expected squared distance} between
\(\bar y_k\) and \(\bar x_k\){, averaged over iterations}.

\begin{theorem}[{Main convergence guarantee for the network-average output model}]
\label{thm:full-rate-x}
Suppose the assumptions of Theorem~\ref{thm:full-rate-z} hold. If
$\eta = c/\sqrt T$ with ${0<c} \le c_0$ as in
Remark~\ref{rem:uniform-smallness}, then
\begin{equation}
\frac1T \sum_{k=0}^{T-1} \E\|\nabla f(\bar x_k)\|^2
\le
\frac{C_{x,1}}{\sqrt T}
+ \frac{C_{x,2}}{T}
+ \frac{C_{x,3}}{T^{3/2}},
\label{eq:full-rate-x-explicit-T}
\end{equation}
where $C_{x,1}, C_{x,2}, C_{x,3}$ depend only on the problem, algorithm,
and network parameters. Explicit constants and the transfer inequality
from $\bar z_k$ to $\bar x_k$ are given in Appendix~\ref{app:thm-x-rate}.
\end{theorem}

\begin{remark}
\label{rem:speedup-x-transfer-main}
{Under the {assumptions} of
Corollary~\ref{cor:linear-speedup-z}, {with
$\eta=c\sqrt{n/T}$}, Appendix~\ref{app:linear-speedup-x} bounds
{the expected squared gradient norm at $\bar x_k$, averaged over
iterations,} by $\tfrac73(D_1/\sqrt{nT}+D_2n/T)$.}
Therefore, {the network-average output sequence} \(\bar x_k\) inherits the same
linear-speedup behavior. {The residual consensus-error term} in
\eqref{eq:linear-speedup-z} {is no larger than the leading term}
{for $D_1>0$} when
\[
T {\ge} \left(\frac{D_2}{D_1}\right)^2 n^3.
\]
For smaller \(T\), {the consensus-error term may} be the larger term.
{Uniform speedup {from this bound} requires $D_1$ to stay
bounded as $n$ grows. Appendix~\ref{app:local-output} extends the
iteration-averaged guarantee to {local output models}, not the last iterate.}
\end{remark}

\section{{Numerical experiments}}
\label{sec:experiments}

We evaluate DPA on decentralized image classification with heterogeneous
data. For fair and reproducible comparisons, we use the public GUT codebase
of \citet{aketi2023global}. We follow the GUT benchmark for datasets,
architectures, data partitions, {and communication topologies}.
We report results for both DPA and {DPA-1G, the variant with one communication round per iteration}.

\subsection{Setup}
\label{subsec:setup}

\paragraph{Datasets and architectures}
We use Fashion-MNIST with LeNet-5, CIFAR-10 with VGG-11, CIFAR-10 with
ResNet-20, and CIFAR-100 with ResNet-20. We use the same preprocessing and
data augmentation pipelines as \citet{aketi2023global}. Additional details
are given in Appendix~\ref{app:architectures}.

\paragraph{Heterogeneous data partitioning}
{We partition the training data among agents using a Dirichlet distribution} with concentration parameter
\(\alpha\). Smaller values of \(\alpha\) {tend to} produce more
skewed label distributions across agents and therefore stronger
heterogeneity. Following \citet{aketi2023global}, we use
\(\alpha\in\{0.1,0.01\}\) to study moderate and strong heterogeneity.

\paragraph{Communication topologies}
The main results use a ring topology with \(n=16\) agents. Additional results
for other graph topologies and network sizes are reported in
Appendix~\ref{app:extended-results}. All graphs use symmetric doubly
stochastic mixing matrices, consistent with Assumption~\ref{ass:mixing}.

\paragraph{Baselines}
We compare DPA and DPA-1G with {five baselines}:
DSGD~\citep{lian2017can}, GT~\citep{nedic2017achieving},
QG-DSGDm~\citep{DBLP:conf/icml/0004KSJ21}, MT~\citep{takezawa2023momentum}, and
QG-GUTm~\citep{aketi2023global}.
{For DSGD, QG-DSGDm, and QG-GUTm, we adopt the hyperparameters
from the public GUT codebase~\citep{aketi2023global},
which we also tested in our experimental settings.}
For DPA and DPA-1G, we tune the optimizer hyperparameters while retaining
the same data partitions, topology, and evaluation protocol as the baselines. {The learning rates and
method-specific hyperparameters are listed in
Appendix~\ref{app:hyperparams}.}

\paragraph{Baseline {learning-rate} schedule}
For DSGD, GT, QG-DSGDm, MT, and QG-GUTm, we follow the
{stepwise learning-rate decay} used in the GUT
benchmark~\citep{aketi2023global}: the learning rate is divided by \(10\)
after \(50\%\) and \(75\%\) of {each run's} training horizon.
DPA and DPA-1G do not use this decay schedule.
{Warmup did not improve baseline {test} accuracy in our
tests, {so we retain the baseline schedule without warmup}.}

\paragraph{{Linear warmup for DPA and DPA-1G}}
The theory uses a constant {learning rate} \(\eta\). In {DPA and DPA-1G}
experiments, we {increase the learning rate linearly} to \(\eta\) over \(T_w\) iterations
{and then keep the learning rate constant}:
\[
\eta_k=
\begin{cases}
\eta \dfrac{k+1}{T_w}, & k<T_w,\\
\eta, & k\ge T_w.
\end{cases}
\]
{When warmup is specified as a fraction of {a reference training
horizon}, we multiply the warmup duration in epochs by the number of
{mini-batches} per epoch to obtain \(T_w\).}
{In the equal-epoch non-IID runs, we} use a warmup over \(25\%\)
of training for \(\alpha=0.1\) and over \(50\%\) for \(\alpha=0.01\).

For DPA and DPA-1G, we use the exponential coefficient schedules described in
Appendix~\ref{app:schedule-details}. The {coefficient} schedule
parameters are listed in Appendix~\ref{app:hyperparams}.

\paragraph{Evaluation protocol}
{In the {equal-epoch} comparisons,} all methods are trained on
identical hardware {for} {100 epochs on Fashion-MNIST and
{200 on CIFAR-10 and CIFAR-100}}. {{Each agent evaluates} its local model on the full
test set.} For each seed, we first average test accuracy across
{{agents at the end of each epoch}}.
{Unless stated otherwise,} tables report both last-epoch accuracy
(LE) and the average accuracy over the last five epochs (L5E){,
computed within each seed}. We then report the mean and {standard deviation (std)}
across three random seeds { (using a divisor of three for the
variance)}. We use L5E as the main metric because it is less sensitive to
fluctuations at the end of training than LE.
{Boldface indicates the highest mean accuracy for each setting,
with LE and L5E compared separately.} {{Test accuracy curves}} show the
mean {test accuracy at each epoch}, with shaded regions
{indicating} \(\pm 1\) standard deviation across seeds.

\begin{remark}
\label{rem:gut-evaluation}
{After training, the GUT code} evaluates the models once more after
$K_{\mathrm{post}}=\min(51,\operatorname{len}(\texttt{train\_loader}))$
{communication rounds of gossip averaging with no gradient updates.
We refer to this phase as post-training averaging.} {We exclude this evaluation from
LE and L5E and do not count these rounds in the communication budget.
For DPA and DPA-1G, the final evaluation simply repeats the last
measurement and is also excluded. Appendix~\ref{app:extra-gossip}
reports the effect of {post-training averaging} separately.}
\end{remark}

\subsection{Accuracy under heterogeneous data}
\label{subsec:main-results}

\begin{table}[t]
\centering
\caption{Test accuracy (\%) {on a 16-agent ring}.}
\label{tab:ring-n16-main}

% Fashion-MNIST DPA/DPA-1G: LE and L5E use the supplied eta=0.1 three-seed summaries.
% CIFAR-10/VGG-11 DPA-1G at alpha=0.01: all three selected logs have tx5 (job17519818).

\resizebox{\linewidth}{!}{
\begin{tabular}{lcccc}
\toprule
Method
& {Fashion-MNIST / LeNet-5}
& CIFAR-10 / VGG-11
& CIFAR-10 / ResNet-20
& CIFAR-100 / ResNet-20 \\
\midrule
\multicolumn{5}{c}{$\alpha=0.1$} \\
\cmidrule(lr){2-5}
& LE $\pm$ std / L5E $\pm$ std
& LE $\pm$ std / L5E $\pm$ std
& LE $\pm$ std / L5E $\pm$ std
& LE $\pm$ std / L5E $\pm$ std \\
\midrule

DSGD
& \acc{{85.605}}{{2.962}}{{85.974}}{{2.374}}
& \acc{{65.99}}{{2.07}}{{65.97}}{{1.95}}
& \acc{57.71}{3.08}{57.49}{3.24}
& \acc{{25.53}}{{0.48}}{{26.19}}{{0.60}} \\

GT
& \acc{{84.517}}{{0.156}}{{84.515}}{{0.154}}
& \acc{{76.13}}{1.16}{76.12}{1.17}
& \acc{69.90}{2.73}{70.25}{2.33}
& \acc{{41.95}}{{1.35}}{{42.15}}{{1.47}} \\

QG-DSGDm
& \acc{{89.717}}{{0.230}}{{89.678}}{{0.273}}
& \acc{{75.57}}{{1.85}}{{75.43}}{{1.24}}
& \acc{81.73}{1.58}{81.79}{1.80}
& \acc{{44.89}}{{1.85}}{{44.64}}{{1.39}} \\

MT
& \acc{{90.064}}{{0.034}}{{90.048}}{{0.050}}
& \acc{{81.90}}{{1.81}}{{82.44}}{{1.18}}
& \acc{79.83}{1.64}{79.79}{0.96}
& \acc{{49.18}}{{3.43}}{{49.54}}{{3.87}} \\

QG-GUTm
& \acc{{89.667}}{{0.002}}{{89.671}}{{0.042}}
& \acc{{81.32}}{{3.02}}{{80.89}}{{3.71}}
& \acc{85.34}{1.42}{85.17}{1.30}
& \acc{{45.04}}{{0.73}}{{45.34}}{{1.11}} \\

DPA-1G
& \acc{89.33}{0.65}{{89.30}}{{0.60}}
& \acc{\textbf{83.62}}{\textbf{1.43}}{{\textbf{83.60}}}{{\textbf{1.46}}}
& \acc{86.58}{0.85}{86.08}{1.46}
& \acc{53.92}{0.55}{53.92}{{0.56}} \\

DPA
& \acc{{\textbf{90.08}}}{{\textbf{0.28}}}{{\textbf{90.13}}}{{\textbf{0.26}}}
& \acc{83.42}{1.56}{{83.46}}{{1.54}}
& \acc{\textbf{86.73}}{\textbf{1.10}}{\textbf{86.65}}{\textbf{1.21}}
& \acc{{\textbf{54.12}}}{{\textbf{0.97}}}{{\textbf{54.08}}}{{\textbf{0.92}}} \\

\midrule
\multicolumn{5}{c}{$\alpha=0.01$} \\
\cmidrule(lr){2-5}
& LE $\pm$ std / L5E $\pm$ std
& LE $\pm$ std / L5E $\pm$ std
& LE $\pm$ std / L5E $\pm$ std
& LE $\pm$ std / L5E $\pm$ std \\
\midrule

DSGD
& \acc{{73.930}}{{2.985}}{{74.310}}{{3.002}}
& \acc{{41.00}}{{3.44}}{{40.72}}{{3.72}}
& \acc{{29.79}}{{4.60}}{{30.26}}{{4.43}}
& \acc{{16.14}}{{0.82}}{{16.14}}{{0.48}} \\

GT
& \acc{{84.102}}{{0.226}}{{84.098}}{{0.227}}
& \acc{70.64}{1.77}{70.64}{1.77}
& \acc{{58.38}}{{0.81}}{{58.29}}{{0.95}}
& \acc{{34.20}}{{1.85}}{{34.40}}{{1.88}} \\

QG-DSGDm
& \acc{{76.456}}{{3.624}}{{76.977}}{{3.128}}
& \acc{{62.92}}{{2.47}}{{63.12}}{{1.18}}
& \acc{{61.47}}{{1.93}}{{60.27}}{{1.44}}
& \acc{{33.57}}{{0.80}}{{33.14}}{{1.80}} \\

MT
& \acc{{82.090}}{{4.793}}{{83.381}}{{3.476}}
& \acc{{65.97}}{{3.92}}{{65.26}}{{3.59}}
& \acc{{48.20}}{{2.49}}{{48.22}}{{0.50}}
& \acc{{30.09}}{{2.63}}{{31.09}}{{3.22}} \\

QG-GUTm
& \acc{{76.360}}{{5.199}}{{76.513}}{{5.839}}
& \acc{{75.49}}{{1.49}}{{75.71}}{{1.40}}
& \acc{{66.47}}{{1.74}}{{65.88}}{{2.57}}
& \acc{{35.54}}{{1.01}}{{35.77}}{{1.33}} \\

DPA-1G
& \acc{{\textbf{85.60}}}{{\textbf{2.46}}}{{\textbf{85.53}}}{{\textbf{2.48}}}
& \acc{{75.73}}{{2.73}}{{75.67}}{{2.86}}
& \acc{78.57}{4.89}{{78.50}}{4.94}
& \acc{48.77}{1.16}{{48.75}}{{1.12}} \\

DPA
& \acc{84.73}{3.97}{{83.15}}{{5.80}}
& \acc{{\textbf{77.06}}}{{\textbf{2.49}}}{{\textbf{76.84}}}{{\textbf{2.54}}}
& \acc{\textbf{82.01}}{\textbf{1.47}}{{\textbf{81.95}}}{{\textbf{1.51}}}
& \acc{\textbf{51.97}}{\textbf{1.18}}{{\textbf{51.86}}}{{\textbf{1.17}}} \\

\bottomrule
\end{tabular}}
\end{table}

Table~\ref{tab:ring-n16-main} reports results on the ring topology with
\(n=16\) agents. Under moderate heterogeneity (\(\alpha=0.1\)),
{DPA has the highest mean L5E accuracy} on CIFAR-10/ResNet-20
(\(86.65\%\)) and CIFAR-100/ResNet-20 (\({54.08}\%\)), while DPA-1G
{has the highest mean L5E accuracy} on CIFAR-10/VGG-11
(\({83.60}\%\)). On Fashion-MNIST/LeNet-5, {DPA
{and MT have similar mean L5E accuracies} {(\({90.13}\%\) and
\(90.05\%\), respectively)}.}

Under stronger heterogeneity (\(\alpha=0.01\)), {DPA's largest gains
are on the two ResNet-20 tasks}. DPA {has the highest mean LE and
L5E} on CIFAR-10/ResNet-20 and CIFAR-100/ResNet-20, improving over QG-GUTm
by {\(16.07\) and \(16.09\)} percentage points in L5E, respectively.
On CIFAR-10/VGG-11, {DPA {also has the highest mean LE and
L5E, but its L5E gain over QG-GUTm is smaller (\(1.13\) {percentage points)}}.}
{With \(T_x=5\), DPA-1G has a mean L5E accuracy of \(75.67\%\).}
On Fashion-MNIST/LeNet-5, DPA-1G {has the highest mean LE and L5E},
{improving over {GT, the strongest baseline,} by \({1.43}\)
percentage points in L5E.}

Additional results for \(n=32\) agents and other graph topologies are given
in Appendix~\ref{app:extended-results}. Figure~\ref{fig:ring16-alpha001-trajectories}
shows the {$n=16$ ring} {test accuracy curves} for \(\alpha=0.01\).
{Among the methods that use one {communication round} per iteration, DPA-1G
has the highest mean {LE accuracy} on CIFAR-100/ResNet-20.}

\begin{remark}
\label{rem:communication-efficiency}
DPA uses two {communication rounds} per iteration, so {DPA sends more data per iteration}
than DSGD-style {methods that use one communication round per iteration}. However,
Figure~\ref{fig:ring16-alpha001-trajectories} shows that on the heterogeneous
ring with \(\alpha=0.01\), DPA reaches high test accuracy in fewer epochs
than QG-DSGDm and QG-GUTm. {At the full matched communication
budget, DPA's mean test accuracy exceeds that of the strongest baseline
by \(7.48\) percentage points on CIFAR-10 and \(16.03\) {percentage points on CIFAR-100}}
{(Appendix~\ref{app:comm-matched}).}
\end{remark}

\begin{figure}[t]
\centering
\begin{subfigure}[t]{0.44\linewidth}
    \centering
    \includegraphics[width=\linewidth]{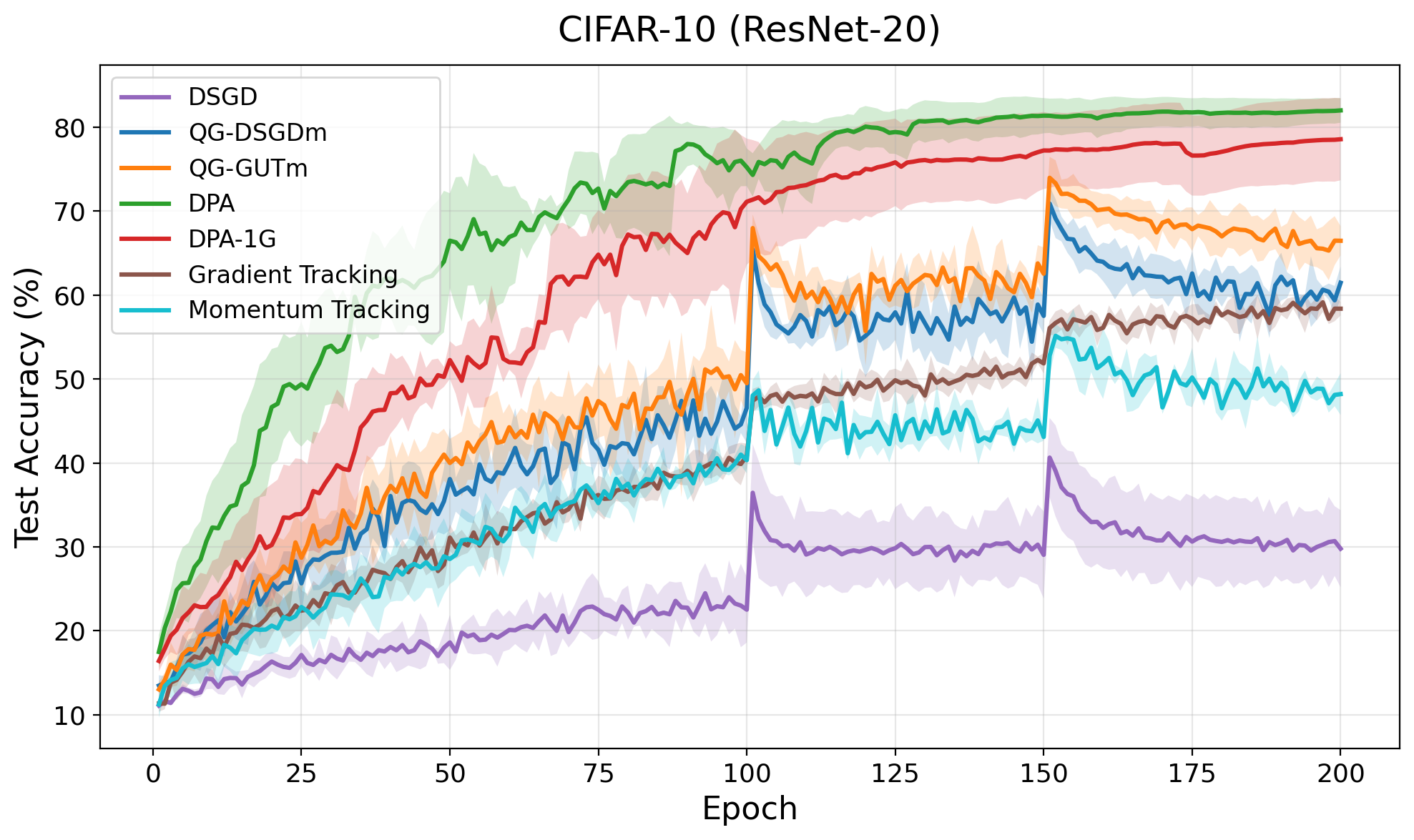}
    \caption{CIFAR-10 / ResNet-20}
    \label{fig:ring16-a001-cifar10-resnet20}
\end{subfigure}
\hfill
\begin{subfigure}[t]{0.44\linewidth}
    \centering
    \includegraphics[width=\linewidth]{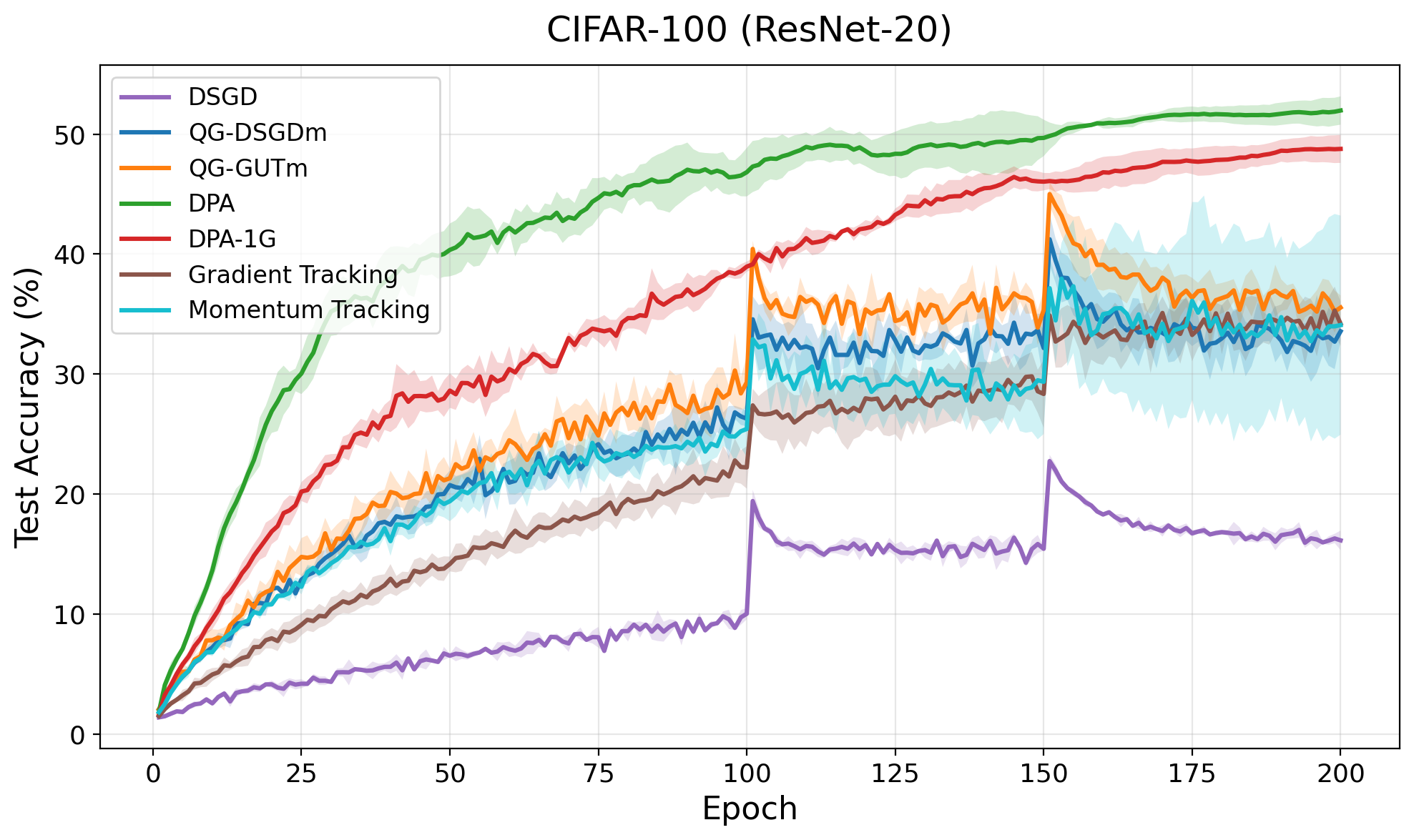}
    \caption{CIFAR-100 / ResNet-20}
    \label{fig:ring16-a001-cifar100-resnet20}
\end{subfigure}
\caption{{Test accuracy during training on a 16-agent ring
($\alpha=0.01$).}}
\label{fig:ring16-alpha001-trajectories}
\end{figure}

{See Appendices~\ref{app:iid-results} and~\ref{app:system-cost} for IID and timing results.}

\section{Conclusion}
\label{sec:conclusion}

We introduced \emph{Decentralized Primal Averaging} (DPA), {a decentralized
learning algorithm} for heterogeneous data. DPA adapts primal averaging to the
gossip-based setting by using three coupled sequences for optimization,
gradient evaluation, and {output averaging}. It also uses a local quasi-global
momentum buffer built from consecutive mixed iterates, without sending an
extra tracking variable. {For smooth nonconvex objectives, we bounded
the expected squared gradient norm at the {network-average output model,} averaged
over iterations. The bound gives linear speedup in the number of agents
when the network- and heterogeneity-dependent term is no larger than the
leading term.} We also introduced DPA-1G, a {variant with one communication round per iteration}
that matches {the} {{per-iteration}} {communication cost} of DSGD-style
baselines. Experiments on decentralized image classification show that DPA
and DPA-1G improve over strong baselines, with the largest gains in highly
heterogeneous settings on sparse ring topologies. Future work includes
extending the analysis to DPA-1G, relaxing the conservative
{{condition on the momentum parameters}}, and adapting primal averaging to time-varying
or asynchronous networks.

\begin{ack}
This work was partially supported by the Research Council of Finland
(Grant 354523). We acknowledge the computational resources provided
by the Aalto Science-IT project.
\end{ack}
% Keep references and appendices after the main paper.
\label{LastMainPage}
\clearpage
\bibliographystyle{plainnat}    
\bibliography{references}

%%%%%%%%%%%%%%%%%%%%%%%%%%%%%%%%%%%%%%%%%%%%%%%%%%%%%%%%%%%%

\appendix

\section{Algorithm details and additional notation}
\label{app:schedules}

The problem statement, algorithm, and standing assumptions are in
Section~\ref{sec:algorithm}. This appendix gives the notation, coefficient bounds, schedule definitions,
and technical conditions used in the proofs.

\subsection{Coefficient bounds and notation}
\label{app:notation}

Throughout the analysis, we use a constant {learning rate} $\eta > 0$. The
coefficient schedules $a_{k+1}$ and $b_{k+1}$ may be time-varying,
provided they satisfy
\[
0 < b_{\min} \le b_{k+1} \le b_{\max} < 1,
\qquad
0 < a_{\min} \le a_{k+1} \le a_{\max} < 1.
\]
The momentum parameters satisfy $\beta_1,\beta_2 \in (0,1)$, and we
require
\begin{equation}
\tilde\eta \coloneqq \frac{\eta}{1-\beta_1} \le \frac{1}{8L},
\label{eq:stepsize-smallness-basic}
\end{equation}
together with the additional {learning-rate} conditions {stated below} when needed.

Let
\[
J \coloneqq \frac{1}{n}\one\one^\top .
\]
For any $A \in \mathbb{R}^{d \times n}$, define
\[
\bar a \coloneqq \frac{1}{n}A\one \in \mathbb{R}^d,
\qquad
\bar A \coloneqq \bar a\one^\top = AJ .
\]
Thus $\bar X_k=\bar x_k\one^\top$, and similarly for
$Y_k,Z_k,U_k,G_k$. {We write $\E_k[\cdot]=\E[\cdot\mid\mathcal F_k]$.}

\begin{remark}
\label{rem:spectral-W}
Under Assumption~\ref{ass:mixing}, the mixing matrix satisfies
$\norm{W}_2 \le 1$ and $\norm{I-W}_2 \le 2$.
\end{remark}

\begin{remark}
\label{rem:frob-contraction}
For every $A \in \mathbb{R}^{d \times n}$,
\[
\norm{AW-\bar A}_F \le (1-\rho)\norm{A-\bar A}_F .
\]
\end{remark}

\subsection{Schedules for $a_{k+1}$ and $b_{k+1}$}
\label{app:schedule-details}

For each coefficient $c \in \{a,b\}$, we use either a linear or an
exponential schedule:
\[
c_{k+1}
=
c_{\mathrm{start}}
+
(c_{\mathrm{end}}-c_{\mathrm{start}})
\min\!\left(1,\frac{k+1}{T_c}\right)
\qquad\text{(linear)},
\]
or
\[
c_{k+1}
=
c_{\mathrm{start}}
\left(\frac{c_{\mathrm{end}}}{c_{\mathrm{start}}}\right)^{
\left(\min\!\left(1,\frac{k+1}{T_c}\right)\right)^{q_c}}
\qquad\text{(exponential)}.
\]
Here $T_c>0$ is the schedule length and $q_c>0$ controls the curvature
of the exponential schedule. The start and end values are chosen so that the bounds in
Appendix~\ref{app:notation} hold at every iteration.

\section{Proof of Theorem~\ref{thm:full-rate-z}}
\label{app:proofs}

{Throughout this proof, we use the assumptions of Section~\ref{subsec:assumptions} and the coefficient bounds in Appendix~\ref{app:notation}.}

% --- Step 1: {auxiliary} iterate and descent inequality ---

\subsection{Step 1: {auxiliary} iterate and descent}
\label{app:step1}
Since \(W\) is doubly stochastic, gossip averaging {preserves the network average.} The {auxiliary} iterate \(\hat x_k\) is a shifted version of
\(\bar y_k\), chosen so that its update is driven only by {the network-average
stochastic gradient} \(\bar g_k\).
\begin{lemma}
\label{lem:averaged-recursions}
Let
\[
\alpha_u \coloneqq 1-\beta_2(1-\beta_1).
\]
Then, for every \(k\ge 0\),
\begin{align}
\bar y_{k+1}
&=
\bar y_k-\eta(\beta_1\bar u_k+\bar g_k),
\label{eq:ybar-rec}
\\
\bar u_{k+1}
&=
\alpha_u \bar u_k+\beta_2\bar g_k,
\label{eq:ubar-rec}
\\
\bar x_{k+1}
&=
(1-a_{k+1})\bar x_k+a_{k+1}\bar y_{k+1}.
\label{eq:xbar-rec}
\end{align}
\end{lemma}
\begin{proof}
Since \(W\one=\one\), we have
\(AW\tfrac1n\one=A\tfrac1n\one\) for any \(A\in\mathbb R^{d\times n}\).
Averaging the \(Y\)-update gives
\[
\bar y_{k+1}
=
\bar y_k-\eta(\beta_1\bar u_k+\bar g_k),
\]
which proves \eqref{eq:ybar-rec}. Similarly, averaging the \(X\)-update gives
\[
\bar x_{k+1}
=
(1-a_{k+1})\bar x_k+a_{k+1}\bar y_{k+1},
\]
which proves \eqref{eq:xbar-rec}. Moreover, {writing \(\bar d_{k+1}=n^{-1}\sum_{i=1}^n d_i^{k+1}\),}
\eqref{eq:ybar-rec} {gives}
\[
{\bar d_{k+1}}
=
\frac{\bar y_k-\bar y_{k+1}}{\eta}
=
\beta_1\bar u_k+\bar g_k .
\]
Averaging the \(U\)-update then yields
\[
\bar u_{k+1}
=
(1-\beta_2)\bar u_k+\beta_2 {\bar d_{k+1}}
=
\bigl(1-\beta_2(1-\beta_1)\bigr)\bar u_k+\beta_2\bar g_k
=
\alpha_u\bar u_k+\beta_2\bar g_k .
\]
This proves \eqref{eq:ubar-rec} and completes the proof.
\end{proof}

\subsubsection{Second-moment bounds}
\label{app:second-moments}
\begin{lemma}
\label{lem:second-moments}
For every \(k\ge 0\),
\begin{align}
\E_k\norm{\bar g_k}^2
&\le
2\norm{\nabla f(\bar z_k)}^2
+
2L^2\left(\frac1n\norm{Z_k-\bar Z_k}_F^2\right)
+
\frac{\sigma^2}{n},
\label{eq:gbar-second-moment}
\\
\E_k\norm{\beta_1\bar u_k+\bar g_k}^2
&\le
2\beta_1^2\norm{\bar u_k}^2
+
4\norm{\nabla f(\bar z_k)}^2
+
4L^2\left(\frac1n\norm{Z_k-\bar Z_k}_F^2\right)
+
\frac{2\sigma^2}{n}.
\label{eq:beta1u-plus-g}
\end{align}
\end{lemma}
\begin{proof}
Since
\[
\E_k[\bar g_k]
=
\frac1n\sum_{i=1}^n\nabla f_i({z_i^k}),
\]
the bias--variance decomposition gives
\begin{align*}
\E_k\norm{\bar g_k}^2
&=
\E_k\norm{
\bar g_k-\frac1n\sum_{i=1}^n\nabla f_i({z_i^k})
}^2
+
\norm{\frac1n\sum_{i=1}^n\nabla f_i({z_i^k})}^2
\\
&\le
\frac{\sigma^2}{n}
+
\norm{\frac1n\sum_{i=1}^n\nabla f_i({z_i^k})}^2
\\
&\le
\frac{\sigma^2}{n}
+
2\norm{\nabla f(\bar z_k)}^2
+
2\norm{
\frac1n\sum_{i=1}^n
\left(\nabla f_i({z_i^k})-\nabla f_i(\bar z_k)\right)
}^2
\\
&\le
\frac{\sigma^2}{n}
+
2\norm{\nabla f(\bar z_k)}^2
+
2L^2\left(\frac1n\norm{Z_k-\bar Z_k}_F^2\right).
\end{align*}
This proves \eqref{eq:gbar-second-moment}. The first inequality uses
Assumption~\ref{ass:bounded-variance} and {conditional independence across agents given \(\mathcal F_k\)}; the
second uses \(\|a+b\|^2\le 2\|a\|^2+2\|b\|^2\); the last uses Jensen's
inequality and Assumption~\ref{ass:smoothness}.

For \eqref{eq:beta1u-plus-g}, applying
\(\norm{a+b}^2\le 2\norm{a}^2+2\norm{b}^2\) and
\eqref{eq:gbar-second-moment} gives
\begin{align*}
\E_k\norm{\beta_1\bar u_k+\bar g_k}^2
&\le
2\beta_1^2\norm{\bar u_k}^2
+
2\E_k\norm{\bar g_k}^2
\\
&\le
2\beta_1^2\norm{\bar u_k}^2
+
4\norm{\nabla f(\bar z_k)}^2
+
4L^2\left(\frac1n\norm{Z_k-\bar Z_k}_F^2\right)
+
\frac{2\sigma^2}{n}.
\end{align*}
This proves \eqref{eq:beta1u-plus-g}.
\end{proof}

\subsubsection{Definition of the {auxiliary} iterate}
\label{app:virtual-iterate}
We now define the {auxiliary} iterate and verify the cancellation of {the network-average
momentum term}.

\begin{lemma}
\label{lem:virtual-iterate}
Let
\[
\tilde\eta \coloneqq \frac{\eta}{1-\beta_1},
\qquad
c_m \coloneqq \frac{\eta\beta_1}{\beta_2(1-\beta_1)},
\qquad
\hat x_k \coloneqq \bar y_k-c_m\bar u_k .
\]
Then \(\hat x_0=\bar x_0\), and for every \(k\ge 0\),
\begin{equation}
\hat x_{k+1}
=
\hat x_k-\tilde\eta\,\bar g_k .
\label{eq:virtual-recursion}
\end{equation}
\end{lemma}
\begin{proof}
By Lemma~\ref{lem:averaged-recursions},
\begin{align*}
\hat x_{k+1}
&=
\bar y_{k+1}-c_m\bar u_{k+1}
\\
&=
\bar y_k-\eta(\beta_1\bar u_k+\bar g_k)
-c_m(\alpha_u\bar u_k+\beta_2\bar g_k)
\\
&=
\hat x_k
+\bigl(c_m(1-\alpha_u)-\eta\beta_1\bigr)\bar u_k
-(\eta+c_m\beta_2)\bar g_k .
\end{align*}
Since \(1-\alpha_u=\beta_2(1-\beta_1)\), the definition of \(c_m\) gives
\[
c_m(1-\alpha_u)=\eta\beta_1,
\qquad
\eta+c_m\beta_2=\tilde\eta .
\]
Thus \eqref{eq:virtual-recursion} follows. Finally,
\[
\hat x_0=\bar y_0-c_m\bar u_0=\bar x_0,
\]
because \(Y_0=X_0\) and \(U_0=0\).
\end{proof}

The update in \eqref{eq:virtual-recursion} has the form of a centralized
stochastic-gradient update with {learning rate} \(\tilde\eta\). Using the smoothness
of \(f\), we obtain the following one-step bound. {The bound accounts for the gap between \(\hat x_k\) and \(\bar z_k\), {the consensus error of the local gradient-evaluation points}, and gradient noise.}

\begin{lemma}
\label{lem:virtual-descent}
Under \eqref{eq:stepsize-smallness-basic}, for every \(k\ge 0\),
\begin{align}
\E_k f(\hat x_{k+1})
\le\;&
f(\hat x_k)
-
\frac{\tilde\eta}{8}\norm{\nabla f(\bar z_k)}^2
+
3\tilde\eta L^2(1-b_{\min})^2\,\|\bar y_k-\bar x_k\|^2
\notag\\
&\quad+
3\tilde\eta L^2 \left(\frac{\eta\beta_1}{\beta_2(1-\beta_1)}\right)^2\,\norm{\bar u_k}^2
\notag\\
&\quad
+
\left(\tilde\eta L^2 + L^3\tilde\eta^2\right)\left(\frac1n \norm{Z_k-\bar Z_k}_F^2\right)
+
\frac{L\tilde\eta^2}{2}\cdot\frac{\sigma^2}{n}.
\label{eq:virtual-descent}
\end{align}
\end{lemma}

\begin{proof}
By Lemma~\ref{lem:virtual-iterate}, \(\hat x_{k+1}=\hat x_k-\tilde\eta\,\bar g_k\). Therefore,
\begin{align}
\E_k f(\hat x_{k+1})
\overset{(a)}{\le}\;&
f(\hat x_k)
-
\tilde\eta
\left\langle
\nabla f(\hat x_k),
\frac1n\sum_{i=1}^n\nabla f_i({z_i^k})
\right\rangle
+
\frac{L\tilde\eta^2}{2}\E_k\norm{\bar g_k}^2
\notag\\
\overset{(b)}{\le}\;&
f(\hat x_k)
-
\frac{\tilde\eta}{2}\norm{\nabla f(\hat x_k)}^2
+
\frac{\tilde\eta}{2}
\norm{
\frac1n\sum_{i=1}^n\nabla f_i({z_i^k})
-
\nabla f(\hat x_k)
}^2
+
\frac{L\tilde\eta^2}{2}\E_k\norm{\bar g_k}^2
\notag\\
\overset{(c)}{\le}\;&
f(\hat x_k)
-
\frac{\tilde\eta}{4}\norm{\nabla f(\bar z_k)}^2
+
\frac{\tilde\eta}{2}\norm{\nabla f(\hat x_k)-\nabla f(\bar z_k)}^2
\notag\\
&\quad
+
\frac{\tilde\eta}{2}
\norm{
\frac1n\sum_{i=1}^n\nabla f_i({z_i^k})
-
\nabla f(\hat x_k)
}^2
+
\frac{L\tilde\eta^2}{2}\E_k\norm{\bar g_k}^2
\notag\\
\overset{(d)}{\le}\;&
f(\hat x_k)
-
\frac{\tilde\eta}{4}\norm{\nabla f(\bar z_k)}^2
+
\frac{3\tilde\eta}{2}\norm{\nabla f(\hat x_k)-\nabla f(\bar z_k)}^2
\notag\\
&\quad
+
\tilde\eta
\norm{
\frac1n\sum_{i=1}^n\nabla f_i({z_i^k})
-
\nabla f(\bar z_k)
}^2
+
\frac{L\tilde\eta^2}{2}\E_k\norm{\bar g_k}^2
\notag\\
\overset{(e)}{\le}\;&
f(\hat x_k)
-
\frac{\tilde\eta}{4}\norm{\nabla f(\bar z_k)}^2
+
\frac{3\tilde\eta L^2}{2}\norm{\hat x_k-\bar z_k}^2
\notag\\
&\quad+
\tilde\eta L^2\left(\frac1n\norm{Z_k-\bar Z_k}_F^2\right)
+
\frac{L\tilde\eta^2}{2}\E_k\norm{\bar g_k}^2,
\label{eq:virtual-proof-main-chain}
\end{align}
where \((a)\) uses Assumption~\ref{ass:smoothness} and
\(\E_k[\bar g_k]=\tfrac1n\sum_{i=1}^n\nabla f_i({z_i^k})\); \((b)\) uses
\(-\langle a,b\rangle\le -\tfrac12\|a\|^2+\tfrac12\|a-b\|^2\); \((c)\) uses
\(-\tfrac12\|a\|^2\le -\tfrac14\|b\|^2+\tfrac12\|a-b\|^2\); \((d)\) uses
\(\|p-r\|^2\le 2\|p-q\|^2+2\|q-r\|^2\); and \((e)\) uses
Assumption~\ref{ass:smoothness} together with the {bound derived using Jensen's inequality in the proof of}
Lemma~\ref{lem:second-moments}.

By definition of \(\hat x_k\) and \(\bar z_k\),
\[
\hat x_k-\bar z_k
=
(1-b_{k+1})(\bar y_k-\bar x_k)-\frac{\eta\beta_1}{\beta_2(1-\beta_1)}\bar u_k,
\]
so \(\norm{a+b}^2\le 2\norm{a}^2+2\norm{b}^2\) and \(b_{k+1}\ge b_{\min}\) give
\begin{equation}
\norm{\hat x_k-\bar z_k}^2
\le
2(1-b_{\min})^2\|\bar y_k-\bar x_k\|^2
+
2\left(\frac{\eta\beta_1}{\beta_2(1-\beta_1)}\right)^2\norm{\bar u_k}^2.
\label{eq:virtual-proof-xhat-z}
\end{equation}
Substituting \eqref{eq:virtual-proof-xhat-z} into \eqref{eq:virtual-proof-main-chain} and applying \eqref{eq:gbar-second-moment} to the \(\E_k\norm{\bar g_k}^2\) term yields
\begin{align*}
\E_k f(\hat x_{k+1})
\le\;&
f(\hat x_k)
-
\left(\frac{\tilde\eta}{4}-L\tilde\eta^2\right)\norm{\nabla f(\bar z_k)}^2
\\
&\quad
+
3\tilde\eta L^2(1-b_{\min})^2\|\bar y_k-\bar x_k\|^2
+
3\tilde\eta L^2 \left(\frac{\eta\beta_1}{\beta_2(1-\beta_1)}\right)^2\norm{\bar u_k}^2
\\
&\quad
+
\left(\tilde\eta L^2+L^3\tilde\eta^2\right)\left(\frac1n\norm{Z_k-\bar Z_k}_F^2\right)
+
\frac{L\tilde\eta^2}{2}\cdot\frac{\sigma^2}{n}.
\end{align*}
Since \(\tilde\eta\le 1/(8L)\), \(\tilde\eta/4-L\tilde\eta^2\ge \tilde\eta/8\), proving \eqref{eq:virtual-descent}.
\end{proof}

The one-step bound in Lemma~\ref{lem:virtual-descent} contains
\(\|\bar y_k-\bar x_k\|^2\) and \(\|\bar u_k\|^2\). The next two lemmas {give recursions for these quantities}. {Proposition~\ref{prop:reduction} then bounds {the expected squared gradient norm, averaged over iterations,} in terms of the consensus error of \(Z_k\), the initial objective gap, and the gradient variance}.

\begin{lemma}
\label{lem:Gamma-coarse}
For every \(k\ge 0\),
\begin{align}
\E_k\|\bar y_{k+1}-\bar x_{k+1}\|^2
\le\;&
(1-a_{\min}) \|\bar y_k-\bar x_k\|^2
+
\frac{(1-a_{\min})^2}{a_{\min}} \eta^2
\Biggl(
2\beta_1^2\norm{\bar u_k}^2
+
4\norm{\nabla f(\bar z_k)}^2
\notag\\
&\qquad\qquad\qquad\qquad
+
4L^2\left(\frac1n \norm{Z_k-\bar Z_k}_F^2\right)
+
\frac{2\sigma^2}{n}
\Biggr).
\label{eq:Gamma-coarse}
\end{align}
\end{lemma}
\begin{proof}
By Lemma~\ref{lem:averaged-recursions},
\[
\bar y_{k+1}-\bar x_{k+1}
=
(1-a_{k+1})
\bigl((\bar y_k-\bar x_k)-\eta(\beta_1\bar u_k+\bar g_k)\bigr).
\]
Young's inequality with \(\tau=a_{k+1}/(1-a_{k+1})\) gives
\[
\|\bar y_{k+1}-\bar x_{k+1}\|^2
\le
(1-a_{k+1})\|\bar y_k-\bar x_k\|^2
+
\frac{(1-a_{k+1})^2}{a_{k+1}}\eta^2
\norm{\beta_1\bar u_k+\bar g_k}^2 .
\]
Since \(a_{k+1}\ge a_{\min}>0\) and \(t\mapsto (1-t)^2/t\) is decreasing on \((0,1)\),
\[
1-a_{k+1}\le 1-a_{\min},
\qquad
\frac{(1-a_{k+1})^2}{a_{k+1}}
\le
\frac{(1-a_{\min})^2}{a_{\min}}.
\]
Taking \(\E_k[\cdot]\) and applying Lemma~\ref{lem:second-moments} gives
\eqref{eq:Gamma-coarse}.
\end{proof}

\begin{lemma}
\label{lem:u-coarse}
For every \(k\ge 0\),
\begin{align}
\E_k\norm{\bar u_{k+1}}^2
\le\;&
\frac{1+\alpha_u^2}{2} \norm{\bar u_k}^2
+
2\beta_2^2
\left(
1+
\frac{2\alpha_u^2}
{1-\alpha_u^2}
\right)
\norm{\nabla f(\bar z_k)}^2
\notag\\
&\quad
+
2L^2\beta_2^2
\left(
1+
\frac{2\alpha_u^2}
{1-\alpha_u^2}
\right)
\left(\frac1n \norm{Z_k-\bar Z_k}_F^2\right)
+
\beta_2^2\frac{\sigma^2}{n}.
\label{eq:u-coarse}
\end{align}
\end{lemma}
\begin{proof}
By Lemma~\ref{lem:averaged-recursions} and
\(\E_k[\bar g_k]=\tfrac1n\sum_{i=1}^n\nabla f_i({z_i^k})\),
\begin{align}
\E_k\norm{\bar u_{k+1}}^2
&=
\alpha_u^2\norm{\bar u_k}^2
+
2\alpha_u\beta_2
\left\langle
\bar u_k,\tfrac1n\sum_{i=1}^n \nabla f_i({z_i^k})
\right\rangle
+
\beta_2^2\E_k\norm{\bar g_k}^2
\notag\\
&\le
\frac{1+\alpha_u^2}{2}\norm{\bar u_k}^2
+
\frac{2\alpha_u^2\beta_2^2}{1-\alpha_u^2}
\norm{\tfrac1n\sum_{i=1}^n \nabla f_i({z_i^k})}^2
+
\beta_2^2\E_k\norm{\bar g_k}^2,
\label{eq:u-coarse-proof-1}
\end{align}
where {we apply} Young's inequality with parameter
\((1-\alpha_u^2)/2\) to the cross term. As in the proof of
Lemma~\ref{lem:second-moments},
\[
\norm{\tfrac1n\sum_{i=1}^n \nabla f_i({z_i^k})}^2
\le
2\norm{\nabla f(\bar z_k)}^2
+
2L^2\left(\tfrac1n\norm{Z_k-\bar Z_k}_F^2\right).
\]
Combining this bound with \eqref{eq:gbar-second-moment} in
\eqref{eq:u-coarse-proof-1} gives \eqref{eq:u-coarse}.
\end{proof}

We will use the following lemma several times. It {bounds the sum of a sequence satisfying a contractive recursion}. We apply it to
{the expectations of} \(\|\bar y_k-\bar x_k\|^2\), \(\|\bar u_k\|^2\), and later to the
{expected value of the consensus Lyapunov function}.

\begin{lemma}
\label{lem:stable-sum}
Let \(\{s_k\}_{k\ge 0}\) and \(\{r_k\}_{k\ge 0}\) be nonnegative sequences such that \(s_0=0\) and
\[
s_{k+1}\le q s_k+r_k,
\qquad
q\in[0,1).
\]
Then, for every \(T\ge 1\),
\begin{equation}
\sum_{k=0}^{T-1} s_k
\le
\frac{1}{1-q}\sum_{k=0}^{T-1} r_k.
\label{eq:stable-sum}
\end{equation}
\end{lemma}
\begin{proof}
Summing \(s_{k+1}\le q s_k+r_k\) over \(k=0,\dots,T-1\) and using \(s_0=0\) gives
\[
(1-q)\sum_{k=0}^{T-1}s_k+s_T \le \sum_{k=0}^{T-1}r_k.
\]
Dropping \(s_T\ge 0\) and dividing by \(1-q\) yields \eqref{eq:stable-sum}.
\end{proof}

Applying Lemma~\ref{lem:stable-sum} to the two preceding inequalities and
substituting the resulting bounds into Lemma~\ref{lem:virtual-descent} gives
the following reduction.

\begin{proposition}
\label{prop:reduction}
Assume that
\[
\tilde\eta=\frac{\eta}{1-\beta_1}\le \frac{1}{8L}.
\]
Let \(K_g,K_z,K_\sigma\) be the constants defined below. If \(K_g>0\), then for every \(T\ge 1\),
\begin{align}
\frac1T\sum_{k=0}^{T-1}\E\norm{\nabla f(\bar z_k)}^2
\le\;&
\frac{f(\bar x_0)-f^\star}{K_g\,T}
+
\frac{K_z}{K_g}
\cdot
\frac1T\sum_{k=0}^{T-1}\E \left(\frac1n \norm{Z_k-\bar Z_k}_F^2\right)
+
\frac{K_\sigma}{K_g}\frac{\sigma^2}{n},
\label{eq:reduction-bound}
\end{align}
where

\begin{align}
K_g
\coloneqq\;&
\frac{\tilde\eta}{8}
-
3\tilde\eta L^2(1-b_{\min})^2
\Biggl[
\frac{4(1-a_{\min})^2\eta^2}{a_{\min}^2}
+
\frac{8\beta_1^2(1-a_{\min})^2\eta^2\beta_2^2}
{a_{\min}^2\bigl(1-\alpha_u^2\bigr)}
\left(
1+
\frac{2\alpha_u^2}
{1-\alpha_u^2}
\right)
\Biggr]
\notag\\
&\quad
-
3\tilde\eta L^2
\left(\frac{\eta\beta_1}{\beta_2(1-\beta_1)}\right)^2
\frac{4\beta_2^2}{1-\alpha_u^2}
\left(
1+
\frac{2\alpha_u^2}
{1-\alpha_u^2}
\right),
\label{eq:Kg-def}
\\
K_z
\coloneqq\;&
3\tilde\eta L^2(1-b_{\min})^2
\Biggl[
\frac{4L^2(1-a_{\min})^2\eta^2}{a_{\min}^2}
+
\frac{8\beta_1^2L^2(1-a_{\min})^2\eta^2\beta_2^2}
{a_{\min}^2\bigl(1-\alpha_u^2\bigr)}
\left(
1+
\frac{2\alpha_u^2}
{1-\alpha_u^2}
\right)
\Biggr]
\notag\\
&\quad+
3\tilde\eta L^2
\left(\frac{\eta\beta_1}{\beta_2(1-\beta_1)}\right)^2
L^2
\frac{4\beta_2^2}{1-\alpha_u^2}
\left(
1+
\frac{2\alpha_u^2}
{1-\alpha_u^2}
\right)
\notag\\
&\quad+
\tilde\eta L^2
+
L^3\tilde\eta^2,
\label{eq:Kz-def}
\\
K_\sigma
\coloneqq\;&
3\tilde\eta L^2(1-b_{\min})^2
\Biggl[
\frac{2(1-a_{\min})^2\eta^2}{a_{\min}^2}
+
\frac{4\beta_1^2(1-a_{\min})^2\eta^2\beta_2^2}
{a_{\min}^2\bigl(1-\alpha_u^2\bigr)}
\Biggr]
\notag\\
&\quad+
3\tilde\eta L^2
\left(\frac{\eta\beta_1}{\beta_2(1-\beta_1)}\right)^2
\frac{2\beta_2^2}{1-\alpha_u^2}
\notag\\
&\quad+
\frac{L\tilde\eta^2}{2}.
\label{eq:Ksigma-def}
\end{align}
\end{proposition}
\begin{proof}
Applying Lemma~\ref{lem:stable-sum} to Lemma~\ref{lem:u-coarse} with
\(s_k=\E\norm{\bar u_k}^2\), \(q=(1+\alpha_u^2)/2\), and
\(1-q=(1-\alpha_u^2)/2\), and using \(\bar u_0=0\), gives
\begin{align}
\sum_{k=0}^{T-1}\E\norm{\bar u_k}^2
\le\;&
\frac{4\beta_2^2}{1-\alpha_u^2}
\left(1+\frac{2\alpha_u^2}{1-\alpha_u^2}\right)
\sum_{k=0}^{T-1}\E\norm{\nabla f(\bar z_k)}^2
\notag\\
&\quad+
\frac{4L^2\beta_2^2}{1-\alpha_u^2}
\left(1+\frac{2\alpha_u^2}{1-\alpha_u^2}\right)
\sum_{k=0}^{T-1}\E\left(\frac1n\norm{Z_k-\bar Z_k}_F^2\right)
+
\frac{2\beta_2^2}{1-\alpha_u^2}\,T\frac{\sigma^2}{n}.
\label{eq:u-sum-reduction}
\end{align}
Similarly, applying Lemma~\ref{lem:stable-sum} to Lemma~\ref{lem:Gamma-coarse} with
\(s_k=\E\|\bar y_k-\bar x_k\|^2\), \(q=1-a_{\min}\), and using
\(\bar y_0=\bar x_0\), yields
\begin{align}
\sum_{k=0}^{T-1}\E\|\bar y_k-\bar x_k\|^2
\le\;&
\frac{2\beta_1^2(1-a_{\min})^2\eta^2}{a_{\min}^2}
\sum_{k=0}^{T-1}\E\norm{\bar u_k}^2
+
\frac{4(1-a_{\min})^2\eta^2}{a_{\min}^2}
\sum_{k=0}^{T-1}\E\norm{\nabla f(\bar z_k)}^2
\notag\\
&\quad+
\frac{4L^2(1-a_{\min})^2\eta^2}{a_{\min}^2}
\sum_{k=0}^{T-1}\E\left(\frac1n\norm{Z_k-\bar Z_k}_F^2\right)
+
\frac{2(1-a_{\min})^2\eta^2}{a_{\min}^2}\,T\frac{\sigma^2}{n}.
\label{eq:Gamma-sum-pre-reduction}
\end{align}
Substituting \eqref{eq:u-sum-reduction} into \eqref{eq:Gamma-sum-pre-reduction} gives
{\small
\begin{align}
&\sum_{k=0}^{T-1}\E\|\bar y_k-\bar x_k\|^2
\notag\\
\le\;&
\Biggl[
\frac{4(1-a_{\min})^2\eta^2}{a_{\min}^2}
+
\frac{8\beta_1^2(1-a_{\min})^2\eta^2\beta_2^2}
{a_{\min}^2\bigl(1-\alpha_u^2\bigr)}
\left(
1+
\frac{2\alpha_u^2}
{1-\alpha_u^2}
\right)
\Biggr]
\sum_{k=0}^{T-1}\E\norm{\nabla f(\bar z_k)}^2
\notag\\
&+
\Biggl[
\frac{4L^2(1-a_{\min})^2\eta^2}{a_{\min}^2}
+
\frac{8\beta_1^2L^2(1-a_{\min})^2\eta^2\beta_2^2}
{a_{\min}^2\bigl(1-\alpha_u^2\bigr)}
\left(
1+
\frac{2\alpha_u^2}
{1-\alpha_u^2}
\right)
\Biggr]
\sum_{k=0}^{T-1}\E\left(\frac1n\norm{Z_k-\bar Z_k}_F^2\right)
\notag\\
&+
\Biggl[
\frac{2(1-a_{\min})^2\eta^2}{a_{\min}^2}
+
\frac{4\beta_1^2(1-a_{\min})^2\eta^2\beta_2^2}
{a_{\min}^2\bigl(1-\alpha_u^2\bigr)}
\Biggr]
T\frac{\sigma^2}{n}.
\label{eq:Gamma-sum-reduction}
\end{align}}
{Taking total expectations in} \eqref{eq:virtual-descent}, {summing} over $k=0,\dots,T-1$, {and} using $\hat x_0=\bar x_0$ and $f(\hat x_T)\ge f^\star$ {gives}
{\small
\begin{align}
\frac{\tilde\eta}{8}
\sum_{k=0}^{T-1}\E\norm{\nabla f(\bar z_k)}^2
\le\;&
f(\bar x_0)-f^\star
+
3\tilde\eta L^2(1-b_{\min})^2
\sum_{k=0}^{T-1}\E\|\bar y_k-\bar x_k\|^2
\notag\\
&+
3\tilde\eta L^2
\left(\frac{\eta\beta_1}{\beta_2(1-\beta_1)}\right)^2
\sum_{k=0}^{T-1}\E\norm{\bar u_k}^2
+
\left(\tilde\eta L^2+L^3\tilde\eta^2\right)
\sum_{k=0}^{T-1}\E\left(\frac1n\norm{Z_k-\bar Z_k}_F^2\right)
\notag\\
&+
\frac{L\tilde\eta^2}{2}\,T\frac{\sigma^2}{n}.
\label{eq:summation-f-z}
\end{align}}
Substituting \eqref{eq:u-sum-reduction} and \eqref{eq:Gamma-sum-reduction}
into \eqref{eq:summation-f-z}, then collecting the coefficients according to
\eqref{eq:Kg-def}--\eqref{eq:Ksigma-def}, gives
\begin{equation}
K_g\sum_{k=0}^{T-1}\E\norm{\nabla f(\bar z_k)}^2
\le
f(\bar x_0)-f^\star
+
K_z\sum_{k=0}^{T-1}\E\left(\frac1n\norm{Z_k-\bar Z_k}_F^2\right)
+
K_\sigma T\frac{\sigma^2}{n}.
\label{eq:Kg-ineq}
\end{equation}
Since $K_g>0$, dividing \eqref{eq:Kg-ineq} by $K_g T$ proves \eqref{eq:reduction-bound}.
\end{proof}

We next give convenient sufficient conditions under which \(K_g\) is positive
and the ratios involving \(K_z\) and \(K_\sigma\) can be bounded by simpler
quantities.

\begin{lemma}
\label{lem:Kg-explicit-sufficient}
Under the assumptions of Proposition~\ref{prop:reduction}, suppose
\begin{equation}
\eta^2
\le
\frac{1}{
192\,L^2
\left[
\dfrac{(1-b_{\min})^2(1-a_{\min})^2}{a_{\min}^2}
\left(
1+
\dfrac{2\beta_1^2(1+\alpha_u^2)}
{(1-\beta_1)^2(1+\alpha_u)^2}
\right)
+
\dfrac{\beta_1^2(1+\alpha_u^2)}
{\beta_2^2(1-\beta_1)^4(1+\alpha_u)^2}
\right]
}.
\label{eq:Kg-stepsize-sufficient}
\end{equation}
Then \(K_g\ge \tilde\eta/16\). In particular, the simpler condition
\[
\eta^2
\le
\frac{a_{\min}^2\beta_2^2(1-\beta_1)^4}{192L^2}
\]
is sufficient.
\end{lemma}

\begin{proof}
Using
\[
1-\alpha_u^2=\beta_2(1-\beta_1)(1+\alpha_u),
\qquad
1+\frac{2\alpha_u^2}{1-\alpha_u^2}
=
\frac{1+\alpha_u^2}{1-\alpha_u^2},
\]
we have
\begin{equation}
\frac{4\beta_2^2}{1-\alpha_u^2}
\left(
1+\frac{2\alpha_u^2}{1-\alpha_u^2}
\right)
=
\frac{4(1+\alpha_u^2)}
{(1-\beta_1)^2(1+\alpha_u)^2}.
\label{eq:Kg-suff-identity}
\end{equation}
Substituting \eqref{eq:Kg-suff-identity} into \eqref{eq:Kg-def} gives
\begin{align}
\frac{\tilde\eta}{8}-K_g
=
12\tilde\eta L^2\eta^2
\Biggl[
&
\frac{(1-b_{\min})^2(1-a_{\min})^2}{a_{\min}^2}
\left(
1+
\frac{2\beta_1^2(1+\alpha_u^2)}
{(1-\beta_1)^2(1+\alpha_u)^2}
\right)
\notag\\
&\quad+
\frac{\beta_1^2(1+\alpha_u^2)}
{\beta_2^2(1-\beta_1)^4(1+\alpha_u)^2}
\Biggr].
\label{eq:Kg-suff-difference}
\end{align}
Thus \eqref{eq:Kg-stepsize-sufficient} implies
\[
\frac{\tilde\eta}{8}-K_g\le \frac{\tilde\eta}{16},
\]
and therefore \(K_g\ge \tilde\eta/16\).

It remains to justify the simpler condition. Let \(B\) denote the bracket in
\eqref{eq:Kg-stepsize-sufficient}. Since
\[
(1-b_{\min})^2\le 1,
\qquad
\frac{1+\alpha_u^2}{(1+\alpha_u)^2}\le 1,
\qquad
\beta_2\le 1,
\]
we obtain the chain
\begin{align}
a_{\min}^2\beta_2^2(1-\beta_1)^4 B
&\le
(1-a_{\min})^2\beta_2^2(1-\beta_1)^4
+
2\beta_1^2(1-a_{\min})^2\beta_2^2(1-\beta_1)^2
+
a_{\min}^2\beta_1^2
\notag\\
&\le
(1-a_{\min})^2(1-\beta_1)^2
\bigl((1-\beta_1)^2+2\beta_1^2\bigr)
+
a_{\min}^2\beta_1^2
\notag\\
&\le
(1-a_{\min})^2+a_{\min}^2
\le 1.
\label{eq:Kg-suff-bracket}
\end{align}
Here the third inequality uses
\[
(1-\beta_1)^2\bigl((1-\beta_1)^2+2\beta_1^2\bigr)\le 1,
\qquad
\beta_1^2\le 1,
\]
for \(\beta_1\in[0,1]\). Hence
\[
B
\le
\frac{1}{a_{\min}^2\beta_2^2(1-\beta_1)^4}.
\]
Therefore,
\[
\eta^2
\le
\frac{a_{\min}^2\beta_2^2(1-\beta_1)^4}{192L^2}
\]
implies \eqref{eq:Kg-stepsize-sufficient}, completing the proof.
\end{proof}

Under the same condition, the next lemma gives simple upper bounds on
\(K_z\) and \(K_\sigma\).

\begin{lemma}
\label{lem:Kz-Ksigma-explicit}
Under the assumptions of Proposition~\ref{prop:reduction} and the {learning-rate} condition
\eqref{eq:Kg-stepsize-sufficient},
\begin{align}
K_z &\le 2\,\tilde\eta L^2,
\label{eq:Kz-bound}
\\
K_\sigma &\le
\left(
\frac{9}{4a_{\min}^2}
+
\frac{3\beta_1^2}{4\beta_2(1-\beta_1)}
+
\frac12
\right)\tilde\eta^2 L.
\label{eq:Ksigma-bound}
\end{align}
\end{lemma}

\begin{proof}
Comparing the definitions of \(K_z\) and \(K_g\) in
\eqref{eq:Kz-def} and \eqref{eq:Kg-def}, we obtain
\[
K_z
=
L^2\left(\frac{\tilde\eta}{8}-K_g\right)
+
\tilde\eta L^2(1+L\tilde\eta).
\]
Thus,
\begin{align}
K_z
&\overset{(a)}{\le}
\frac{\tilde\eta L^2}{16}
+
\tilde\eta L^2(1+L\tilde\eta)
\notag\\
&\overset{(b)}{\le}
\frac{\tilde\eta L^2}{16}
+
\frac{9}{8}\tilde\eta L^2
\le
2\tilde\eta L^2.
\label{eq:Kz-chain}
\end{align}
Here, \((a)\) uses Lemma~\ref{lem:Kg-explicit-sufficient}, namely
\(K_g\ge\tilde\eta/16\), and \((b)\) uses \(L\tilde\eta\le 1/8\).
This proves \eqref{eq:Kz-bound}.

For \(K_\sigma\), using \((1-b_{\min})^2\le 1\),
\((1-a_{\min})^2\le 1\), \(1-\alpha_u^2\ge \beta_2(1-\beta_1)\), and
\[
\left(\frac{\eta\beta_1}{\beta_2(1-\beta_1)}\right)^2
=
\frac{\tilde\eta^2\beta_1^2}{\beta_2^2},
\]
we have
\begin{align}
K_\sigma
&\overset{(c)}{\le}
3\tilde\eta L^2
\left(
\frac{2\eta^2}{a_{\min}^2}
+
\frac{4\beta_1^2\beta_2\eta^2}{a_{\min}^2(1-\beta_1)}
\right)
+
3\tilde\eta L^2
\frac{\tilde\eta^2\beta_1^2}{\beta_2^2}
\cdot
\frac{2\beta_2^2}{\beta_2(1-\beta_1)}
+
\frac{L\tilde\eta^2}{2}
\notag\\
&\overset{(d)}{\le}
\frac{18\tilde\eta^3 L^2}{a_{\min}^2}
+
\frac{6\beta_1^2\tilde\eta^3 L^2}{\beta_2(1-\beta_1)}
+
\frac{L\tilde\eta^2}{2}
\notag\\
&\overset{(e)}{\le}
\frac{9\tilde\eta^2 L}{4a_{\min}^2}
+
\frac{3\beta_1^2\tilde\eta^2 L}{4\beta_2(1-\beta_1)}
+
\frac{L\tilde\eta^2}{2}.
\label{eq:Ksigma-chain}
\end{align}
Here, \((c)\) uses the elementary bounds stated above. For \((d)\), we use
\[
\frac{2\eta^2}{a_{\min}^2}
+
\frac{4\beta_1^2\beta_2\eta^2}{a_{\min}^2(1-\beta_1)}
\le
\frac{6\tilde\eta^2}{a_{\min}^2},
\]
which follows from
\[
2(1-\beta_1)^2+4\beta_1^2\beta_2(1-\beta_1)\le 6.
\]
Finally, \((e)\) uses \(L\tilde\eta\le 1/8\). This proves
\eqref{eq:Ksigma-bound}.
\end{proof}

\begin{remark}
\label{rem:K-scaling}
Under the basic {learning-rate} condition \eqref{eq:stepsize-smallness-basic} and
the sufficient condition \eqref{eq:Kg-stepsize-sufficient},
Lemma~\ref{lem:Kg-explicit-sufficient} and
Lemma~\ref{lem:Kz-Ksigma-explicit} yield
\[
K_g \ge \frac{\tilde\eta}{16},
\qquad
K_z \le 2\,\tilde\eta L^2,
\qquad
K_\sigma \le
\left(
\frac{9}{4a_{\min}^2}
+
\frac{3\beta_1^2}{4\beta_2(1-\beta_1)}
+
\frac12
\right)
\tilde\eta^2 L.
\]
Consequently,
\[
\frac{K_z}{K_g}\le 32L^2,
\qquad
\frac{K_\sigma}{K_g}\le
16
\left(
\frac{9}{4a_{\min}^2}
+
\frac{3\beta_1^2}{4\beta_2(1-\beta_1)}
+
\frac12
\right)
\tilde\eta L.
\]
These bounds are used in the {convergence results} below.
\end{remark}

Proposition~\ref{prop:reduction} shows that it remains to control {the expected consensus error of} \(Z_k\){, averaged over iterations}. In particular, an \(O(\eta^2)\) bound on
\[
\frac1T\sum_{k=0}^{T-1}
\E\left(\frac1n\|Z_k-\bar Z_k\|_F^2\right)
\]
gives an \(O(1/\sqrt T)\) {bound on the iteration average of \(\E\|\nabla f(\bar z_k)\|^2\)} when
\(\eta=\Theta(1/\sqrt T)\). We now prove {such a consensus-error bound}.

\subsection{Step 2: consensus error bound}
\label{app:step2}

We now control the consensus error of \(Z_k\). Since \(Z_k\) is formed from
\(X_k\) and \(Y_k\), we first bound the consensus errors of \(Y_k\), \(U_k\),
and \(X_k\), and then combine them to control \(Z_k\). {For the momentum variable} \(U_k\), {we use the scaled consensus error}
\[
V_k \coloneqq \eta^2\left(\frac1n\|U_k-\bar U_k\|_F^2\right).
\]
These bounds are then combined through a weighted Lyapunov function.

\begin{remark}
\label{rem:consensus-init-consequence}
By Assumption~\ref{ass:consensus-init},
\[
\frac1n\norm{X_0-\bar X_0}_F^2
=
\frac1n\norm{Y_0-\bar Y_0}_F^2
=
\eta^2\frac1n\norm{U_0-\bar U_0}_F^2
=
0.
\]
Thus, {the initial consensus errors are zero}, as required
{by} Lemma~\ref{lem:stable-sum}.
\end{remark}

We first bound the {deviations of the local stochastic gradients {from their network average}} in terms of the
consensus error of \(Z_k\), the {gradient} variance, and the {data} heterogeneity.

\begin{lemma}
\label{lem:G-consensus}
For every \(k\ge 0\),
\begin{equation}
\E_k \left(\frac1n \norm{G_k-\bar G_k}_F^2\right)
\le
4L^2 \left(\frac1n \norm{Z_k-\bar Z_k}_F^2\right)
+
4\zeta^2
+
2\sigma^2.
\label{eq:G-consensus}
\end{equation}
\end{lemma}

\begin{proof}
Define
$H_k \coloneqq [\nabla f_1(z_1^k),\dots,\nabla f_n(z_n^k)] .$
Since \(\bar G_k=G_kJ\) and \(\bar H_k=H_kJ\), we have
\[
G_k-\bar G_k
=
G_k(I-J)
=
(G_k-H_k)(I-J)+(H_k-\bar H_k).
\]
Taking {the} squared Frobenius norm and {then the} conditional expectation gives
\begin{align}
\E_k\|G_k-\bar G_k\|_F^2
&\le
2\,\E_k\|(G_k-H_k)(I-J)\|_F^2
+
2\,\|H_k-\bar H_k\|_F^2
\notag\\
&\le
2\,\E_k\|G_k-H_k\|_F^2
+
2\,\|H_k-\bar H_k\|_F^2
\notag\\
&\le
2n\sigma^2
+
2\,\|H_k-\bar H_k\|_F^2
\notag\\
&\le
2n\sigma^2
+
2\sum_{i=1}^n
\|\nabla f_i(z_i^k)-\nabla f(\bar z_k)\|^2
\notag\\
&\le
2n\sigma^2
+
4\sum_{i=1}^n
\|\nabla f_i(z_i^k)-\nabla f_i(\bar z_k)\|^2
+
4\sum_{i=1}^n
\|\nabla f_i(\bar z_k)-\nabla f(\bar z_k)\|^2
\notag\\
&\le
2n\sigma^2
+
4L^2\,\|Z_k-\bar Z_k\|_F^2
+
4n\zeta^2 .
\end{align}
The second inequality uses \(\|I-J\|_2\le 1\), and the third uses
Assumption~\ref{ass:bounded-variance}. The fourth inequality uses the fact
that the column average minimizes the sum of squared distances. The last two
inequalities use \(\|a+b\|^2\le 2\|a\|^2+2\|b\|^2\),
Assumption~\ref{ass:smoothness}, and
Assumption~\ref{ass:bounded-heterogeneity}. Dividing by \(n\) gives
\eqref{eq:G-consensus}.
\end{proof}

The next bound tracks how the \(Y\)-update {affects the consensus error.}

\begin{lemma}
\label{lem:Y-consensus}
Under Assumption~\ref{ass:mixing}, for every \(k\ge 0\),
\begin{align}
&\E_k \left(\frac1n \norm{Y_{k+1}-\bar Y_{k+1}}_F^2\right)
\notag\\
\le\;&
\left(1-\frac{\rho(2-\rho)}{2}\right)
\left(\frac1n \norm{Y_k-\bar Y_k}_F^2\right)
+
\frac{4\beta_1^2}{\rho(2-\rho)}\,\eta^2
\left(\frac1n \norm{U_k-\bar U_k}_F^2\right)
\notag\\
&\quad
+
\frac{4\eta^2}{\rho(2-\rho)}\,
\E_k \left(\frac1n \norm{G_k-\bar G_k}_F^2\right).
\label{eq:Y-consensus}
\end{align}
\end{lemma}

\begin{proof}
From the \(Y\)-update \eqref{eq:dpa-y} and \(W\one=\one\), we have
\begin{equation}
Y_{k+1}-\bar Y_{k+1}
=
\bigl((Y_k-\bar Y_k)
-\eta(\beta_1(U_k-\bar U_k)+(G_k-\bar G_k))\bigr)W.
\label{eq:Ycons-decomp}
\end{equation}
Applying Young's inequality
\[
\|p+q\|_F^2
\le
(1+\tau)\|p\|_F^2+
\left(1+\frac1\tau\right)\|q\|_F^2
\]
to \eqref{eq:Ycons-decomp} with
\[
\tau=\frac{\rho(2-\rho)}{2(1-\rho)^2},
\]
we obtain
\begin{align}
\|Y_{k+1}-\bar Y_{k+1}\|_F^2
&\le
(1+\tau)\|(Y_k-\bar Y_k)W\|_F^2
\notag\\
&\quad+
\left(1+\frac1\tau\right)\eta^2
\|(\beta_1(U_k-\bar U_k)+(G_k-\bar G_k))W\|_F^2
\notag\\
&\overset{(a)}{\le}
(1+\tau)(1-\rho)^2\|Y_k-\bar Y_k\|_F^2
\notag\\
&\quad+
\left(1+\frac1\tau\right)\eta^2
\|\beta_1(U_k-\bar U_k)+(G_k-\bar G_k)\|_F^2
\notag\\
&\overset{(b)}{\le}
\left(1-\frac{\rho(2-\rho)}{2}\right)
\|Y_k-\bar Y_k\|_F^2
+
\frac{2\eta^2}{\rho(2-\rho)}
\|\beta_1(U_k-\bar U_k)+(G_k-\bar G_k)\|_F^2
\notag\\
&\overset{(c)}{\le}
\left(1-\frac{\rho(2-\rho)}{2}\right)
\|Y_k-\bar Y_k\|_F^2
+
\frac{4\beta_1^2\eta^2}{\rho(2-\rho)}
\|U_k-\bar U_k\|_F^2
\notag\\
&\quad+
\frac{4\eta^2}{\rho(2-\rho)}
\|G_k-\bar G_k\|_F^2 .
\label{eq:Ydisc-proof-final}
\end{align}
Here, \((a)\) uses Remark~\ref{rem:frob-contraction} applied to
\(A=Y_k-\bar Y_k\), which is centered, for the first term, and
\(\|W\|_2\le 1\) from Remark~\ref{rem:spectral-W} for the second term.
\((b)\) follows from the choice of \(\tau\), since
\[
(1+\tau)(1-\rho)^2
=
1-\frac{\rho(2-\rho)}{2},
\qquad
1+\frac1\tau
\le
\frac{2}{\rho(2-\rho)}.
\]
\((c)\) uses
\[
\|\beta_1(U_k-\bar U_k)+(G_k-\bar G_k)\|_F^2
\le
2\beta_1^2\|U_k-\bar U_k\|_F^2
+
2\|G_k-\bar G_k\|_F^2.
\]
Dividing \eqref{eq:Ydisc-proof-final} by \(n\) and taking
\(\E_k[\cdot]\) gives \eqref{eq:Y-consensus}.
\end{proof}

We next control the consensus error of the {momentum} variable.

\begin{lemma}
\label{lem:U-consensus}
Define
\[
V_k \coloneqq \eta^2 \left(\frac1n \norm{U_k-\bar U_k}_F^2\right),
\qquad
\alpha_{u,\rho}
\coloneqq
(1-\beta_2)+\beta_2\beta_1(1-\rho).
\]
Then, under Assumption~\ref{ass:mixing}, for every \(k\ge 0\),
\begin{align}
\E_k V_{k+1}
\le\;&
\frac{1+\alpha_{u,\rho}^2}{2}\,V_k
\notag\\
&\quad+
\frac{
8\left(1+\alpha_{u,\rho}^2\right)\beta_2^2
}{
1-\alpha_{u,\rho}^2
}
\left(\frac1n \norm{Y_k-\bar Y_k}_F^2\right)
\notag\\
&\quad+
\frac{
2\left(1+\alpha_{u,\rho}^2\right)\beta_2^2
}{
1-\alpha_{u,\rho}^2
}
\,\eta^2
\E_k \left(\frac1n \norm{G_k-\bar G_k}_F^2\right).
\label{eq:U-consensus}
\end{align}
\end{lemma}
\begin{proof}
{Multiplying the \(U\)-update by $\eta$, we get}
\[
\eta U_{k+1}
=
(1-\beta_2)\eta U_k
+
\beta_2(Y_k-Y_{k+1}).
\]
Using the \(Y\)-update \eqref{eq:dpa-y}, this becomes
\[
\eta U_{k+1}
=
(1-\beta_2)\eta U_k
+
\beta_2 Y_k(I-W)
+
\beta_2\eta(\beta_1U_k+G_k)W.
\]
Since \(W\one=\one\), {subtracting the column average from each column} gives
\begin{align}
\eta(U_{k+1}-\bar U_{k+1})
&=
\eta(U_k-\bar U_k)\bigl((1-\beta_2)I+\beta_2\beta_1W\bigr)
\notag\\
&\quad+
\beta_2(Y_k-\bar Y_k)(I-W)
+
\beta_2\eta(G_k-\bar G_k)W.
\label{eq:Udisc-decomp}
\end{align}

For the first term in \eqref{eq:Udisc-decomp}, Assumption~\ref{ass:mixing}
gives
\begin{align}
&\left\|
\eta(U_k-\bar U_k)\bigl((1-\beta_2)I+\beta_2\beta_1W\bigr)
\right\|_F
\notag\\
&\quad\le
\eta(1-\beta_2)\|U_k-\bar U_k\|_F
+
\eta\beta_2\beta_1\|(U_k-\bar U_k)W\|_F
\notag\\
&\quad\le
\eta\bigl((1-\beta_2)+\beta_2\beta_1(1-\rho)\bigr)
\|U_k-\bar U_k\|_F
\notag\\
&\quad=
\alpha_{u,\rho}\eta\|U_k-\bar U_k\|_F .
\label{eq:Mu-contraction}
\end{align}
The first inequality uses the triangle inequality, and the second uses
Remark~\ref{rem:frob-contraction}. Applying Young's inequality in the form
\[
\|a+b+c\|_F^2
\le
(1+\gamma)\|a\|_F^2
+
2\left(1+\frac1\gamma\right)
\bigl(\|b\|_F^2+\|c\|_F^2\bigr)
\]
to \eqref{eq:Udisc-decomp}, with
\[
\gamma=
\frac{1-\alpha_{u,\rho}^2}{2\alpha_{u,\rho}^2},
\]
and using \eqref{eq:Mu-contraction}, we get
\begin{align}
\eta^2\|U_{k+1}-\bar U_{k+1}\|_F^2
&\le
\frac{1+\alpha_{u,\rho}^2}{2}\,
\eta^2\|U_k-\bar U_k\|_F^2
\notag\\
&\quad+
\frac{
2(1+\alpha_{u,\rho}^2)\beta_2^2
}{
1-\alpha_{u,\rho}^2
}
\|(Y_k-\bar Y_k)(I-W)\|_F^2
\notag\\
&\quad+
\frac{
2(1+\alpha_{u,\rho}^2)\beta_2^2
}{
1-\alpha_{u,\rho}^2
}
\eta^2\|(G_k-\bar G_k)W\|_F^2
\notag\\
&\le
\frac{1+\alpha_{u,\rho}^2}{2}\,
\eta^2\|U_k-\bar U_k\|_F^2
\notag\\
&\quad+
\frac{
8(1+\alpha_{u,\rho}^2)\beta_2^2
}{
1-\alpha_{u,\rho}^2
}
\|Y_k-\bar Y_k\|_F^2
\notag\\
&\quad+
\frac{
2(1+\alpha_{u,\rho}^2)\beta_2^2
}{
1-\alpha_{u,\rho}^2
}
\eta^2\|G_k-\bar G_k\|_F^2 .
\label{eq:Udisc-proof-final}
\end{align}
The last inequality uses \(\|I-W\|_2\le 2\) and \(\|W\|_2\le 1\).
Dividing \eqref{eq:Udisc-proof-final} by \(n\) and taking
\(\E_k[\cdot]\) gives \eqref{eq:U-consensus}.
\end{proof}

The {consensus error of {the output sequence}} \(X\) is {bounded in terms of} the consensus errors of \(X_k\) and
\(Y_{k+1}\).

\begin{lemma}
\label{lem:X-consensus}
Under Assumption~\ref{ass:mixing}, for every \(k\ge 0\),
\begin{align}
\E_k \left(\frac1n \norm{X_{k+1}-\bar X_{k+1}}_F^2\right)
\le\;&
(1-\rho)^2
\Biggl[
(1-a_{k+1})
\left(\frac1n \norm{X_k-\bar X_k}_F^2\right)
\notag\\
&\qquad\qquad
+
a_{k+1}\E_k\left(\frac1n \norm{Y_{k+1}-\bar Y_{k+1}}_F^2\right)
\Biggr].
\label{eq:X-consensus}
\end{align}
\end{lemma}

\begin{proof}
From the \(X\)-update \eqref{eq:dpa-x} and \(W\one=\one\),
\begin{equation}
X_{k+1}-\bar X_{k+1}
=
\bigl((1-a_{k+1})(X_k-\bar X_k)
+
a_{k+1}(Y_{k+1}-\bar Y_{k+1})\bigr)W .
\label{eq:Xdisc-decomp}
\end{equation}
Moreover,
\[
\bigl((1-a_{k+1})(X_k-\bar X_k)
+
a_{k+1}(Y_{k+1}-\bar Y_{k+1})\bigr)\one=0.
\]
Therefore, by Remark~\ref{rem:frob-contraction},
\begin{align}
\|X_{k+1}-\bar X_{k+1}\|_F^2
&\le
(1-\rho)^2
\bigl\|
(1-a_{k+1})(X_k-\bar X_k)
+
a_{k+1}(Y_{k+1}-\bar Y_{k+1})
\bigr\|_F^2
\notag\\
&\le
(1-\rho)^2
\Bigl[
(1-a_{k+1})\|X_k-\bar X_k\|_F^2
+
a_{k+1}\|Y_{k+1}-\bar Y_{k+1}\|_F^2
\Bigr].
\label{eq:Xdisc-proof-final}
\end{align}
The second inequality uses convexity of \(\|\cdot\|_F^2\) and
\(a_{k+1}\in[0,1]\). Dividing by \(n\) and taking \(\E_k[\cdot]\) proves
\eqref{eq:X-consensus}.
\end{proof}

{The definition of} \(Z_k\) {as a weighted average gives the following consensus error bound}.

\begin{lemma}
\label{lem:Z-from-XY}
For every \(k\ge 0\),
\begin{equation}
\frac1n \norm{Z_k-\bar Z_k}_F^2
\le
2\left(\frac1n \norm{X_k-\bar X_k}_F^2\right)
+
2\left(\frac1n \norm{Y_k-\bar Y_k}_F^2\right).
\label{eq:Z-from-XY}
\end{equation}
\end{lemma}

\begin{proof}
From the \(Z\)-update \eqref{eq:dpa-z},
\[
Z_k-\bar Z_k
=
b_{k+1}(Y_k-\bar Y_k)
+
(1-b_{k+1})(X_k-\bar X_k).
\]
Therefore,
\begin{align}
\norm{Z_k-\bar Z_k}_F^2
&\le
2b_{k+1}^2\norm{Y_k-\bar Y_k}_F^2
+
2(1-b_{k+1})^2\norm{X_k-\bar X_k}_F^2
\notag\\
&\le
2\norm{Y_k-\bar Y_k}_F^2
+
2\norm{X_k-\bar X_k}_F^2 .
\label{eq:Zdisc-proof-final}
\end{align}
The first inequality uses \(\norm{p+q}_F^2\le
2\norm{p}_F^2+2\norm{q}_F^2\), and the second uses
\(b_{k+1}\in[0,1]\). Dividing by \(n\) proves
\eqref{eq:Z-from-XY}.
\end{proof}

\subsection{Weighted Lyapunov bound}
\label{app:lyapunov}
We combine the preceding consensus-error inequalities into a single Lyapunov
{function}. The weights are chosen so that the terms involving \(X_k\), \(Y_k\),
and \(V_k\) {contract jointly in conditional expectation, up to {terms due to gradient variance and data heterogeneity}}. For weights \(\lambda_X,\lambda_Y,\lambda_V>0\), define
\[
\Psi_k
\coloneqq
\lambda_X\!\left(\tfrac1n \norm{X_k-\bar X_k}_F^2\right)
+\lambda_Y\!\left(\tfrac1n \norm{Y_k-\bar Y_k}_F^2\right)
+\lambda_V V_k.
\]
Lemma~\ref{lem:weighted-consensus-recursion} gives a one-step inequality for
\(\Psi_k\). Proposition~\ref{prop:explicit-weights} then gives explicit
weights for which
\[
\E_k\Psi_{k+1}\le (1-\chi)\Psi_k+C_\Psi,
\]
and Proposition~\ref{prop:consensus-explicit} sums this inequality to obtain
{the bound on the expected consensus error, averaged over iterations.}

\begin{lemma}
\label{lem:weighted-consensus-recursion}
Under Assumption~\ref{ass:mixing} with $\rho\in(0,1)$, for any
$\lambda_X,\lambda_Y,\lambda_V>0$,
\begin{equation}
\E_k\Psi_{k+1}
\le
\theta_X\left(\tfrac1n\norm{X_k-\bar X_k}_F^2\right)
+
\theta_Y\left(\tfrac1n\norm{Y_k-\bar Y_k}_F^2\right)
+
\theta_V V_k
+
C_\Psi,
\label{eq:weighted-consensus-recursion}
\end{equation}
where
\begin{align}
\theta_X
\coloneqq\;&
\lambda_X(1-\rho)^2
+
8L^2\eta^2
\Biggl[
\bigl(\lambda_X(1-\rho)^2+\lambda_Y\bigr)
\frac{4}{\rho(2-\rho)}
+
\lambda_V
\frac{
2(1+\alpha_{u,\rho}^2)\beta_2^2
}{
1-\alpha_{u,\rho}^2
}
\Biggr],
\label{eq:thetaX-def}
\\
\theta_Y
\coloneqq\;&
\lambda_X(1-\rho)^2\left(1-\frac{\rho(2-\rho)}{2}\right)
+
\lambda_Y\left(1-\frac{\rho(2-\rho)}{2}\right)
+
\lambda_V
\frac{
8(1+\alpha_{u,\rho}^2)\beta_2^2
}{
1-\alpha_{u,\rho}^2
}
\notag\\
&\quad+
8L^2\eta^2
\Biggl[
\lambda_X(1-\rho)^2\frac{4}{\rho(2-\rho)}
+
\lambda_Y\frac{4}{\rho(2-\rho)}
+
\lambda_V
\frac{
2(1+\alpha_{u,\rho}^2)\beta_2^2
}{
1-\alpha_{u,\rho}^2
}
\Biggr],
\label{eq:thetaY-def}
\\
\theta_V
\coloneqq\;&
\bigl(\lambda_X(1-\rho)^2+\lambda_Y\bigr)
\frac{4\beta_1^2}{\rho(2-\rho)}
+
\lambda_V
\frac{1+\alpha_{u,\rho}^2}{2},
\label{eq:thetaV-def}
\\
C_\Psi
\coloneqq\;&
\Biggl[
\bigl(\lambda_X(1-\rho)^2+\lambda_Y\bigr)
\frac{4}{\rho(2-\rho)}
+
\lambda_V
\frac{
2(1+\alpha_{u,\rho}^2)\beta_2^2
}{
1-\alpha_{u,\rho}^2
}
\Biggr]\eta^2(4\zeta^2+2\sigma^2).
\label{eq:CPsi-def}
\end{align}
In particular, if for some $\chi\in(0,1)$,
\begin{equation}
\theta_X\le (1-\chi)\lambda_X,
\qquad
\theta_Y\le (1-\chi)\lambda_Y,
\qquad
\theta_V\le (1-\chi)\lambda_V,
\label{eq:contraction-conditions}
\end{equation}
then
\[
\E_k\Psi_{k+1}\le (1-\chi)\Psi_k+C_\Psi.
\]
\end{lemma}

\begin{proof}
Combining Lemma~\ref{lem:G-consensus} with Lemma~\ref{lem:Z-from-XY}
gives
\begin{equation}
\E_k\!\left(\tfrac1n\norm{G_k-\bar G_k}_F^2\right)
\le
8L^2\!\left(\tfrac1n\norm{X_k-\bar X_k}_F^2\right)
+
8L^2\!\left(\tfrac1n\norm{Y_k-\bar Y_k}_F^2\right)
+
4\zeta^2+2\sigma^2.
\label{eq:G-consensus-XY}
\end{equation}
Substituting \eqref{eq:G-consensus-XY} into Lemma~\ref{lem:Y-consensus}
gives
\begin{align}
\E_k\!\left(\tfrac1n\norm{Y_{k+1}-\bar Y_{k+1}}_F^2\right)
\le\;&
8L^2\eta^2\frac{4}{\rho(2-\rho)}
\left(\tfrac1n\norm{X_k-\bar X_k}_F^2\right)
\notag\\
&+
\left(
1-\frac{\rho(2-\rho)}{2}
+
8L^2\eta^2\frac{4}{\rho(2-\rho)}
\right)
\left(\tfrac1n\norm{Y_k-\bar Y_k}_F^2\right)
\notag\\
&+
\frac{4\beta_1^2}{\rho(2-\rho)}V_k
+
\frac{4}{\rho(2-\rho)}\eta^2(4\zeta^2+2\sigma^2).
\label{eq:weighted-proof-Y}
\end{align}
Similarly, Lemma~\ref{lem:U-consensus} and \eqref{eq:G-consensus-XY} give
\begin{align}
\E_k V_{k+1}
\le\;&
8L^2\eta^2
\frac{
2(1+\alpha_{u,\rho}^2)\beta_2^2
}{
1-\alpha_{u,\rho}^2
}
\left(\tfrac1n\norm{X_k-\bar X_k}_F^2\right)
\notag\\
&+
\Biggl[
\frac{
8(1+\alpha_{u,\rho}^2)\beta_2^2
}{
1-\alpha_{u,\rho}^2
}
+
8L^2\eta^2
\frac{
2(1+\alpha_{u,\rho}^2)\beta_2^2
}{
1-\alpha_{u,\rho}^2
}
\Biggr]
\left(\tfrac1n\norm{Y_k-\bar Y_k}_F^2\right)
\notag\\
&+
\frac{1+\alpha_{u,\rho}^2}{2}V_k
+
\frac{
2(1+\alpha_{u,\rho}^2)\beta_2^2
}{
1-\alpha_{u,\rho}^2
}
\eta^2(4\zeta^2+2\sigma^2).
\label{eq:weighted-proof-V}
\end{align}
Lemma~\ref{lem:X-consensus}, combined with \eqref{eq:weighted-proof-Y} and the
bounds \(1-a_{k+1}\le 1\) and \(a_{k+1}\le 1\), gives
\begin{align}
\E_k\!\left(\tfrac1n\norm{X_{k+1}-\bar X_{k+1}}_F^2\right)
\le\;&
\left(
(1-\rho)^2
+
8L^2\eta^2(1-\rho)^2\frac{4}{\rho(2-\rho)}
\right)
\left(\tfrac1n\norm{X_k-\bar X_k}_F^2\right)
\notag\\
&+
(1-\rho)^2
\left(
1-\frac{\rho(2-\rho)}{2}
+
8L^2\eta^2\frac{4}{\rho(2-\rho)}
\right)
\left(\tfrac1n\norm{Y_k-\bar Y_k}_F^2\right)
\notag\\
&+
(1-\rho)^2\frac{4\beta_1^2}{\rho(2-\rho)}V_k
+
(1-\rho)^2\frac{4}{\rho(2-\rho)}
\eta^2(4\zeta^2+2\sigma^2).
\label{eq:weighted-proof-X}
\end{align}
Multiplying \eqref{eq:weighted-proof-X}, \eqref{eq:weighted-proof-Y}, and
\eqref{eq:weighted-proof-V} by \(\lambda_X,\lambda_Y,\lambda_V\),
respectively, and summing gives \eqref{eq:weighted-consensus-recursion}
with the coefficients in \eqref{eq:thetaX-def}--\eqref{eq:CPsi-def}.

If \eqref{eq:contraction-conditions} holds, then the first three terms in
\eqref{eq:weighted-consensus-recursion} are bounded by \((1-\chi)\Psi_k\).
Hence
\[
\E_k\Psi_{k+1}\le (1-\chi)\Psi_k+C_\Psi,
\]
which proves the claim.
\end{proof}

\begin{proposition}
\label{prop:explicit-weights}
Under Assumption~\ref{ass:mixing}, let
\[
\delta_\rho \coloneqq \rho(2-\rho),
\qquad
\mathcal M
\coloneqq
\frac{
\frac{4\beta_1^2}{\delta_\rho}
\cdot
\frac{
8(1+\alpha_{u,\rho}^2)\beta_2^2
}{
1-\alpha_{u,\rho}^2
}
}{
1-\alpha_{u,\rho}^2
}.
\]
Suppose that
\begin{equation}
64\,\mathcal M \le \delta_\rho .
\label{eq:momentum-smallness-explicit}
\end{equation}
Choose
\begin{align}
\lambda_X &\coloneqq 1,
\label{eq:lambdaX-choice}
\\
\lambda_Y
&\coloneqq
\frac{
16\left[
(1-\rho)^2\left(1-\frac{\delta_\rho}{2}\right)
+
8(1-\rho)^2\mathcal M
+
\frac{7\delta_\rho}{8}
\right]
}{
\delta_\rho
},
\label{eq:lambdaY-choice}
\\
\lambda_V
&\coloneqq
\frac{
8\bigl((1-\rho)^2+\lambda_Y\bigr)
\frac{4\beta_1^2}{\delta_\rho}
}{
1-\alpha_{u,\rho}^2
}.
\label{eq:lambdaV-choice}
\end{align}
Suppose further that
\begin{equation}
\eta^2
\le
\min\!\left\{
\frac{7\delta_\rho}
{64L^2\left[
\bigl((1-\rho)^2+\lambda_Y\bigr)\frac{4}{\delta_\rho}
+
\lambda_V
\frac{
2(1+\alpha_{u,\rho}^2)\beta_2^2
}{
1-\alpha_{u,\rho}^2
}
\right]},
\;
\frac{\lambda_Y\delta_\rho}
{64L^2\left[
\lambda_Y\frac{4}{\delta_\rho}
+
\lambda_V
\frac{
2(1+\alpha_{u,\rho}^2)\beta_2^2
}{
1-\alpha_{u,\rho}^2
}
\right]}
\right\}.
\label{eq:eta-smallness-explicit}
\end{equation}
Then, with
\[
\chi
\coloneqq
\frac18
\min\!\left\{
\delta_\rho,\,
1-\alpha_{u,\rho}^2
\right\},
\]
the contraction conditions \eqref{eq:contraction-conditions} hold.
\end{proposition}

\begin{proof}
We verify the three inequalities in \eqref{eq:contraction-conditions}.

First, we verify the condition for \(\theta_V\). Since \(\lambda_X=1\), the
definition of \(\theta_V\) gives
\begin{align}
\theta_V
&=
\left((1-\rho)^2+\lambda_Y\right)
\frac{4\beta_1^2}{\delta_\rho}
+
\lambda_V\frac{1+\alpha_{u,\rho}^2}{2}
\notag\\
&=
\frac{1-\alpha_{u,\rho}^2}{8}\lambda_V
+
\lambda_V\frac{1+\alpha_{u,\rho}^2}{2}
\notag\\
&=
\lambda_V
\left(
\frac{1-\alpha_{u,\rho}^2}{8}
+
\frac{1+\alpha_{u,\rho}^2}{2}
\right)
\notag\\
&=
\lambda_V
\left(
\frac{5+3\alpha_{u,\rho}^2}{8}
\right)
\notag\\
&=
\lambda_V
\left(
1-\frac{3(1-\alpha_{u,\rho}^2)}{8}
\right).
\label{eq:thetaV-bound-explicit}
\end{align}
Here, the second equality uses the choice of \(\lambda_V\) in
\eqref{eq:lambdaV-choice}. Therefore,
\(\theta_V\le (1-\chi)\lambda_V\) whenever
\begin{equation}
\chi
\le
\frac{3(1-\alpha_{u,\rho}^2)}{8}.
\label{eq:chi-V-constraint}
\end{equation}

Next, we verify the condition for \(\theta_Y\). Define
\[
A
\coloneqq
(1-\rho)^2\left(1-\frac{\delta_\rho}{2}\right)
+
8(1-\rho)^2\mathcal M .
\]
By the choice of \(\lambda_V\) in \eqref{eq:lambdaV-choice},
\[
\lambda_V
\frac{8(1+\alpha_{u,\rho}^2)\beta_2^2}{1-\alpha_{u,\rho}^2}
=
8\bigl((1-\rho)^2+\lambda_Y\bigr)\mathcal M .
\]
Using this identity in the definition of \(\theta_Y\), we get
\begin{align}
\theta_Y
&=
A
+
\lambda_Y\left(1-\frac{\delta_\rho}{2}+8\mathcal M\right)
+
8L^2\eta^2
\left[
(1-\rho)^2\frac{4}{\delta_\rho}
+
\lambda_Y\frac{4}{\delta_\rho}
+
\lambda_V\frac{2(1+\alpha_{u,\rho}^2)\beta_2^2}{1-\alpha_{u,\rho}^2}
\right]
\notag\\
&\le
A
+
\lambda_Y\left(1-\frac{\delta_\rho}{2}+8\mathcal M\right)
+
\frac{7\delta_\rho}{8}
+
\frac{\delta_\rho}{8}\lambda_Y
\notag\\
&=
A
+
\frac{7\delta_\rho}{8}
+
\lambda_Y\left(1-\frac{3\delta_\rho}{8}+8\mathcal M\right)
\notag\\
&\le
A
+
\frac{7\delta_\rho}{8}
+
\lambda_Y\left(1-\frac{\delta_\rho}{4}\right)
\notag\\
&=
\frac{\delta_\rho}{16}\lambda_Y
+
\lambda_Y\left(1-\frac{\delta_\rho}{4}\right)
\notag\\
&=
\lambda_Y\left(1-\frac{3\delta_\rho}{16}\right).
\label{eq:thetaY-bound-explicit}
\end{align}
Here, the first inequality uses the {learning-rate} condition
\eqref{eq:eta-smallness-explicit}; the second inequality uses
\eqref{eq:momentum-smallness-explicit}, namely
\(8\mathcal M\le \delta_\rho/8\); and the {penultimate line} uses the definition
of \(\lambda_Y\) in \eqref{eq:lambdaY-choice}. Therefore,
\(\theta_Y\le (1-\chi)\lambda_Y\) whenever
\begin{equation}
\chi
\le
\frac{3\delta_\rho}{16}.
\label{eq:chi-Y-constraint}
\end{equation}

Finally, we verify the condition for \(\theta_X\). Since \(\lambda_X=1\), the
definition of \(\theta_X\) gives
\begin{align}
\theta_X
&=
(1-\rho)^2
+
8L^2\eta^2
\left[
\bigl((1-\rho)^2+\lambda_Y\bigr)\frac{4}{\delta_\rho}
+
\lambda_V\frac{2(1+\alpha_{u,\rho}^2)\beta_2^2}{1-\alpha_{u,\rho}^2}
\right]
\notag\\
&\le
(1-\rho)^2+\frac{7\delta_\rho}{8}
\notag\\
&=
1-\delta_\rho+\frac{7\delta_\rho}{8}
\notag\\
&=
1-\frac{\delta_\rho}{8}.
\label{eq:thetaX-bound-explicit}
\end{align}
Here, the inequality uses the first part of the {learning-rate} condition
\eqref{eq:eta-smallness-explicit}, and the first equality after it uses
\((1-\rho)^2=1-\delta_\rho\). Therefore,
\(\theta_X\le (1-\chi)\lambda_X\) whenever
\begin{equation}
\chi
\le
\frac{\delta_\rho}{8}.
\label{eq:chi-X-constraint}
\end{equation}

It remains to check that our choice of \(\chi\) satisfies the three
requirements. Since
\[
\chi
=
\frac18
\min\!\left\{
\delta_\rho,\,
1-\alpha_{u,\rho}^2
\right\},
\]
we have
\[
\chi
\le
\frac{1-\alpha_{u,\rho}^2}{8}
\le
\frac{3(1-\alpha_{u,\rho}^2)}{8},
\qquad
\chi
\le
\frac{\delta_\rho}{8}
\le
\frac{3\delta_\rho}{16}.
\]
Thus \eqref{eq:chi-V-constraint}, \eqref{eq:chi-Y-constraint}, and
\eqref{eq:chi-X-constraint} all hold. Moreover,
\(\delta_\rho>0\) and \(\alpha_{u,\rho}<1\) because
\(\rho,\beta_1,\beta_2\in(0,1)\), so \(\chi>0\). Hence
\eqref{eq:contraction-conditions} holds.
\end{proof}

\begin{remark}
\label{rem:momentum-smallness-conservative}
{The condition on the momentum parameters} \eqref{eq:momentum-smallness-explicit} is used only to
{obtain an explicit contraction factor for the weighted consensus Lyapunov function}. It restricts the
admissible momentum parameters as a function of the spectral gap \(\rho\):
{as} \(\rho\) {decreases, the condition on the momentum parameters becomes more restrictive}. For example, when \(\beta_2\) is bounded away
from zero, it suffices to take {\(\beta_1\le c_\beta\rho\) with a sufficiently small constant \(c_\beta>0\)}.

Analogous {conditions on the momentum parameters} also appear in convergence analyses
of quasi-global momentum methods{~\citep{DBLP:conf/icml/0004KSJ21}}; for example, QG-DSGDm imposes a condition of
the form \(\beta/(1-\beta)\lesssim \rho\). These restrictions are sufficient conditions for the proof, not practical tuning rules.
{They arise from the particular Lyapunov weights and bounds from Young's inequality
used to obtain a joint contraction.}

In experiments, we tune \(\beta_1\) and \(\beta_2\) empirically. As is common
for momentum methods, {training can remain stable and effective outside the range covered by these sufficient conditions}; in our experiments, the selected ring values of $\beta_1$ are $0.90$ and $0.95$.
\end{remark}

Summing the Lyapunov inequality gives {the following bound on the expected consensus error, averaged over iterations.}
\begin{proposition}
\label{prop:consensus-explicit}
{Suppose} Assumptions~\ref{ass:mixing} and~\ref{ass:consensus-init} and the
assumptions of Proposition~\ref{prop:explicit-weights} {hold}. Then, for every
\(k\ge 0\),
\begin{equation}
\E_k\Psi_{k+1} \le (1-\chi)\Psi_k+C_\Psi,
\label{eq:Psi-recursion-explicit}
\end{equation}
and, for every \(T\ge 1\),
\begin{equation}
\frac1T\sum_{k=0}^{T-1}\E\!\left(\tfrac1n\norm{Z_k-\bar Z_k}_F^2\right)
\le
\frac{2C_\Psi}{\chi\,\min\{\lambda_X,\lambda_Y\}} .
\label{eq:EZ-final-explicit}
\end{equation}
\end{proposition}

\begin{proof}
By Proposition~\ref{prop:explicit-weights}, the contraction conditions
\eqref{eq:contraction-conditions} hold with the stated \(\chi\). Substituting
them into Lemma~\ref{lem:weighted-consensus-recursion} gives
\eqref{eq:Psi-recursion-explicit}.

Taking total expectation in \eqref{eq:Psi-recursion-explicit} and applying
Lemma~\ref{lem:stable-sum} with
\[
s_k=\E\Psi_k,\qquad q=1-\chi,\qquad r_k=C_\Psi,
\]
and using \(\Psi_0=0\), yields
\begin{equation}
\sum_{k=0}^{T-1}\E\Psi_k
\le
\frac{C_\Psi}{\chi}T.
\label{eq:Psi-sum-bound}
\end{equation}
Here, \(\Psi_0=0\) follows from Assumption~\ref{ass:consensus-init} and the
definition of \(\Psi_k\).

By Lemma~\ref{lem:Z-from-XY},
\[
\tfrac1n\norm{Z_k-\bar Z_k}_F^2
\le
2\left(\tfrac1n\norm{X_k-\bar X_k}_F^2\right)
+
2\left(\tfrac1n\norm{Y_k-\bar Y_k}_F^2\right)
\le
\frac{2}{\min\{\lambda_X,\lambda_Y\}}\Psi_k.
\]
Summing over \(k=0,\dots,T-1\), taking expectation, dividing by \(T\), and
using \eqref{eq:Psi-sum-bound} gives \eqref{eq:EZ-final-explicit}.
\end{proof}
\begin{remark}
\label{rem:CPsi-scaling}
With the choices in Proposition~\ref{prop:explicit-weights},
\(\lambda_X,\lambda_Y,\lambda_V\) and \(\chi\) depend only on
\(\rho,\beta_1,\beta_2\). Hence, from \eqref{eq:CPsi-def},
\[
C_\Psi
=
O\!\left(\eta^2(\sigma^2+\zeta^2)\right),
\]
where the hidden constant depends only on \(\rho,\beta_1,\beta_2\). Therefore,
under the conditions of Proposition~\ref{prop:consensus-explicit},
\[
\frac1T\sum_{k=0}^{T-1}
\E\!\left(\frac1n\|Z_k-\bar Z_k\|_F^2\right)
=
O\!\left(\eta^2(\sigma^2+\zeta^2)\right).
\]
\end{remark}
\begin{remark}
\label{rem:uniform-smallness}
Let \(\eta_{\mathrm{cons}}^2\) denote the right-hand side of
\eqref{eq:eta-smallness-explicit}. Assume the {condition on the momentum parameters}
\eqref{eq:momentum-smallness-explicit} holds. Let
\[
\eta=\frac{c}{\sqrt T},
\qquad
0<c\le c_0,
\]
where \(c_0>0\) is chosen so that
\[
c_0^2
\le
\min\!\left\{
\frac{(1-\beta_1)^2}{64L^2},
\;
\frac{a_{\min}^2\beta_2^2(1-\beta_1)^4}{192L^2},
\;
\eta_{\mathrm{cons}}^2
\right\}.
\]
Then, for every \(T\ge 1\), the choice \(\eta=c/\sqrt T\) satisfies the
{basic learning-rate condition} \eqref{eq:stepsize-smallness-basic}, the sufficient
condition \eqref{eq:Kg-stepsize-sufficient}, and the consensus {learning-rate}
condition \eqref{eq:eta-smallness-explicit}.
\end{remark}

% --- Combining the two steps ---

\subsection{Proof of Theorem~\ref{thm:full-rate-z}}
\label{app:combine-z}

\begin{proof}
By Proposition~\ref{prop:reduction} and \eqref{eq:EZ-final-explicit},
\[
\frac1T\sum_{k=0}^{T-1}\E\norm{\nabla f(\bar z_k)}^2
\le
\frac{f(\bar x_0)-f^\star}{K_g T}
+
\frac{K_z}{K_g}\cdot\frac{2C_\Psi}{\chi\min\{\lambda_X,\lambda_Y\}}
+
\frac{K_\sigma}{K_g}\cdot\frac{\sigma^2}{n}.
\]
Under \eqref{eq:Kg-stepsize-sufficient}, Lemmas~\ref{lem:Kg-explicit-sufficient}
and~\ref{lem:Kz-Ksigma-explicit} give
\[
\frac1{K_g}\le \frac{16}{\tilde\eta},
\qquad
\frac{K_z}{K_g}\le 32L^2,
\qquad
\frac{K_\sigma}{K_g}\le
16\!\left(
\frac{9}{4a_{\min}^2}
+
\frac{3\beta_1^2}{4\beta_2(1-\beta_1)}
+
\frac12
\right)\!\tilde\eta L.
\]
Substituting these bounds proves \eqref{eq:full-rate-z}.

Now choose \(\eta=c/\sqrt T\). Then
\[
\tilde\eta=\frac{c}{(1-\beta_1)\sqrt T},
\qquad
\eta^2=\frac{c^2}{T}.
\]
{With these identities, the term involving the initial objective gap in} \eqref{eq:full-rate-z} {becomes}
\[
\frac{16(1-\beta_1)(f(\bar x_0)-f^\star)}{c\sqrt T},
\]
{the gradient-variance term becomes}
\[
\frac{16}{1-\beta_1}\!\left(
\frac{9}{4a_{\min}^2}
+
\frac{3\beta_1^2}{4\beta_2(1-\beta_1)}
+
\frac12
\right)\!\frac{Lc\sigma^2}{n\sqrt T},
\]
and the consensus-error term, by \eqref{eq:CPsi-def}, becomes
\[
\frac{64L^2c^2}{\chi\min\{\lambda_X,\lambda_Y\}\,T}
\Biggl[
\bigl(\lambda_X(1-\rho)^2+\lambda_Y\bigr)\frac{4}{\delta_\rho}
+
\lambda_V\frac{2(1+\alpha_{u,\rho}^2)\beta_2^2}{1-\alpha_{u,\rho}^2}
\Biggr](4\zeta^2+2\sigma^2).
\]
Collecting $1/\sqrt T$ terms into $C_{z,1}$ and $1/T$ terms into
$C_{z,2}$ proves \eqref{eq:full-rate-z-explicit}, where
\begin{align}
C_{z,1}
&\coloneqq
\frac{16(1-\beta_1)(f(\bar x_0)-f^\star)}{c}
+
\frac{16}{1-\beta_1}\!\left(
\frac{9}{4a_{\min}^2}
+
\frac{3\beta_1^2}{4\beta_2(1-\beta_1)}
+
\frac12
\right)\!\frac{Lc\sigma^2}{n},
\label{eq:Cz1-def}
\\
C_{z,2}
&\coloneqq
\frac{64L^2c^2}{\chi\min\{\lambda_X,\lambda_Y\}}
\Biggl[
\bigl(\lambda_X(1-\rho)^2+\lambda_Y\bigr)\frac{4}{\delta_\rho}
+
\lambda_V\frac{2(1+\alpha_{u,\rho}^2)\beta_2^2}{1-\alpha_{u,\rho}^2}
\Biggr](4\zeta^2+2\sigma^2).
\label{eq:Cz2-def}
\end{align}
\end{proof}

\begin{proof}[Proof of Corollary~\ref{cor:linear-speedup-z}]
By \eqref{eq:c-condition-speedup}, the choice
\[
\eta=c\sqrt{\frac{n}{T}}
\]
satisfies the {learning-rate} conditions used in Theorem~\ref{thm:full-rate-z}.
Moreover,
\[
\tilde\eta
=
\frac{c}{1-\beta_1}\sqrt{\frac{n}{T}},
\qquad
\eta^2
=
\frac{c^2 n}{T}.
\]
{With these identities, the term involving the initial objective gap in} \eqref{eq:full-rate-z} {becomes}
\[
\frac{16(1-\beta_1)(f(\bar x_0)-f^\star)}{c\sqrt{nT}},
\]
{the gradient-variance term becomes}
\[
\frac{16}{1-\beta_1}\!\left(
\frac{9}{4a_{\min}^2}
+
\frac{3\beta_1^2}{4\beta_2(1-\beta_1)}
+
\frac12
\right)\!
\frac{cL\sigma^2}{\sqrt{nT}},
\]
and, by \eqref{eq:CPsi-def}, {the consensus-error term becomes}
\[
\frac{64L^2c^2 n}{\chi\min\{\lambda_X,\lambda_Y\}\,T}
\Biggl[
\bigl(\lambda_X(1-\rho)^2+\lambda_Y\bigr)\frac{4}{\delta_\rho}
+
\lambda_V\frac{2(1+\alpha_{u,\rho}^2)\beta_2^2}{1-\alpha_{u,\rho}^2}
\Biggr]
(4\zeta^2+2\sigma^2).
\]
Collecting the first two terms into \(D_1/\sqrt{nT}\) and the third term into
\(D_2n/T\) gives \eqref{eq:linear-speedup-z}, where
\begin{align}
D_1
&\coloneqq
\frac{16(1-\beta_1)(f(\bar x_0)-f^\star)}{c}
+
\frac{16}{1-\beta_1}\!\left(
\frac{9}{4a_{\min}^2}
+
\frac{3\beta_1^2}{4\beta_2(1-\beta_1)}
+
\frac12
\right)cL\sigma^2,
\label{eq:D1-def}
\\
D_2
&\coloneqq
\frac{64L^2c^2}{\chi\min\{\lambda_X,\lambda_Y\}}
\Biggl[
\bigl(\lambda_X(1-\rho)^2+\lambda_Y\bigr)\frac{4}{\delta_\rho}
+
\lambda_V\frac{2(1+\alpha_{u,\rho}^2)\beta_2^2}{1-\alpha_{u,\rho}^2}
\Biggr]
(4\zeta^2+2\sigma^2).
\label{eq:D2-def}
\end{align}
\end{proof}

\section{Proof of Theorem~\ref{thm:full-rate-x}}
\label{app:thm-x-rate}

We transfer the {bound on the expected squared gradient norm, averaged over iterations,} from \(\bar z_k\) to the {network-average} output
model \(\bar x_k\). Since
\[
\bar z_k-\bar x_k=b_{k+1}(\bar y_k-\bar x_k),
\]
it remains to control {the expected squared distance} between \(\bar y_k\) and \(\bar x_k\){, averaged over iterations}.

For compactness, define
\[
B_\Psi
\coloneqq
\Biggl[
\bigl(\lambda_X(1-\rho)^2+\lambda_Y\bigr)\frac{4}{\delta_\rho}
+
\lambda_V
\frac{2(1+\alpha_{u,\rho}^2)\beta_2^2}{1-\alpha_{u,\rho}^2}
\Biggr](4\zeta^2+2\sigma^2).
\]
Then \(C_\Psi=\eta^2 B_\Psi\). Under the choice \(\eta=c/\sqrt T\),
Proposition~\ref{prop:consensus-explicit} gives
\[
\frac1T\sum_{k=0}^{T-1}
\E\!\left(\frac1n\norm{Z_k-\bar Z_k}_F^2\right)
\le
\frac{C_{Z,2}}{T},
\qquad
C_{Z,2}
\coloneqq
\frac{2c^2B_\Psi}{\chi\min\{\lambda_X,\lambda_Y\}}.
\]
The constants in \eqref{eq:full-rate-x-explicit-T} may be chosen as
\begin{align}
C_{x,1}
&\coloneqq
2C_{z,1},
\label{eq:Cx1-def}
\\
C_{x,2}
&\coloneqq
2C_{z,2}
+
\frac{12L^2b_{\max}^2c^2}{a_{\min}^2(1-\beta_1)^2}
\cdot\frac{\sigma^2}{n},
\label{eq:Cx2-def}
\\
C_{x,3}
&\coloneqq
\frac{56L^2b_{\max}^2c^2}{a_{\min}^2(1-\beta_1)^2}
\left(C_{z,1}+C_{z,2}\right)
+
\frac{56L^4b_{\max}^2c^2}{a_{\min}^2(1-\beta_1)^2}
C_{Z,2}.
\label{eq:Cx3-def}
\end{align}

\begin{proof}
By $\bar z_k-\bar x_k=b_{k+1}(\bar y_k-\bar x_k)$,
$\|p+q\|^2\le 2\|p\|^2+2\|q\|^2$, Assumption~\ref{ass:smoothness}, and
$b_{k+1}\le b_{\max}$,
\[
\norm{\nabla f(\bar x_k)}^2
\le
2\norm{\nabla f(\bar z_k)}^2
+
2L^2 b_{\max}^2\|\bar y_k-\bar x_k\|^2.
\]
Taking expectation, summing over $k=0,\dots,T-1$, and dividing by $T$ gives
\begin{equation}
\frac1T\sum_{k=0}^{T-1}\E\norm{\nabla f(\bar x_k)}^2
\le
\frac{2}{T}\sum_{k=0}^{T-1}\E\norm{\nabla f(\bar z_k)}^2
+
2L^2 b_{\max}^2\cdot\frac1T\sum_{k=0}^{T-1}\E\|\bar y_k-\bar x_k\|^2.
\label{eq:x-from-z-gamma}
\end{equation}

Dividing \eqref{eq:Gamma-sum-reduction} by $T$ gives
\begin{equation}
\frac1T\sum_{k=0}^{T-1}\E\|\bar y_k-\bar x_k\|^2
\le
E_1\cdot\frac1T\sum_{k=0}^{T-1}\E\norm{\nabla f(\bar z_k)}^2
+
E_2\cdot\frac1T\sum_{k=0}^{T-1}\E\!\left(\tfrac1n\norm{Z_k-\bar Z_k}_F^2\right)
+
E_3\cdot\frac{\sigma^2}{n},
\label{eq:Gamma-average-explicit}
\end{equation}
where $E_1, E_2, E_3$ are {the corresponding coefficients in}
\eqref{eq:Gamma-sum-reduction}. Using
\[
\frac{(1-a_{\min})^2}{a_{\min}}\le \frac{1}{a_{\min}},
\qquad
1-\alpha_u^2\ge \beta_2(1-\beta_1),
\]
{{together with} $1-\alpha_u^2=\beta_2(1-\beta_1)(1+\alpha_u)$ and $(1+\alpha_u^2)/(1+\alpha_u)^2\le1$, {we obtain}}
\[
E_1\le \frac{28\eta^2}{a_{\min}^2(1-\beta_1)^2},
\qquad
E_2\le \frac{28L^2\eta^2}{a_{\min}^2(1-\beta_1)^2},
\qquad
E_3\le \frac{6\eta^2}{a_{\min}^2(1-\beta_1)^2}.
\]

Now choose \(\eta=c/\sqrt T\). By Theorem~\ref{thm:full-rate-z},
\[
\frac1T\sum_{k=0}^{T-1}\E\norm{\nabla f(\bar z_k)}^2
\le
\frac{C_{z,1}}{\sqrt T}
+
\frac{C_{z,2}}{T}.
\]
Also, by Proposition~\ref{prop:consensus-explicit} and \(C_\Psi=\eta^2B_\Psi\),
\[
\frac1T\sum_{k=0}^{T-1}
\E\!\left(\tfrac1n\norm{Z_k-\bar Z_k}_F^2\right)
\le
\frac{C_{Z,2}}{T}.
\]
Moreover,
\[
E_1\le
\frac{28c^2}{a_{\min}^2(1-\beta_1)^2}\cdot\frac1T,
\qquad
E_2\le
\frac{28L^2c^2}{a_{\min}^2(1-\beta_1)^2}\cdot\frac1T,
\qquad
E_3\le
\frac{6c^2}{a_{\min}^2(1-\beta_1)^2}\cdot\frac1T.
\]
Substituting these bounds into \eqref{eq:Gamma-average-explicit} gives
\begin{align}
\frac1T\sum_{k=0}^{T-1}\E\|\bar y_k-\bar x_k\|^2
\le\;&
\frac{28c^2C_{z,1}}{a_{\min}^2(1-\beta_1)^2}\cdot\frac1{T^{3/2}}
+
\frac{28c^2C_{z,2}}{a_{\min}^2(1-\beta_1)^2}\cdot\frac1{T^2}
\notag\\
&+
\frac{28L^2c^2C_{Z,2}}{a_{\min}^2(1-\beta_1)^2}\cdot\frac1{T^2}
+
\frac{6c^2}{a_{\min}^2(1-\beta_1)^2}
\cdot\frac{\sigma^2}{n}\cdot\frac1T .
\label{eq:Gamma-average-final-rate}
\end{align}
Since \(T^{-2}\le T^{-3/2}\) for \(T\ge1\), we obtain
\begin{align}
\frac1T\sum_{k=0}^{T-1}\E\|\bar y_k-\bar x_k\|^2
\le\;&
\frac{6c^2}{a_{\min}^2(1-\beta_1)^2}
\cdot\frac{\sigma^2}{n}\cdot\frac1T
\notag\\
&+
\frac{28c^2}{a_{\min}^2(1-\beta_1)^2}
\left(C_{z,1}+C_{z,2}+L^2C_{Z,2}\right)
\frac1{T^{3/2}}.
\label{eq:Gamma-average-final-rate-simplified}
\end{align}
Finally, substituting \eqref{eq:Gamma-average-final-rate-simplified} into
\eqref{eq:x-from-z-gamma} gives
\[
\frac1T\sum_{k=0}^{T-1}\E\norm{\nabla f(\bar x_k)}^2
\le
\frac{C_{x,1}}{\sqrt T}
+
\frac{C_{x,2}}{T}
+
\frac{C_{x,3}}{T^{3/2}},
\]
with \(C_{x,1},C_{x,2},C_{x,3}\) defined in
\eqref{eq:Cx1-def}--\eqref{eq:Cx3-def}. This proves
\eqref{eq:full-rate-x-explicit-T}.
\end{proof}

% ============================================================

\subsection{Proof of Corollary~\ref{cor:linear-speedup-x}}
\label{app:linear-speedup-x}

\begin{corollary}[{Linear-speedup bound for the network-average output sequence}]
\label{cor:linear-speedup-x}
Under the assumptions of Corollary~\ref{cor:linear-speedup-z}, including
\eqref{eq:c-condition-speedup}, the choice $\eta=c\sqrt{n/T}$ gives
\begin{equation}
\frac1T\sum_{k=0}^{T-1}\E\|\nabla f(\bar x_k)\|^2
\le
\frac73\left(\frac{D_1}{\sqrt{nT}}+\frac{D_2n}{T}\right).
\label{eq:linear-speedup-x}
\end{equation}
\end{corollary}

\begin{proof}
Write $S_{x,T}=T^{-1}\sum_{k=0}^{T-1}\E\|\nabla f(\bar x_k)\|^2$,
with $S_{z,T}$ defined analogously. Let $E_{Z,T}$ be the left-hand side
of \eqref{eq:EZ-final-explicit}. Equations~\eqref{eq:x-from-z-gamma}
and~\eqref{eq:Gamma-average-explicit}, together with the bounds on
$E_1,E_2,E_3$, give
\begin{equation}
S_{x,T}\le (2+q)S_{z,T}+qL^2E_{Z,T}+V,
\label{eq:output-transfer-absorption}
\end{equation}
where
\[
q=\frac{56L^2b_{\max}^2\tilde\eta^2}{a_{\min}^2},
\qquad
V=\frac{12L^2b_{\max}^2\tilde\eta^2\sigma^2}{a_{\min}^2n}.
\]

Write $I,N,R$ for {the} {{terms involving the initial objective gap, consensus error, and gradient variance
in} \eqref{eq:full-rate-z}{, respectively, so that}
$S_{z,T}\le I+N+R$.}
By \eqref{eq:c-condition-speedup},
$q\le(7/24)b_{\max}^2\beta_2^2(1-\beta_1)^2\le7/24$.
Also, $L^2E_{Z,T}\le N/32$ by \eqref{eq:EZ-final-explicit}.
Since $L\tilde\eta\le1/8$ and
$R\ge36L\tilde\eta\sigma^2/(a_{\min}^2n)$, we have $V\le R/24$.
Hence
\[
\begin{aligned}
S_{x,T}
&\le \tfrac{55}{24}I
 +\left(\tfrac{55}{24}+\tfrac{7}{768}\right)N
 +\tfrac73 R \\
&\le \tfrac73(I+N+R).
\end{aligned}
\]
{For $\eta=c\sqrt{n/T}$, \eqref{eq:D1-def} and \eqref{eq:D2-def}
give $I+R=D_1/\sqrt{nT}$ and $N=D_2n/T$, which proves
\eqref{eq:linear-speedup-x}.}
\end{proof}

\begin{remark}
\label{rem:speedup-x-transfer}
Keeping the individual terms in \eqref{eq:output-transfer-absorption}
gives, for $\eta=c\sqrt{n/T}$,
\begin{equation}
\begin{aligned}
S_{x,T}\le{}&
\frac{2D_1}{\sqrt{nT}}+\frac{2D_2n}{T}\\
&+\mathcal O\!\left(
\frac{\sqrt n}{T^{3/2}}+\frac{n^2}{T^2}+\frac{\sigma^2}{T}
\right).
\end{aligned}
\label{eq:linear-speedup-x-transfer}
\end{equation}
The constants are independent of $T$ for fixed problem, network, and
algorithm parameters: $q=O(n/T)$, $E_{Z,T}=O(n/T)$, and
$V=O(\sigma^2/T)$.
{Remark~\ref{rem:speedup-x-transfer-main} gives the threshold
at which $D_2n/T$ is no larger than $D_1/\sqrt{nT}$.}
\end{remark}

\subsection{{Squared gradient norm at} the local output models}
\label{app:local-output}
Under the assumptions of Theorem~\ref{thm:full-rate-x},
\begin{equation}
\begin{aligned}
\frac1{nT}\sum_{k=0}^{T-1}\sum_{i=1}^n\E\|\nabla f(x_i^k)\|^2
\le{}&\frac2T\sum_{k=0}^{T-1}\E\|\nabla f(\bar x_k)\|^2\\
&+\frac{2L^2C_\Psi}{\chi\lambda_X}.
\end{aligned}
\label{eq:local-output-stationarity}
\end{equation}

Smoothness of $f$ gives
\[
\frac1n\sum_i\|\nabla f(x_i^k)\|^2
\le 2\|\nabla f(\bar x_k)\|^2
+\frac{2L^2}{n}\|X_k-\bar X_k\|_F^2.
\]
The definition of $\Psi_k$ implies
$n^{-1}\|X_k-\bar X_k\|_F^2\le\Psi_k/\lambda_X$.
Taking expectations, averaging over $k$, and using
\eqref{eq:Psi-sum-bound} proves
\eqref{eq:local-output-stationarity}. Because $C_\Psi=\eta^2B_\Psi$,
{{the consensus-error term is}} $O(1/T)$ for $\eta=c/\sqrt T$.
This bound {concerns the global gradient {at the local output models and}}
is averaged over agents and iterations; {it is not a last-iterate
guarantee}.

\section{Additional experiments and hyperparameters}
\label{app:experiments}

\subsection{{DPA-1G with} {{one communication round per iteration}}}
\label{app:DPA-1G}

DPA uses two {communication rounds} per iteration, one for the optimization sequence
and one for the {output sequence}. To match the {per-iteration}
{communication cost of DSGD, QG-DSGDm, and}
QG-GUTm, we also study a practical variant called DPA-1G. DPA-1G performs
exactly one \textsc{SendReceive} operation per iteration.

DPA-1G {uses an integer period} \(T_x>1\).
On {optimization} iterations, {agents update the optimization sequence and apply gossip averaging},
update the quasi-global momentum buffer, and update the {output
sequence} locally. Every \(T_x\) iterations, the method instead performs {a
gradient-free iteration. It} skips the stochastic-gradient computation, keeps the
optimization and momentum variables fixed, and uses {the single communication round} to mix the {output sequence}.

\begin{algorithm}[t]
\caption{{DPA-1G with} {{one communication round per iteration}}}
\label{alg:DPA-1G}
\textbf{Input:} Initial model $x_i^0$ at agent $i$,
$y_i^0=x_i^0$, $u_i^0=0$, number of iterations $T$,
learning rate $\eta$, momentum parameters
$\beta_1,\beta_2\in(0,1)$, schedules $\{a_k\}_{k\ge1}$ and
$\{b_k\}_{k\ge1}$, mixing weights $\{w_{ij}\}_{j\in\mathcal N(i)}$,
neighbor set $\mathcal N(i)$ including $i$, and integer period
$T_x>1$.

\begin{algorithmic}[1]
\Procedure{{AGENT-}$i$}{}
  \For{$k=0,1,\dots,T-1$}
    \If{$(k+1)\bmod T_x \neq 0$}
      \Comment{{Optimization iteration with one communication round}}
      \State $z_i^k=b_{k+1}y_i^k+(1-b_{k+1})x_i^k$
      \State Sample $\xi_i^k\sim\mathcal D_i$ and compute
      $g_i^k=\nabla F_i(z_i^k;\xi_i^k)$
      \State $v_i^k=y_i^k-\eta(\beta_1u_i^k+g_i^k)$
      \State \textsc{SendReceive}$(v_i^k)$
      \State $y_i^{k+1}=\sum_{j\in\mathcal N(i)}w_{ij}v_j^k$
      \State $d_i^{k+1}=(y_i^k-y_i^{k+1})/\eta$
      \State $u_i^{k+1}=(1-\beta_2)u_i^k+\beta_2 d_i^{k+1}$
      \State $x_i^{k+1}=(1-a_{k+1})x_i^k+a_{k+1}y_i^{k+1}$
      \Comment{{Local averaging}}
    \Else
      \Comment{{Gradient-free iteration with one communication round}}
      \State $y_i^{k+1}=y_i^k$
      \State $u_i^{k+1}=u_i^k$
      \State $s_i^k=(1-a_{k+1})x_i^k+a_{k+1}y_i^k$
      \State \textsc{SendReceive}$(s_i^k)$
      \State $x_i^{k+1}=\sum_{j\in\mathcal N(i)}w_{ij}s_j^k$
    \EndIf
  \EndFor
  \State \Return $x_i^T$
\EndProcedure
\end{algorithmic}
\end{algorithm}

On {optimization} iterations, DPA-1G communicates only the {locally updated model}
\(v_i^k\). The {output sequence} is updated locally as
\[
  x_i^{k+1}
  =
  (1-a_{k+1})x_i^k+a_{k+1}y_i^{k+1}.
\]
No {additional} communication is used for \(x_i^{k+1}\) on these
iterations. On {gradient-free iterations}, DPA-1G keeps \(y_i^k\) and \(u_i^k\)
fixed, skips the stochastic-gradient computation, and uses the single {communication round} to compute
\[
  x_i^{k+1}
  =
  \sum_{j\in\mathcal N(i)}w_{ij}
  \bigl((1-a_{k+1})x_j^k+a_{k+1}y_j^k\bigr).
\]
Thus, {each agent sends one model-sized vector to each neighbor per
iteration}. Smaller values of \(T_x\) {mix the output models more
often}, while larger values {allow more stochastic-gradient updates
between {gradient-free iterations}}.

In all reported experiments, we use \(T_x=5\). Thus, four out of every five
iterations perform a stochastic-gradient update and {apply gossip averaging
to the optimization sequence}, while the remaining iteration is a {gradient-free
iteration.} {{The iteration counter continues across epoch
boundaries.} DPA-1G computes $K-\lfloor K/T_x\rfloor$ {mini-batch} gradients
per agent in {the first} $K$ iterations.}
{The warmup and coefficient schedules advance on every iteration,
including {gradient-free iterations}.}

\subsection{Architectures and training details}
\label{app:architectures}

We follow the architecture and training setup of the public GUT
codebase~\citep{aketi2023global}. This includes the model definitions,
{EvoNorm normalization for the CIFAR models}, preprocessing, data augmentation, and {the} simulated
distributed training setup. We use LeNet-5 for Fashion-MNIST, VGG-11 and
ResNet-20 for CIFAR-10, and ResNet-20 for CIFAR-100. This matches the setup
used for the GUT baselines. {LeNet-5 has no normalization layer; VGG-11 uses the benchmark's reduced-width configuration. Torus neighbors are $i\pm1,i\pm4$ modulo $n$. Uniform mixing gives {ring, torus, and Dyck} self-weights $1/3$, $1/5$ and $1/4$, respectively.}

\subsection{Hyperparameter tables}
\label{app:hyperparams}

For DSGD, QG-DSGDm, and QG-GUTm, we adopt the hyperparameters
from the public GUT codebase~\citep{aketi2023global},
which we also tested in our experimental settings.

For the equal-epoch non-IID comparisons,
Table~\ref{tab:learning-rates-main} lists the learning rates for the
16-agent ring. Table~\ref{tab:learning-rates-extended} gives the learning
rates for CIFAR-10/ResNet-20 on the 32-agent ring.
Table~\ref{tab:learning-rates-extended-four-methods} gives the learning
rates for the other 32-agent CIFAR-10 comparisons, using VGG-11 on a
ring and ResNet-20 on the torus and Dyck graphs.
In these three tables, the last column applies to both DPA and DPA-1G.

We train for 100 epochs on Fashion-MNIST and 200 epochs on CIFAR-10
and CIFAR-100. Each agent uses a mini-batch size of \(32\).
The total mini-batch size across agents is \(512\) for \(n=16\)
and \(1024\) for \(n=32\). For these comparisons, we use random
seeds \(0\), \(1\), and \(2\).

\begin{table}[htbp]
\centering
\caption{Learning rates on the 16-agent ring.}
\label{tab:learning-rates-main}
\setlength{\tabcolsep}{6pt}
\resizebox{\linewidth}{!}{%
\begin{tabular}{@{}lccccccc@{}}
\toprule
Dataset / Model & $\alpha$ & DSGD & GT & QG-DSGDm & MT & QG-GUTm &
\shortstack{DPA /\\DPA-1G} \\
\midrule
Fashion-MNIST / LeNet-5 & $0.1$ & $0.1$ & $0.01$ & $0.1$ & $0.05$ & $0.1$ & $0.1$ \\
Fashion-MNIST / LeNet-5 & $0.01$ & $0.1$ & $0.01$ & $0.1$ & $0.05$ & $0.1$ & $0.1$ \\
CIFAR-10 / VGG-11 & $0.1$ & $0.1$ & $0.1$ & $0.1$ & $0.1$ & $0.1$ & $5.0$ \\
CIFAR-10 / VGG-11 & $0.01$ & $0.1$ & $0.1$ & $0.1$ & $0.05$ & $0.1$ & $5.0$ \\
CIFAR-10 / ResNet-20 & $0.1$ & $0.1$ & $0.1$ & $0.1$ & $0.1$ & $0.1$ & $5.0$ \\
CIFAR-10 / ResNet-20 & $0.01$ & $0.1$ & $0.1$ & $0.1$ & $0.1$ & $0.1$ & $5.0$ \\
CIFAR-100 / ResNet-20 & $0.1$ & $0.1$ & $0.1$ & $0.1$ & $0.1$ & $0.1$ & $5.0$ \\
CIFAR-100 / ResNet-20 & $0.01$ & $0.1$ & $0.1$ & $0.1$ & $0.05$ & $0.1$ & $5.0$ \\
\bottomrule
\end{tabular}}
\end{table}

\begin{table}[htbp]
\centering
\caption{ResNet-20 learning rates on the 32-agent ring.}
\label{tab:learning-rates-extended}
\setlength{\tabcolsep}{6pt}
\resizebox{\linewidth}{!}{%
\begin{tabular}{@{}lccccccc@{}}
\toprule
Model / Graph & $\alpha$ & DSGD & GT & QG-DSGDm & MT & QG-GUTm &
\shortstack{DPA /\\DPA-1G} \\
\midrule
ResNet-20 / Ring & $0.1$ & $0.1$ & $0.05$ & $0.1$ & $0.05$ & $0.1$ & $5.0$ \\
ResNet-20 / Ring & $0.01$ & $0.1$ & $0.05$ & $0.1$ & $0.05$ & $0.1$ & $5.0$ \\
\bottomrule
\end{tabular}}
\end{table}

\begin{table}[htbp]
\centering
\caption{Learning rates for additional 32-agent comparisons.}
\label{tab:learning-rates-extended-four-methods}
\setlength{\tabcolsep}{6pt}
\small
\begin{tabular}{@{}lcccc@{}}
\toprule
Model / Graph & $\alpha$ & QG-DSGDm & QG-GUTm &
\shortstack{DPA /\\DPA-1G} \\
\midrule
VGG-11 / Ring & $0.1$ & $0.1$ & $0.1$ & $5.0$ \\
VGG-11 / Ring & $0.01$ & $0.1$ & $0.1$ & $5.0$ \\
ResNet-20 / Torus & $0.01$ & $0.1$ & $0.1$ & $5.0$ \\
ResNet-20 / Dyck & $0.01$ & $0.1$ & $0.1$ & $5.0$ \\
\bottomrule
\end{tabular}
\end{table}

For DSGD, GT, QG-DSGDm, MT, and QG-GUTm, we divide the learning
rate by \(10\) after \(50\%\) and \(75\%\) of training.
For DPA and DPA-1G, we increase the learning rate linearly to
\(\eta\) during warmup and then keep it constant.

QG-DSGDm and QG-GUTm use a momentum parameter of \(0.9\).
Table~\ref{tab:qg-gutm-scaling} lists the QG-GUTm scaling coefficients
for the equal-epoch non-IID comparisons.

\begin{table}[htbp]
\centering
\caption{QG-GUTm scaling coefficients.}
\label{tab:qg-gutm-scaling}
\setlength{\tabcolsep}{6pt}
\small
\begin{tabular}{@{}lclcc@{}}
\toprule
Dataset / Model & $n$ & Graph & $\alpha$ & Scaling coefficient \\
\midrule
Fashion-MNIST / LeNet-5 & $16$ & Ring & $0.1$ & $0.01$ \\
Fashion-MNIST / LeNet-5 & $16$ & Ring & $0.01$ & $0.005$ \\
CIFAR-10 / VGG-11 & $16$ & Ring & $0.1$ & $0.08$ \\
CIFAR-10 / VGG-11 & $16$ & Ring & $0.01$ & $0.09$ \\
CIFAR-10 / ResNet-20 & $16$ & Ring & $0.1,\ 0.01$ & $0.06$ \\
CIFAR-100 / ResNet-20 & $16$ & Ring & $0.1,\ 0.01$ & $0.005$ \\
CIFAR-10 / VGG-11 & $32$ & Ring & $0.1,\ 0.01$ & $0.08$ \\
CIFAR-10 / ResNet-20 & $32$ & Ring & $0.1,\ 0.01$ & $0.04$ \\
CIFAR-10 / ResNet-20 & $32$ & Torus & $0.01$ & $0.05$ \\
CIFAR-10 / ResNet-20 & $32$ & Dyck & $0.01$ & $0.05$ \\
\bottomrule
\end{tabular}
\end{table}

MT uses a momentum parameter of \(0.9\).
The implementation settings for GT and MT are specified in the
released code.
% The trainer describes --mt-paper-init as gradient initialization
% using an all-reduce operation.
% AUTHOR NOTE: These common auxiliary settings are directly supported by the
% supplied configurations and logs for several settings, including all three
% seeds of CIFAR-10/VGG-11 on the 16-agent ring at alpha=0.1 (job17536258).
% For the following historical runs, they use the author's recollection that
% weight decay and initialization were shared across models and datasets.
% GT weight decay is assumed to be zero for the 16-agent ring on
% CIFAR-10/VGG-11 at alpha=0.01 and both ResNet-20 tasks at both alpha values.
% MT weight decay is assumed to be zero and --mt-paper-init enabled for the
% 16-agent ring on CIFAR-10/ResNet-20 at both alpha values and
% CIFAR-100/ResNet-20 at alpha=0.1.

Table~\ref{tab:dpa-hyperparams} lists the training epochs, learning
rates, and momentum parameters used in the equal-epoch non-IID comparisons.

\begin{table}[htbp]
\centering
\caption{DPA and DPA-1G hyperparameters.}
\label{tab:dpa-hyperparams}
\setlength{\tabcolsep}{8pt}
\resizebox{\linewidth}{!}{%
\begin{tabular}{@{}llccccc@{}}
\toprule
Dataset / Model & Graph & $n$ & $\alpha$ & Epochs & $\eta$ & $(\beta_1,\beta_2)$ \\
\midrule
Fashion-MNIST / LeNet-5 & Ring & 16 & $0.1,\ 0.01$ & 100 & $0.1$ & $(0.95,0.85)$ \\
CIFAR-10 / VGG-11 & Ring & 16, 32 & $0.1,\ 0.01$ & 200 & $5.0$ & $(0.95,0.85)$ \\
CIFAR-10 / ResNet-20 & Ring & 16, 32 & $0.1,\ 0.01$ & 200 & $5.0$ & $(0.90,0.80)$ \\
CIFAR-100 / ResNet-20 & Ring & 16 & $0.1,\ 0.01$ & 200 & $5.0$ & $(0.90,0.80)$ \\
CIFAR-10 / ResNet-20 & Torus & 32 & $0.01$ & 200 & $5.0$ & $(0.90,0.70)$ \\
CIFAR-10 / ResNet-20 & Dyck & 32 & $0.01$ & 200 & $5.0$ & $(0.95,0.65)$ \\
\bottomrule
\end{tabular}}
\end{table}

For DPA and DPA-1G, we use the exponential coefficient schedules
for \(a_{k+1}\) and \(b_{k+1}\) described in
Appendix~\ref{app:schedule-details}, with
\[
a_{\mathrm{start}}=0.005,\qquad a_{\mathrm{end}}=0.0001,\qquad
b_{\mathrm{start}}=0.3,\qquad b_{\mathrm{end}}=0.1,\qquad q_a=q_b=1.
\]
Let \(T\) denote the total number of training iterations.
Table~\ref{tab:dpa-schedule-lengths} gives the coefficient schedule lengths
\(T_a\) and \(T_b\), and the warmup duration \(T_w\), as fractions
of \(T\). For these comparisons, the reference training horizon
equals the total training horizon.

DPA uses two communication rounds per iteration, one for the optimization
sequence and one for the output sequence. For DPA-1G, we set \(T_x=5\).
The warmup and coefficient schedules advance on every iteration,
including the gradient-free iterations of DPA-1G.

\begin{table}[htbp]
\centering
\caption{Coefficient schedule lengths and warmup.}
\label{tab:dpa-schedule-lengths}
\begin{tabular}{@{}cccc@{}}
\toprule
$\alpha$ & $T_a/T$ & $T_b/T$ & $T_w/T$ \\
\midrule
$0.1$ & $1.0$ & $0.15$ & $0.25$ \\
$0.01$ & $1.0$ & $0.30$ & $0.50$ \\
\bottomrule
\end{tabular}
\end{table}

The coefficient bounds in Appendix~\ref{app:notation} hold with
\(a_{\min}=10^{-4}\), \(a_{\max}=5\times10^{-3}\),
\(b_{\min}=0.1\), and \(b_{\max}=0.3\).

\subsection{Effect of momentum and the \(b_k\) schedule}
\label{app:ablation-bk}

\begin{figure}[h!]
\centering
% AUTHOR CHECK (figure asset): the uploaded ablation still draws DPA in blue.
% Regenerate with DPA green and distinct colors/clear labels for the momentum-free
% variants to match the fixed-color convention; do not change the underlying data.
\includegraphics[width=0.6\linewidth]{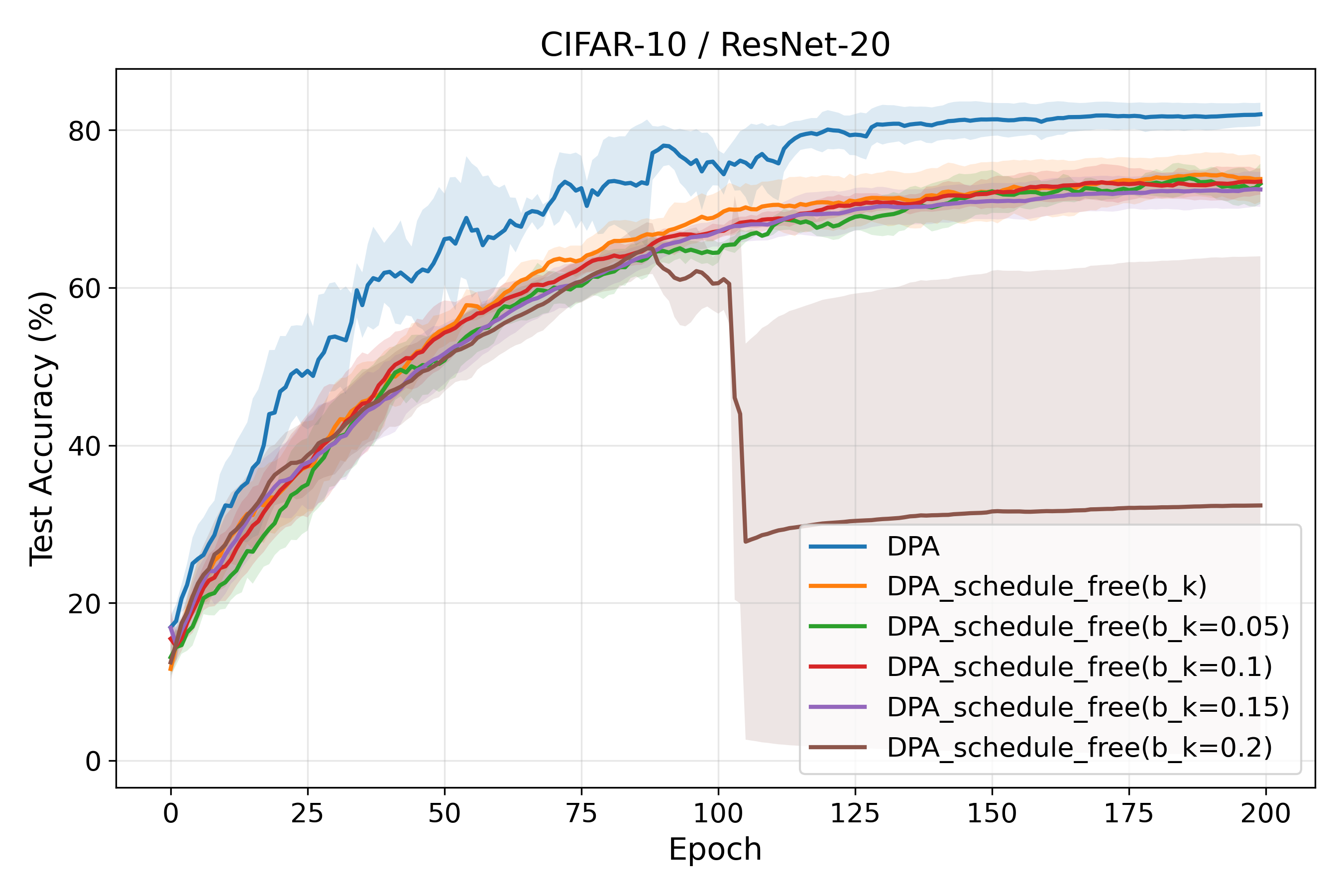}
\caption{{{Effect of quasi-global momentum and the $b_k$
schedule} on CIFAR-10/ResNet-20 ($n=16$, ring, $\alpha=0.01$).}}
\label{fig:dpa-ablation-bk}
\end{figure}

Figure~\ref{fig:dpa-ablation-bk} compares DPA with variants without
momentum on CIFAR-10/ResNet-20 using a ring topology with
$n=16$ agents and $\alpha=0.01$.
In the legend, labels starting with \texttt{DPA\_schedule\_free}
denote variants without momentum. The label with $b_k$ alone
uses a decreasing schedule, and the labels with numerical values
use fixed coefficients.

DPA has higher test accuracy early in training and at the last
epoch than the variants without momentum.
Among the variants without momentum, the decreasing $b_k$
schedule gives slightly higher test accuracy than the fixed
values tested. Training with $b_k=0.2$ is unstable in this
experiment.

\subsection{Extended results}
\label{app:extended-results}

\begin{table}[t]
\centering
\caption{{CIFAR-10/ResNet-20 test accuracy (\%) on a 32-agent ring.}}
\label{tab:n32-ring-cifar10-resnet20}
\begin{tabular}{llcc}
\toprule
$\alpha$ & Method & LE $\pm$ std & L5E $\pm$ std \\
\midrule
\multirow{7}{*}{$0.01$}
& DSGD
& {$21.52\pm6.74$} & {$21.50\pm6.70$} \\
& GT
& {$49.17\pm0.53$} & {$49.16\pm0.53$} \\
& QG-DSGDm
& {$54.52\pm1.26$} & {$55.15\pm1.34$} \\
& MT
& {$44.50\pm2.66$} & {$44.01\pm3.34$} \\
& QG-GUTm
& {$60.35\pm2.58$} & {$60.26\pm2.97$} \\
& DPA-1G
& $\mathbf{76.60 \pm 1.81}$ & {$\mathbf{76.34\pm1.81}$} \\
& DPA
& $76.04 \pm 2.35$ & {$76.13\pm2.27$} \\
\midrule
\multirow{7}{*}{$0.1$}
& DSGD
& $44.85 \pm 6.29$ & $45.13 \pm 6.31$ \\
& GT
& {$52.54\pm1.03$} & {$52.53\pm1.03$} \\
& QG-DSGDm
& $80.30 \pm 2.57$ & $80.32 \pm 2.56$ \\
& MT
& {$72.66\pm3.65$} & {$72.36\pm3.50$} \\
& QG-GUTm
& $82.16 \pm 1.50$ & $82.14 \pm 1.53$ \\
& DPA-1G
& $81.52 \pm 3.19$ & $81.48 \pm 3.18$ \\
& DPA
& $\mathbf{85.62 \pm 0.50}$ & $\mathbf{85.57 \pm 0.49}$ \\
\bottomrule
\end{tabular}
\end{table}

\begin{figure}[t]
\centering
\includegraphics[width=0.48\linewidth]{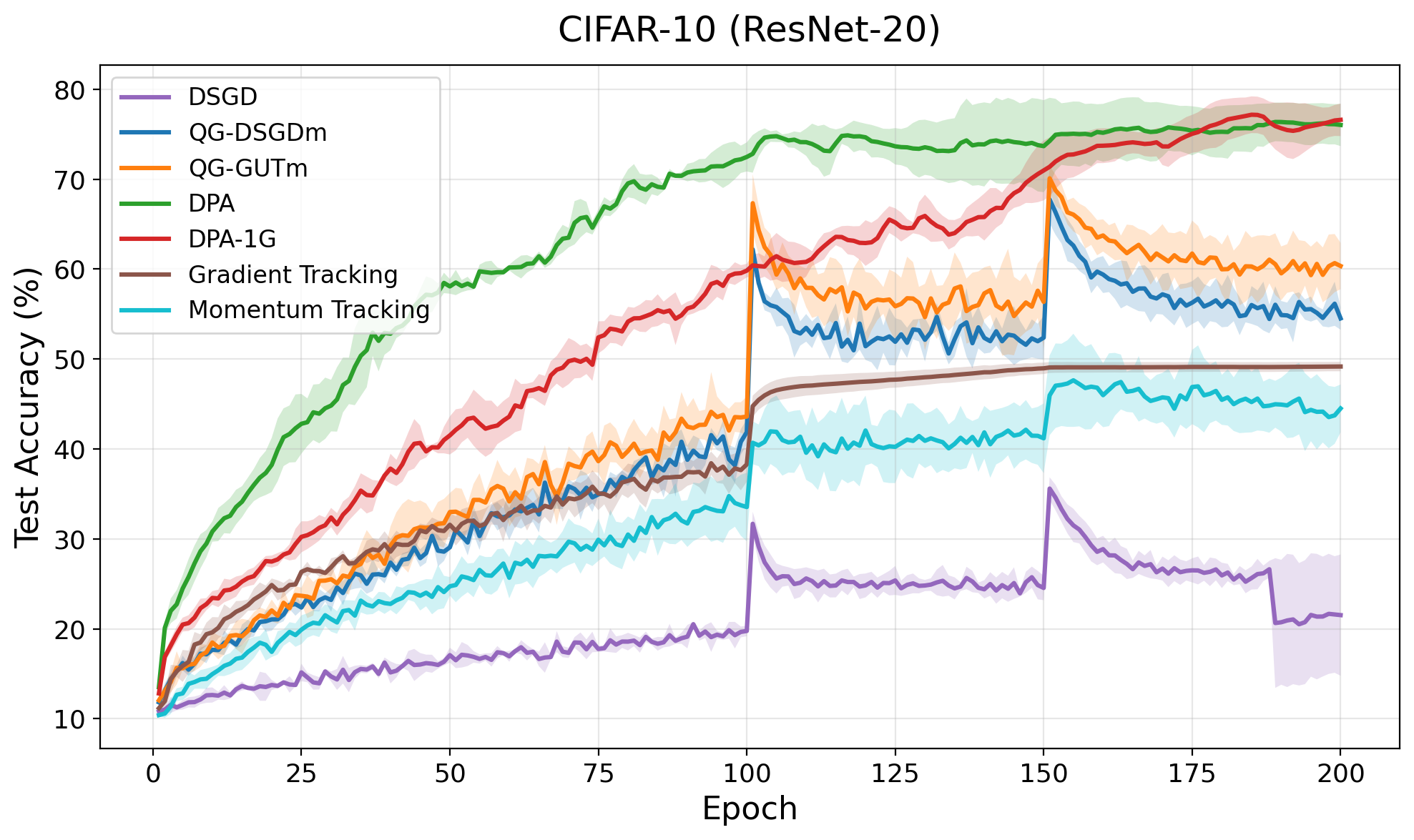}
\hfill
\includegraphics[width=0.48\linewidth]{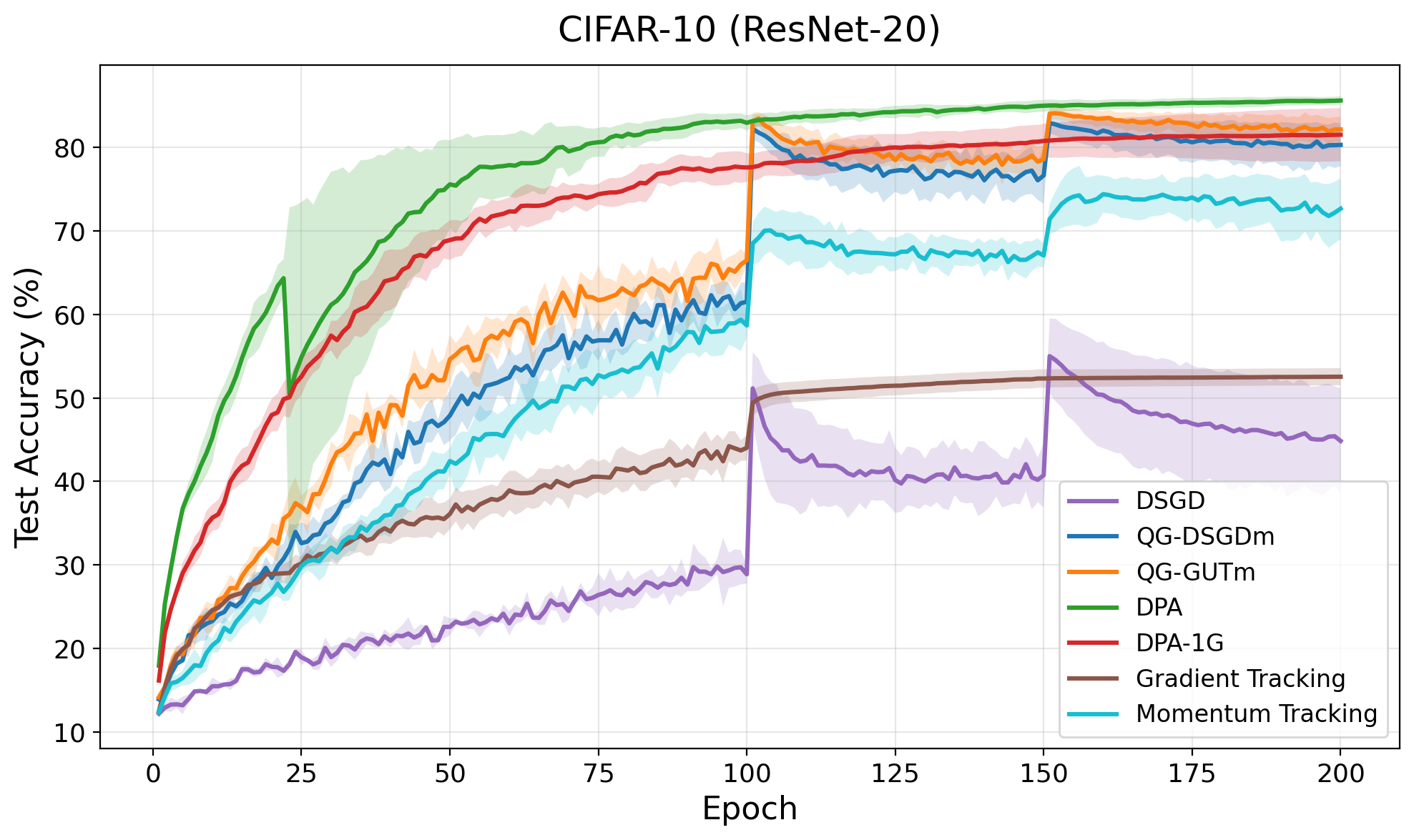}
\caption{Test accuracy curves on CIFAR-10/ResNet-20 with $n=32$ agents on
the ring topology. Left: $\alpha=0.01$. Right: $\alpha=0.1$.}
\label{fig:n32-ring-cifar10-resnet20}
\end{figure}

Figure~\ref{fig:n32-ring-cifar10-resnet20} shows the {test accuracy curves}
corresponding to Table~\ref{tab:n32-ring-cifar10-resnet20}. Under stronger
heterogeneity (\(\alpha=0.01\)), DPA-1G has the {highest mean LE
and L5E, with an L5E of \(76.34\%\) versus \(76.13\%\) for DPA}. At
moderate heterogeneity (\(\alpha=0.1\)), DPA gives the best
{mean} LE and L5E accuracy, improving over the strongest baseline,
QG-GUTm, by {\(3.43\) percentage points in L5E}.

\begin{table}[t]
\centering
\caption{{CIFAR-10/ResNet-20 test accuracy (\%) {on torus and Dyck}
graphs ($n=32$, $\alpha=0.01$).}}
\label{tab:n32-torus-dyck-cifar10-resnet20}
\begin{tabular}{llcc}
\toprule
Graph & Method & LE $\pm$ std & L5E $\pm$ std \\
\midrule
\multirow{4}{*}{Torus}
& QG-DSGDm
& {$84.77\pm1.05$} & {$84.77\pm1.06$} \\
& QG-GUTm
& {$85.92\pm0.39$} & {$85.94\pm0.36$} \\
& DPA-1G
& {$85.35\pm0.78$} & {$85.33\pm0.79$} \\
& DPA
& {$\mathbf{86.42\pm0.51}$}
& {$\mathbf{86.40\pm0.53}$} \\
\midrule
\multirow{4}{*}{Dyck}
& QG-DSGDm
& {$71.19\pm2.50$} & {$71.50\pm2.23$} \\
& QG-GUTm
& {$84.77\pm0.24$} & {$84.73\pm0.24$} \\
& DPA-1G
& {$83.54\pm0.51$} & {$83.55\pm0.51$} \\
& DPA
& {$\mathbf{84.83\pm0.48}$}
& {$\mathbf{84.82\pm0.50}$} \\
\bottomrule
\end{tabular}
\end{table}

\begin{figure}[h!]
\centering
\includegraphics[width=0.48\linewidth]{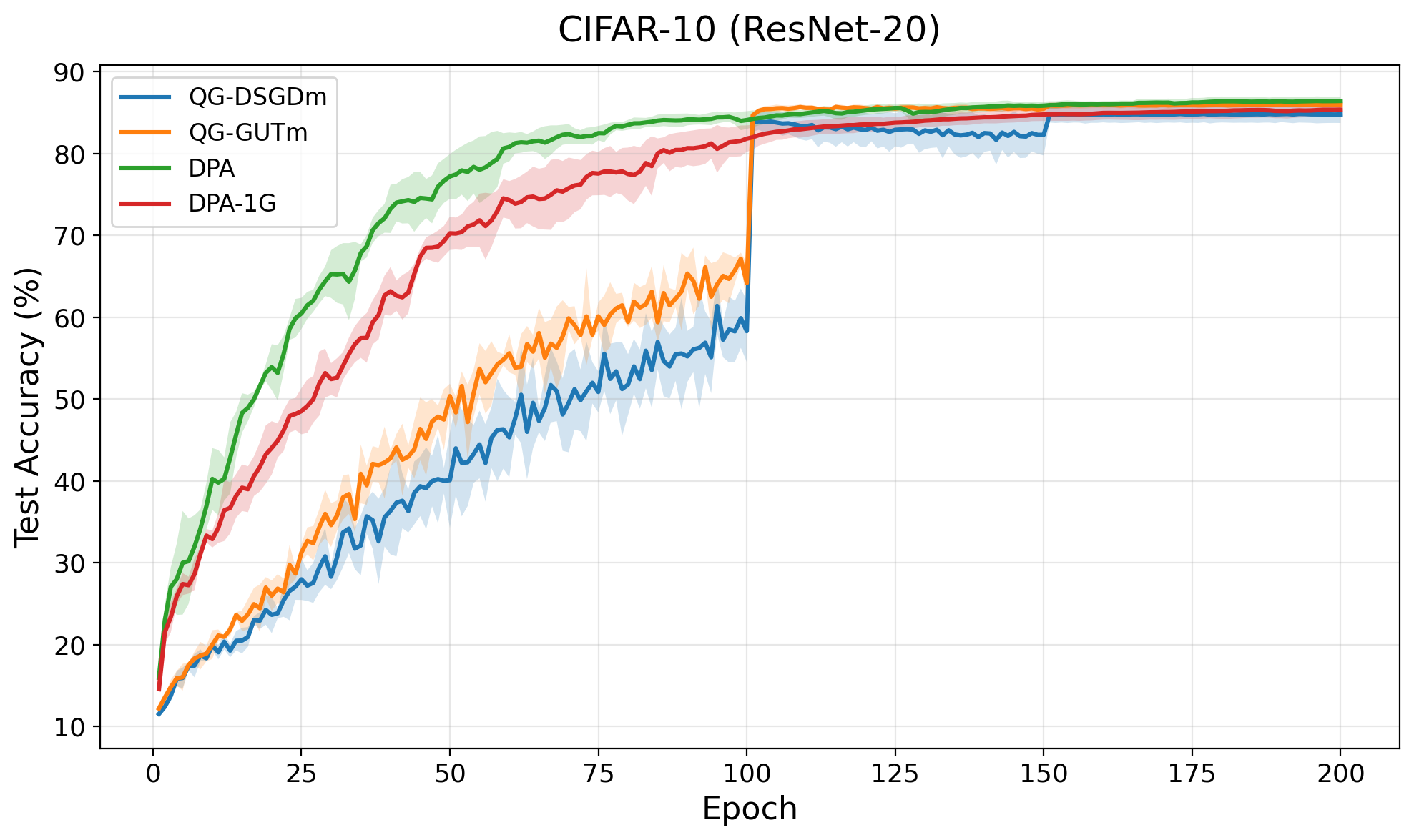}
\hfill
\includegraphics[width=0.48\linewidth]{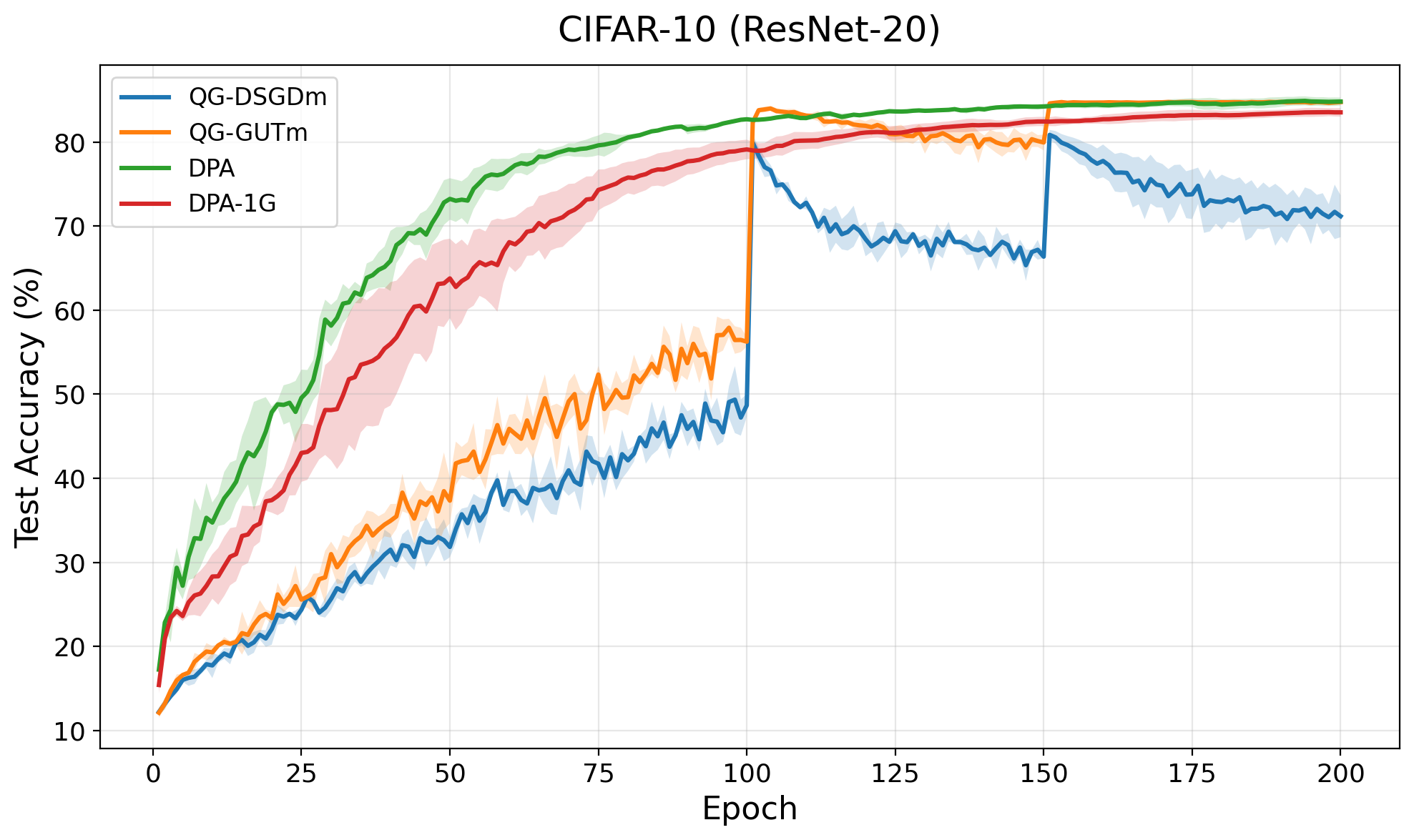}
\caption{Test accuracy curves on CIFAR-10/ResNet-20 with $n=32$ agents and
$\alpha=0.01$ on better-connected topologies. Left: Torus. Right: Dyck.}
\label{fig:n32-torus-dyck-cifar10-resnet20}
\end{figure}

Figure~\ref{fig:n32-torus-dyck-cifar10-resnet20} and
Table~\ref{tab:n32-torus-dyck-cifar10-resnet20} report results {on the torus}
and Dyck graphs at \(\alpha=0.01\). {DPA achieves the highest mean
{LE accuracy} after {choosing the {momentum parameters} separately for each graph}, with QG-GUTm
close behind.} {The mean L5E gap between the two methods is
\(0.46\) percentage points {on the torus} and \(0.09\) {on the Dyck graph}, compared with
\(15.87\) on the 32-agent ring.}

\begin{table}[h!]
\centering
\caption{{CIFAR-10/VGG-11 test accuracy (\%) on a 32-agent ring.}}
\label{tab:n32-ring-cifar10-vgg11}
% AUTHOR CHECK: QG-DSGDm, alpha=0.01, job17536264: the supplied seed-0 raw file has
% 102 and 201 evaluations per rank in separate logged attempts. In the second block,
% seed-0 LE=56.8928125 and L5E=56.195625 (32-rank means). Confirm all three
% seed-level values from complete blocks before finalizing the mean and std.
\begin{tabular}{llcc}
\toprule
$\alpha$ & Method & LE $\pm$ std & L5E $\pm$ std \\
\midrule
\multirow{4}{*}{$0.01$}
& QG-DSGDm
& {$57.48\pm0.43$} & {$57.82\pm1.32$} \\
& QG-GUTm
& $68.53 \pm 2.74$ & $69.05 \pm 2.47$ \\
& DPA-1G
& $69.26 \pm 1.77$ & $69.10 \pm 1.84$ \\
& DPA
& $\mathbf{77.47 \pm 3.99}$ & $\mathbf{77.10 \pm 4.24}$ \\
\midrule
\multirow{4}{*}{$0.1$}
& QG-DSGDm
& {$74.75\pm1.03$} & {$74.38\pm0.85$} \\
& QG-GUTm
& {$\mathbf{83.54\pm0.35}$}
& {$\mathbf{83.56\pm0.34}$} \\
& DPA-1G
& $77.27 \pm 2.47$ & {$77.21\pm2.44$} \\
& DPA
& $81.78 \pm 0.81$ & {$81.75\pm0.85$} \\
\bottomrule
\end{tabular}
\end{table}

\begin{figure}[htbp]
\centering
\includegraphics[width=0.48\linewidth]{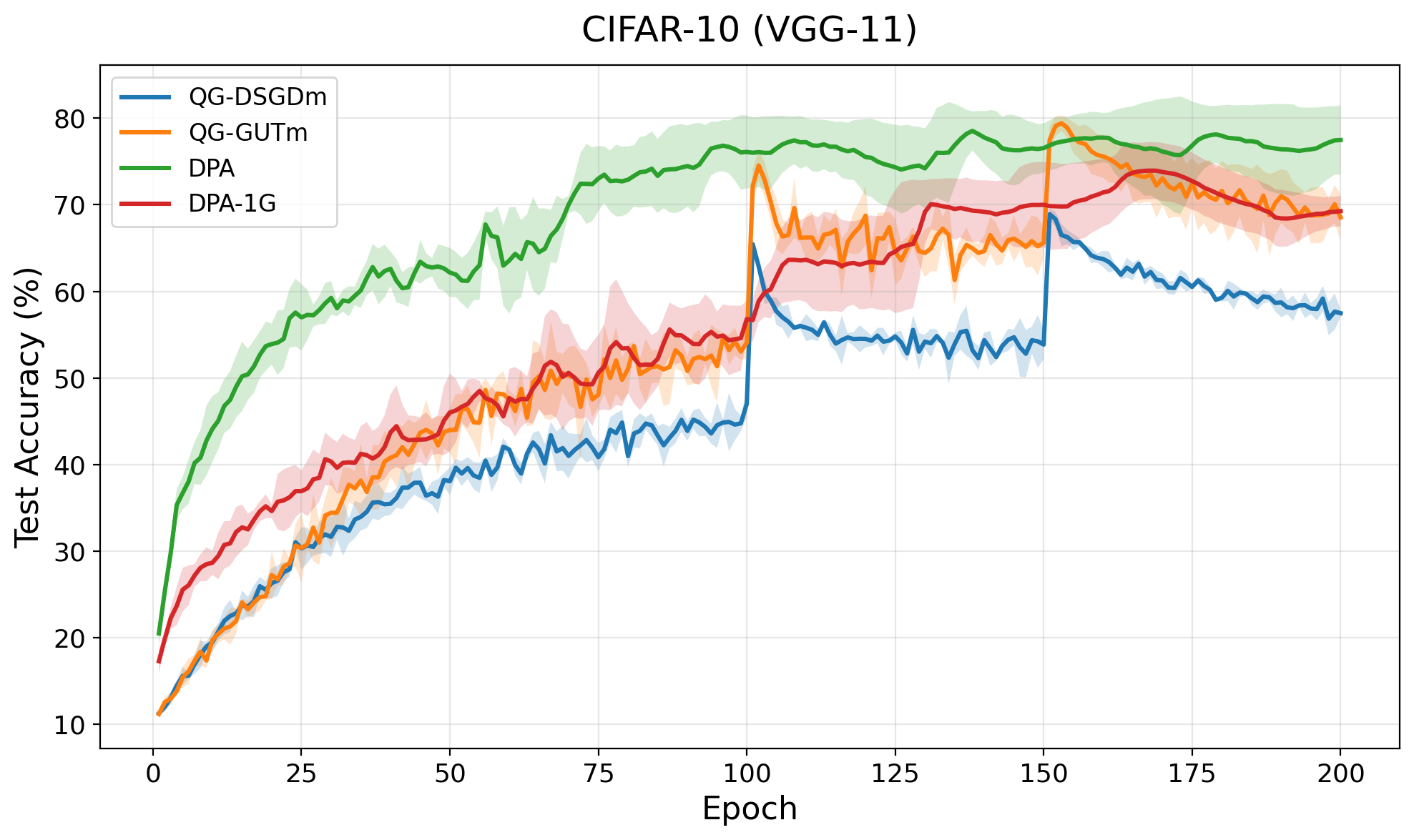}
\hfill
\includegraphics[width=0.48\linewidth]{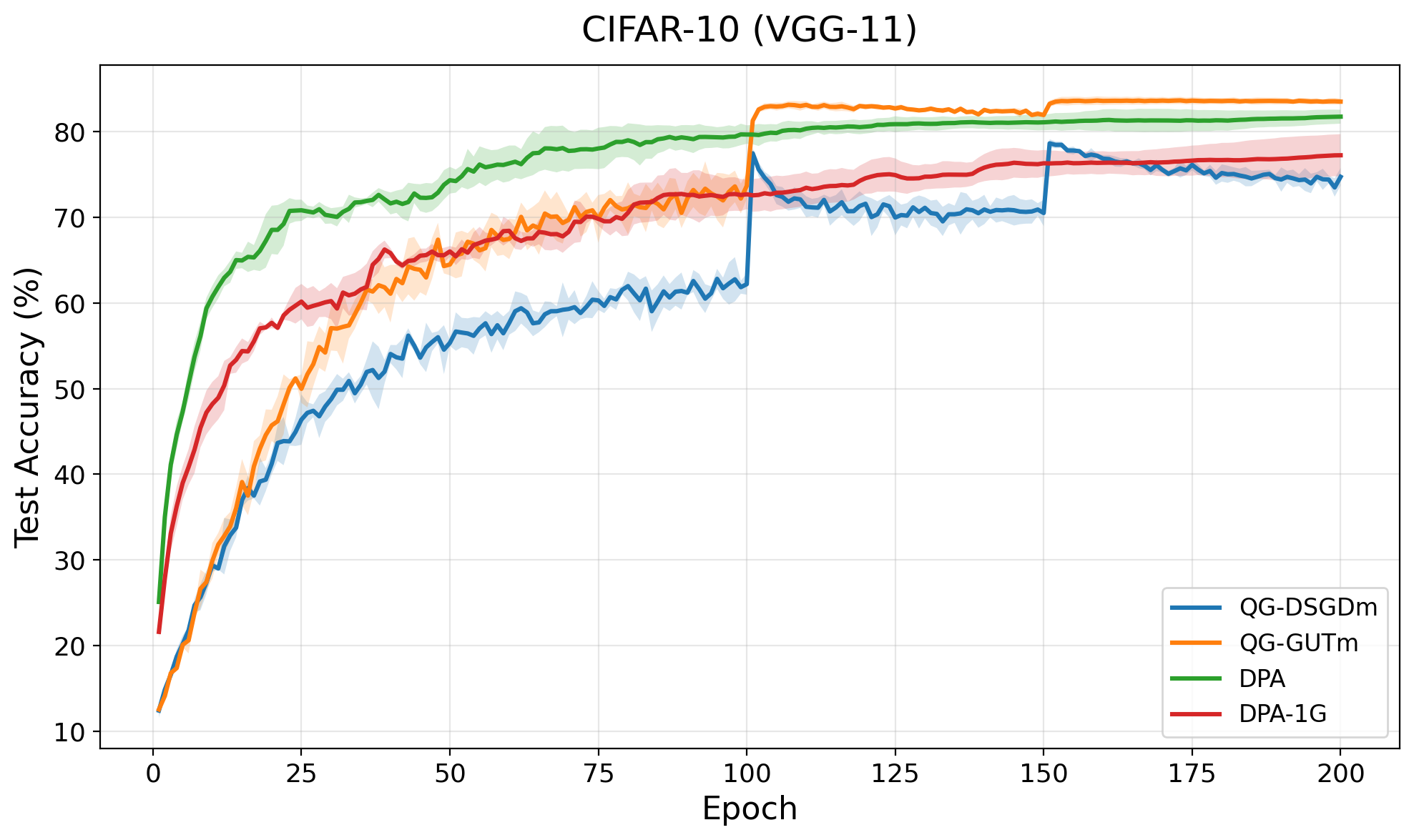}
\caption{Test accuracy curves on CIFAR-10/VGG-11 with $n=32$ agents on the
ring topology. Left: $\alpha=0.01$. Right: $\alpha=0.1$.}
\label{fig:n32-ring-cifar10-vgg11}
\end{figure}

Figure~\ref{fig:n32-ring-cifar10-vgg11} shows the CIFAR-10/VGG-11 curves
from Table~\ref{tab:n32-ring-cifar10-vgg11}. {We compare the two QG
baselines with DPA and DPA-1G at both heterogeneity levels.} At
\(\alpha=0.01\), DPA gives the best {mean} LE and L5E accuracy,
improving over QG-GUTm by about \(8\) percentage points in L5E. At
\(\alpha=0.1\), QG-GUTm {has the highest mean LE and L5E, followed
by DPA}.

\subsection{{Comparison at the same communication budget}}
\label{app:comm-matched}

DPA communicates two model-sized vectors per neighbor per iteration,
while DSGD, QG-DSGDm, QG-GUTm, and DPA-1G communicate one. GT and MT
communicate approximately two model-sized vectors per neighbor per
iteration. We therefore compare 200 epochs of DPA, GT, and MT with
separate 400-epoch runs of the {methods that communicate one model-sized
vector per neighbor per iteration}.
{We apply the baseline learning-rate decay} after epochs 100 and 150 in the
200-epoch runs and after epochs 200 and 300 in the 400-epoch runs.

For the 400-epoch DPA-1G runs, we use the same optimizer
hyperparameters as in the corresponding 200-epoch runs in
Appendix~\ref{app:hyperparams}. The reference horizon for the
coefficient schedules and warmup remains 200 epochs.

{We measure communication cost by the cumulative number of}
tensor bytes sent per {agent}, summed over {all}
neighbors and excluding received bytes, transport overhead, and {post-training averaging}. For {earlier runs}, we estimate {the bytes sent
by the end of each epoch by multiplying} the recorded training total
{by the fraction of epochs completed}. For each seed,
we average test accuracy across {agents}. {We then estimate accuracy
at each communication budget using the straight line connecting the
two adjacent epoch-end evaluations}. {We report} the mean
and standard deviation across three seeds. The full common budgets are
$B=85.45$ {GB/agent} on CIFAR-10 and $B=87.28$ {GB/agent} on CIFAR-100.

\begin{figure}[h!]
\centering
\begin{subfigure}[t]{0.49\linewidth}
  \centering
  \includegraphics[width=\linewidth]{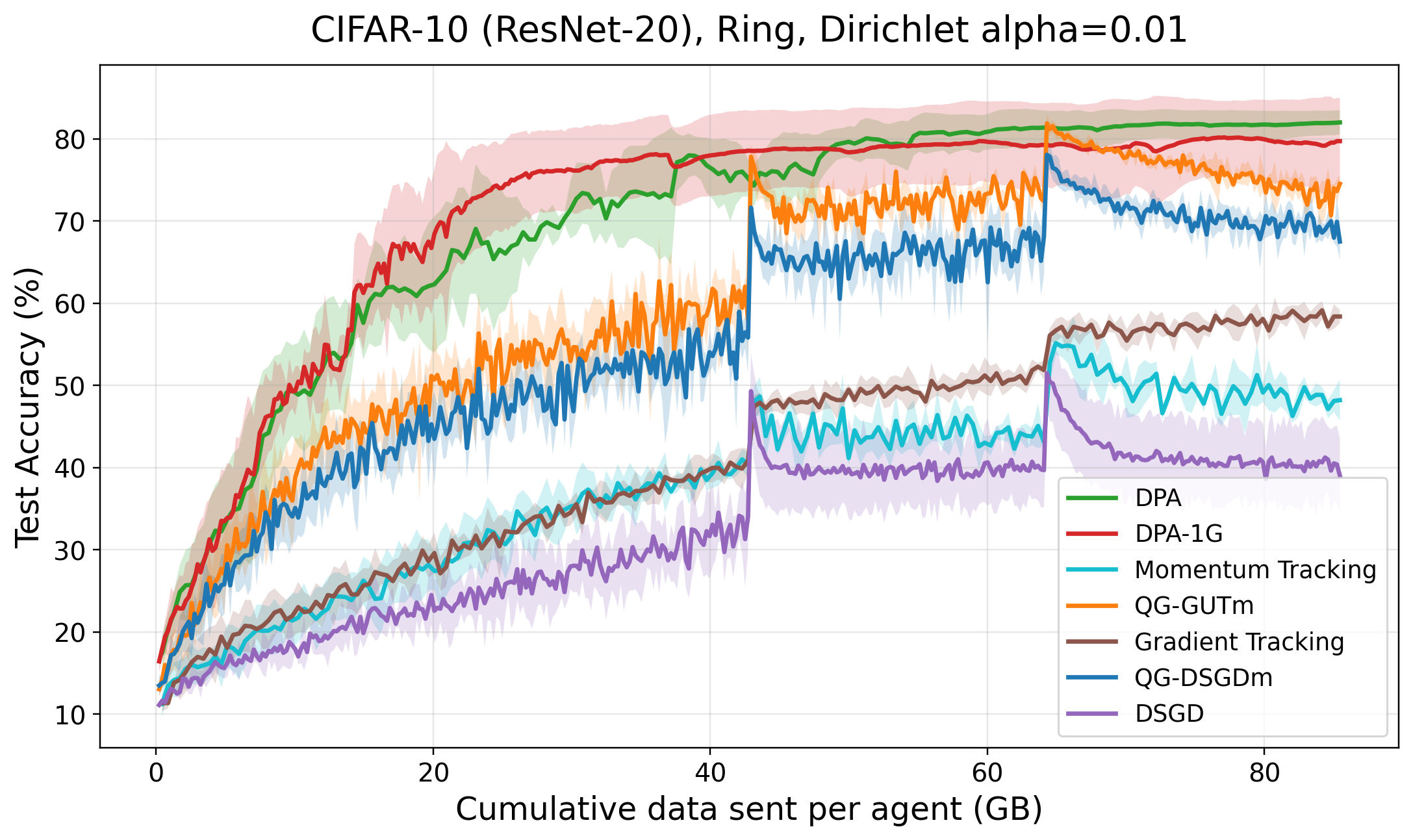}
  \caption{CIFAR-10 / ResNet-20.}
\end{subfigure}
\hfill
\begin{subfigure}[t]{0.49\linewidth}
  \centering
  \includegraphics[width=\linewidth]{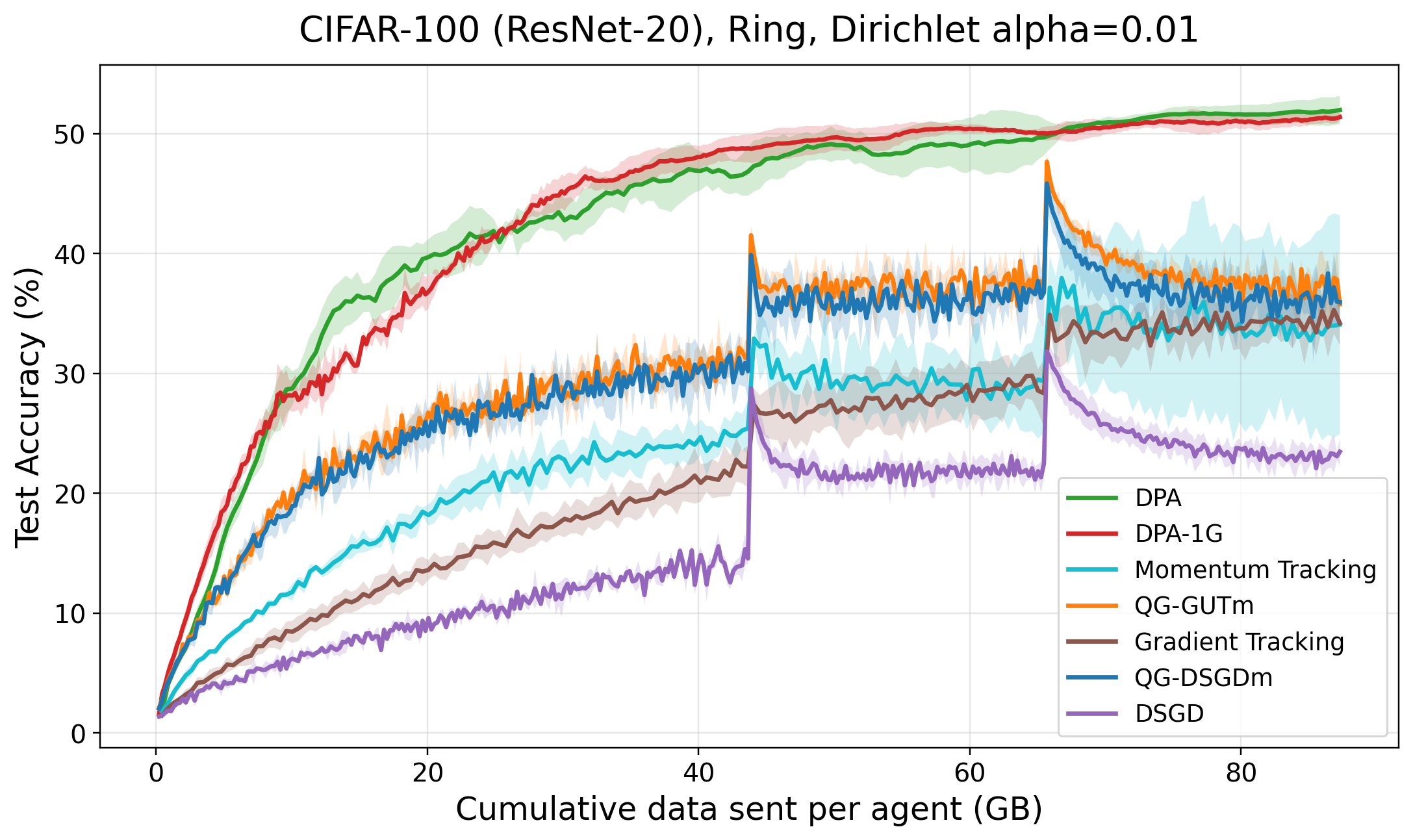}
  \caption{CIFAR-100 / ResNet-20.}
\end{subfigure}
\caption{{Test accuracy versus cumulative data sent per {agent}
(GB) on a 16-agent ring ($\alpha=0.01$).}}
\label{fig:comm-matched-curves}
\end{figure}

Figure~\ref{fig:comm-matched-curves} shows all seven methods, with mean
curves and shaded regions indicating $\pm1$ standard deviation.
Tables~\ref{tab:comm-fractions-c10} and~\ref{tab:comm-fractions-c100}
report {test accuracy for} DPA, DPA-1G, and the
{baseline with the highest mean at each budget}.
These entries are accuracies at the specified budgets, not L5E.
Boldface marks the highest mean in each column.
{A run can reach the full common budget between two evaluations,
so accuracy at $B$ may differ slightly from last-epoch accuracy (LE).}

% AUTHOR CHECK: the rebuttal promised a full seven-method budget table.
% Add the omitted method rows from verified numerical exports before
% finalizing; do not infer their values from the plotted curves.
% Keep training-byte totals separate from final-gossip totals, and
% interpolate within each seed before computing the mean and std.

\begin{table}[h!]
\centering
\caption{{{CIFAR-10/ResNet-20 test accuracy} (\%) at fractions of
$B=85.45$ {GB/agent}.}}
\label{tab:comm-fractions-c10}
\resizebox{0.95\linewidth}{!}{%
\begin{tabular}{lccccc}
\toprule
Method & $0.2B$ & $0.4B$ & $0.6B$ & $0.8B$ & $B$ \\
\midrule
{Best baseline (QG-GUTm)}
& $46.61\pm5.71$ & $56.00\pm4.58$ & $73.47\pm2.48$
& $78.39\pm0.27$ & $74.52\pm1.10$ \\
DPA-1G
& $\mathbf{67.22\pm7.11}$ & $\mathbf{77.40\pm4.71}$
& $78.72\pm5.07$ & $78.58\pm6.10$ & $79.73\pm5.23$ \\
DPA
& $61.99\pm4.62$ & $73.46\pm5.69$
& $\mathbf{80.08\pm2.46}$ & $\mathbf{81.30\pm2.23}$
& $\mathbf{82.00\pm1.47}$ \\
\bottomrule
\end{tabular}}
\end{table}

\begin{table}[h!]
\centering
\caption{{{CIFAR-100/ResNet-20 test accuracy} (\%) at fractions of
$B=87.28$ {GB/agent}.}}
\label{tab:comm-fractions-c100}
\resizebox{0.98\linewidth}{!}{%
\begin{tabular}{lccccc}
\toprule
Method & $0.2B$ & $0.4B$ & $0.6B$ & $0.8B$ & $B$ \\
\midrule
{Best baseline method}
& QG-DSGDm & QG-DSGDm & QG-GUTm & QG-GUTm & QG-DSGDm \\
{Best baseline accuracy}
& $23.36\pm1.54$ & $29.76\pm0.22$ & $37.82\pm2.92$
& $39.97\pm0.31$ & $35.94\pm1.69$ \\
DPA-1G
& $34.66\pm0.87$ & $\mathbf{46.82\pm1.28}$
& $\mathbf{49.53\pm0.97}$ & $50.51\pm0.61$ & $51.38\pm0.37$ \\
DPA
& $\mathbf{37.99\pm2.62}$ & $45.57\pm1.14$ & $48.60\pm1.26$
& $\mathbf{50.89\pm0.40}$ & $\mathbf{51.97\pm1.18}$ \\
\bottomrule
\end{tabular}}
\end{table}

DPA and DPA-1G have the two highest mean accuracies at every reported
budget fraction. At the full budget, DPA improves over the
{best baseline} by $7.48$ percentage points on CIFAR-10 and
$16.03$ {percentage points on CIFAR-100}. The corresponding gains for DPA-1G are
$5.21$ and $15.44$ {percentage points, respectively}.
{Our primary comparison uses} the full common budget;
intermediate columns {show test accuracy at smaller budgets}
under each method's own training schedule.
Equal communication does not imply identical learning-rate
schedules or equal numbers of gradient evaluations.

\subsection{IID evaluation}
\label{app:iid-results}

\begin{figure}[h!]
\centering
\includegraphics[width=0.78\linewidth]{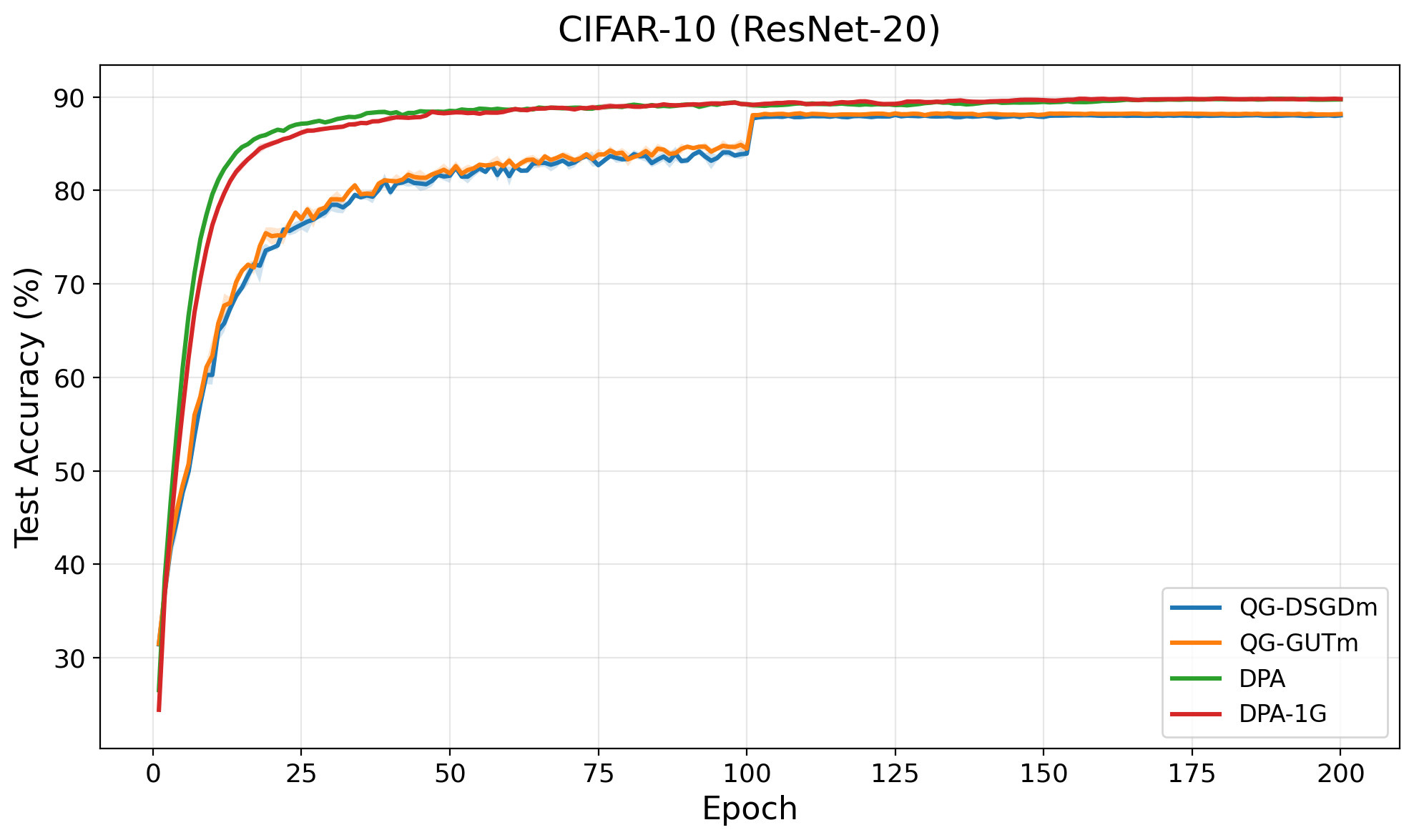}
\caption{{IID CIFAR-10/ResNet-20 test accuracy on a 16-agent ring.}}
\label{fig:iid-curves}
\end{figure}

{All four methods train ResNet-20 for 200 epochs on a 16-agent
ring, using {equal-sized}, disjoint random subsets of the
CIFAR-10 training data and three random seeds. QG-DSGDm and QG-GUTm use an initial learning rate of $0.1$
and a momentum parameter of $0.9$. QG-GUTm uses a scaling
coefficient of $0.04$. DPA and DPA-1G use a base learning
rate of $5.0$ and momentum parameters
$(\beta_1,\beta_2)=(0.90,0.80)$.
For both DPA variants, the coefficient schedules for $a_k$
and $b_k$ follow the $\alpha=0.1$ settings in
Appendix~\ref{app:hyperparams}, while learning-rate warmup
covers the first $50\%$ of training. With these IID splits,}
DPA and DPA-1G reach high {test} accuracy {in fewer epochs}
{than the QG baselines under their selected learning-rate schedules}
(Figure~\ref{fig:iid-curves}) and exceed QG-GUTm by $1.58$ and $1.64$
{percentage points in mean L5E}, respectively
(Table~\ref{tab:iid-progress}).
{Table entries report means $\pm$ standard deviations across
three seeds. Curves show the means, with shaded regions indicating
$\pm1$ standard deviation.}

\begin{table}[h!]
\centering
\caption{{IID CIFAR-10/ResNet-20: test accuracy (\%) at selected
epochs and L5E.}}
\label{tab:iid-progress}
\resizebox{0.88\linewidth}{!}{%
\begin{tabular}{lcccc}
\toprule
Training progress & QG-DSGDm & QG-GUTm & DPA-1G & DPA \\
\midrule
10\% (epoch 20)
& $73.82\pm0.04$ & $75.10\pm0.95$
& $85.01\pm0.32$ & $\mathbf{86.24\pm0.08}$ \\
20\% (epoch 40)
& $79.82\pm0.47$ & $81.02\pm0.17$
& $87.71\pm0.37$ & $\mathbf{88.25\pm0.15}$ \\
40\% (epoch 80)
& $83.36\pm0.24$ & $83.33\pm0.76$
& $89.03\pm0.08$ & $\mathbf{89.07\pm0.31}$ \\
50\% (epoch 100)
& $83.93\pm0.26$ & $84.47\pm0.47$
& $\mathbf{89.22\pm0.09}$ & $89.20\pm0.29$ \\
60\% (epoch 120)
& $87.91\pm0.16$ & $88.14\pm0.10$
& $\mathbf{89.53\pm0.19}$ & $89.21\pm0.12$ \\
80\% (epoch 160)
& $87.97\pm0.20$ & $88.19\pm0.13$
& $\mathbf{89.80\pm0.14}$ & $89.58\pm0.05$ \\
100\% (epoch 200)
& $88.03\pm0.25$ & $88.17\pm0.10$
& $\mathbf{89.78\pm0.14}$ & $89.73\pm0.17$ \\
L5E
& $88.01\pm0.26$ & $88.14\pm0.12$
& $\mathbf{89.78\pm0.14}$ & $89.72\pm0.22$ \\
\bottomrule
\end{tabular}}
\end{table}

\subsection{{{Training time}} {and GPU} {{usage}}}
\label{app:system-cost}

We run all seven methods sequentially {on the same} four
V100-32GB GPUs. Each run uses ResNet-20 on CIFAR-10, a 16-agent ring,
$\alpha=0.01$, seed 0, and 200 epochs
(Table~\ref{tab:system-cost}). {Total time} includes training,
{evaluation}, and scheduled checkpoint saving.
{{Training time} covers {mini-batch} training. GPU-hours include
all four GPUs. {Communication cost and the number of mini-batch gradient
evaluations are reported per agent}.}

{In this timing experiment}, DSGD and QG-DSGDm use direct
implementations of their published updates.
{These implementations avoid the GUT reconstruction and
tracking code that still runs in the benchmark when the corresponding
coefficients are set to zero.} For QG-GUTm, we use the
{original GUT implementation}.
{We report these single-seed timings separately from the
benchmark accuracy results, which use three seeds.}

% AUTHOR CHECK: specify how GPU utilization and peak memory are
% measured and aggregated across the four GPUs.

\begin{table}[t]
\centering
\caption{{Training costs on four V100 GPUs
(200 epochs, seed 0).}}
\label{tab:system-cost}
\resizebox{\linewidth}{!}{%
\begin{tabular}{lrrrrrrr}
\toprule
Method & {Total time} & {Training time} & GPU-hours & GPU util. & Peak mem.
& {GB/agent} & {Grad. evals/agent} \\
\midrule
DSGD (direct)
& 45.84 min & 34.52 min & 3.056 & 61.83\% & 4982 MiB
& 42.72 & 19,600 \\
QG-DSGDm (direct)
& 46.81 min & 35.51 min & 3.121 & 62.64\% & 4990 MiB
& 42.72 & 19,600 \\
QG-GUTm
& 115.80 min & 104.78 min & 7.720 & 84.40\% & 5006 MiB
& 42.72 & 19,600 \\
{GT}
& 51.62 min & 40.33 min & 3.441 & 60.98\% & 4990 MiB
& 85.45 & 19,600 \\
{MT}
& 52.59 min & 41.23 min & 3.506 & 67.42\% & 4990 MiB
& 85.45 & 19,600 \\
DPA-1G
& 47.17 min & 35.79 min & 3.145 & 62.79\% & 4998 MiB
& 42.72 & 15,680 \\
DPA
& 54.13 min & 42.82 min & 3.609 & 67.89\% & 4996 MiB
& 85.45 & 19,600 \\
\bottomrule
\end{tabular}}
\end{table}

DPA-1G takes $0.8\%$ more {total time} than direct QG-DSGDm at the same
{communication cost}. DPA takes $15.6\%$ more time than direct QG-DSGDm,
but $4.9\%$ and $2.9\%$ more than GT and MT, respectively, which
{have the same communication cost as DPA}. Peak memory varies by less than
$0.5\%$ across methods.
{These measurements apply to the implementations and hardware
used in this experiment.}

\subsection{{Effect of post-training averaging}}
\label{app:extra-gossip}

Table~\ref{tab:extra-gossip} compares the five baselines before and after
the {{post-training averaging described}} in
Remark~\ref{rem:gut-evaluation}.
{After training, each baseline} performs $K_{\mathrm{post}}$
{communication rounds} without taking a gradient step. DPA and DPA-1G do not
use this phase; their LE values are shown for reference.
Values are means $\pm$ standard deviations across three seeds.
Dashes indicate that {no post-training averaging was used}.

We report this evaluation separately because {{post-training averaging increases communication cost and changes} the local models after
training ends}. {The evaluation after post-training averaging} is not part of LE or L5E, and
these rounds are excluded from the matched training budgets.

\begin{table}[h!]
\centering
\caption{{Test accuracy (\%) before and after {post-training averaging} ($n=16$, ring).}}
\label{tab:extra-gossip}
\begingroup
\small
\setlength{\tabcolsep}{4pt}
\resizebox{\linewidth}{!}{%
\begin{tabular}{lcccccc}
\toprule
& \multicolumn{2}{c}{Fashion-MNIST / LeNet-5}
& \multicolumn{2}{c}{CIFAR-10 / ResNet-20}
& \multicolumn{2}{c}{CIFAR-100 / ResNet-20} \\
\cmidrule(lr){2-3}\cmidrule(lr){4-5}\cmidrule(lr){6-7}
Method & LE & {After averaging} & LE & {After averaging} & LE & {After averaging} \\
\midrule
\multicolumn{7}{c}{$\alpha=0.1$} \\
\midrule
DSGD
& $85.605\pm2.962$ & $88.643\pm0.906$
& $57.709\pm3.081$ & $70.041\pm2.130$
& $25.526\pm0.475$ & $38.643\pm0.803$ \\
GT
& $84.517\pm0.156$ & $84.520\pm0.156$
& $69.902\pm2.731$ & $71.925\pm2.230$
& $41.946\pm1.353$ & $43.790\pm0.978$ \\
QG-DSGDm
& $89.717\pm0.230$ & $89.820\pm0.193$
& $81.732\pm1.576$ & $84.430\pm1.247$
& $44.892\pm1.846$ & $51.227\pm3.480$ \\
MT
& $90.064\pm0.034$ & $90.052\pm0.025$
& $79.829\pm1.644$ & $81.453\pm1.185$
& $49.179\pm3.432$ & $54.243\pm1.066$ \\
QG-GUTm
& $89.667\pm0.002$ & $89.815\pm0.121$
& $85.338\pm1.422$ & $86.307\pm0.645$
& $45.038\pm0.731$ & $53.368\pm0.528$ \\
\midrule
DPA-1G
& $89.33\pm0.65$ & \textemdash
& $86.582\pm0.854$ & \textemdash
& $53.920\pm0.546$ & \textemdash \\
DPA
& $90.08\pm0.28$ & \textemdash
& $86.727\pm1.103$ & \textemdash
& $54.120\pm0.970$ & \textemdash \\
\midrule
\multicolumn{7}{c}{$\alpha=0.01$} \\
\midrule
DSGD
& $73.930\pm2.985$ & $84.179\pm0.348$
& $29.786\pm4.599$ & $48.558\pm4.961$
& $16.136\pm0.817$ & $31.525\pm1.458$ \\
GT
& $84.102\pm0.226$ & $84.100\pm0.220$
& $58.381\pm0.811$ & $61.389\pm0.741$
& $34.195\pm1.852$ & $38.968\pm0.407$ \\
QG-DSGDm
& $76.456\pm3.624$ & $82.170\pm2.618$
& $61.470\pm1.933$ & $74.365\pm2.636$
& $33.565\pm0.797$ & $48.484\pm2.842$ \\
MT
& $82.090\pm4.793$ & $82.585\pm4.823$
& $48.204\pm2.493$ & $43.180\pm5.789$
& {$30.090\pm2.627$} & {$39.278\pm3.460$} \\
QG-GUTm
& $76.360\pm5.199$ & $84.459\pm1.710$
& $66.474\pm1.744$ & $77.795\pm1.480$
& $35.544\pm1.008$ & $48.018\pm1.883$ \\
\midrule
DPA-1G
& $85.60\pm2.46$ & \textemdash
& $78.570\pm4.890$ & \textemdash
& $48.770\pm1.164$ & \textemdash \\
DPA
& $84.73\pm3.97$ & \textemdash
& $82.005\pm1.467$ & \textemdash
& $51.971\pm1.180$ & \textemdash \\
\bottomrule
\end{tabular}%
}
\endgroup
\end{table}

{{Post-training averaging} can improve test accuracy, but the
effect varies by method and data distribution.}
At $\alpha=0.01$, both DPA and DPA-1G have higher mean accuracy
{without post-training averaging} than all five baselines {after post-training averaging} on the
three tasks shown. On CIFAR-10 and CIFAR-100, DPA
{exceeds the best baseline {after post-training averaging}} by $4.21$ and $3.49$
percentage points, respectively.
At $\alpha=0.1$ on CIFAR-100, {MT has a slightly higher mean accuracy after
post-training averaging than DPA's mean LE accuracy}
($54.243\%$ versus $54.120\%$).
{These differences in mean accuracy do not establish
statistical significance.}
These evaluations do not show whether the gains persist
{when local gradient updates resume}.

\end{document}